\documentclass[10pt,twocolumn,letterpaper]{article}

\RequirePackage{snapshot}
\usepackage{cvpr}              %

\usepackage{graphicx}
\usepackage{amsmath}
\usepackage{amssymb}
\usepackage{booktabs}
\usepackage{balance}
\usepackage{acronym} 
\usepackage{comment}
\usepackage{tabularx}
\usepackage[numbers,sort,compress]{natbib}

\usepackage{mdframed}%

\usepackage[pagebackref,breaklinks,colorlinks]{hyperref}

\usepackage[capitalize]{cleveref}
\crefname{section}{Sec.}{Secs.}
\Crefname{section}{Section}{Sections}
\Crefname{table}{Table}{Tables}
\crefname{table}{Tab.}{Tabs.}

\acrodef{MVS}{multi-view stereo}
\acrodef{3DMM}{3D morphable model}
\acrodef{ICP}{iterative closest point}
\acrodef{SDF}{signed distance field}

\def\cvprPaperID{376} %
\def\confName{CVPR}
\def\confYear{2023}

\newcommand{\red}[1]{\textcolor{red}{#1}}
\renewcommand{\paragraph}[1]{\noindent\textbf{#1}}

\newcommand{\modelname}{TEMPEH\xspace}
\newcommand{\modelnamelong}{\modelname~(Towards Estimation of 3D Meshes from Performances of Expressive Heads)\xspace}

\begin{document}

\title{Instant Multi-View Head Capture through Learnable Registration}

\author{
Timo Bolkart$^1$
\and
Tianye Li$^2$
\and
Michael J. Black$^1$
\and
\textrm{
$^1$MPI for Intelligent Systems, T{\"u}bingen  
\qquad 
$^2$University of Southern California}
}

\newcommand{\colmargin}{0.0}

\newcommand{\imgsize}{0.14\linewidth}
\newcommand{\imgcropleft}{0}
\newcommand{\imgcroplower}{400}
\newcommand{\imgcropright}{400}
\newcommand{\imgcropupper}{0}

\newcommand{\meshsize}{0.14\linewidth}
\newcommand{\meshimgcropleft}{110}
\newcommand{\meshimgcroplower}{75}
\newcommand{\meshimgcropright}{120}
\newcommand{\meshimgcropupper}{40}

\twocolumn[{
    \renewcommand\twocolumn[1][]{#1}
    \maketitle
    \centering
    \begin{minipage}{1.00\textwidth}
        \centering
        \begin{tabular}{@{}c@{\hskip \colmargin\linewidth}c@{\hskip \colmargin\linewidth}c@{\hskip \colmargin\linewidth}c@{\hskip \colmargin\linewidth}c@{\hskip \colmargin\linewidth}c@{\hskip \colmargin\linewidth}c@{\hskip \colmargin\linewidth}c@{\hskip \colmargin\linewidth}c@{}}
            \includegraphics[width=\imgsize, clip]{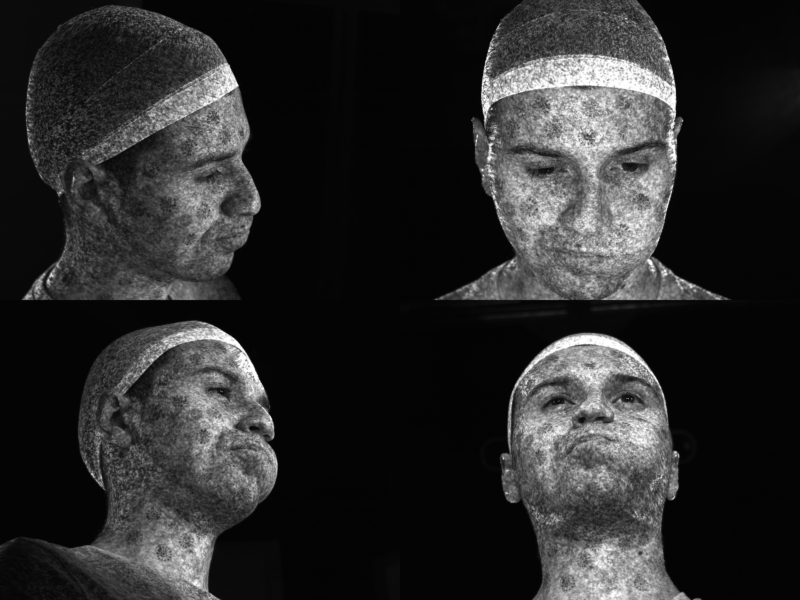} &
            \includegraphics[width=\imgsize, clip]{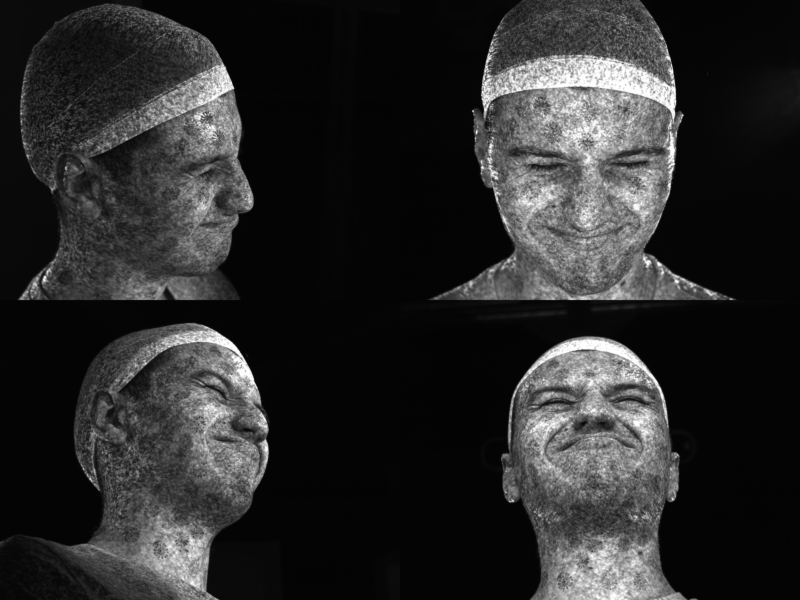} &
            \includegraphics[width=\imgsize, clip]{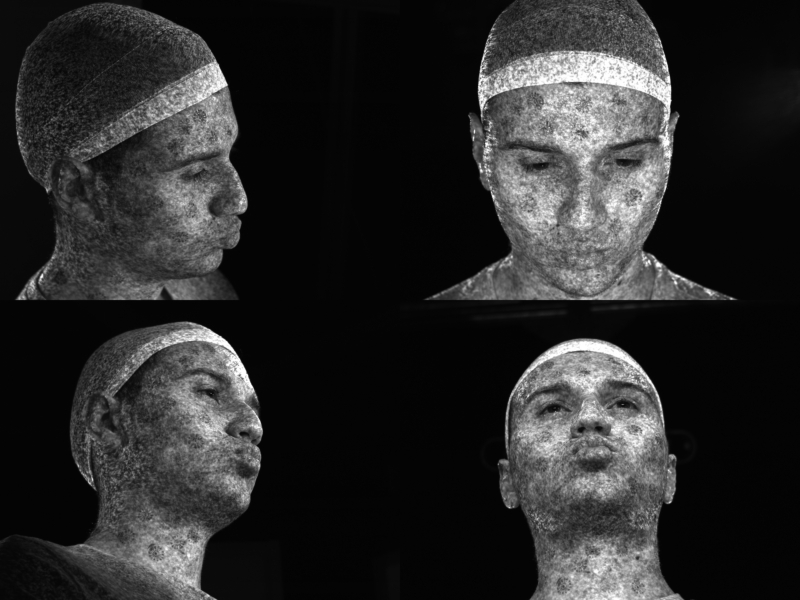} &
            \includegraphics[width=\imgsize, clip]{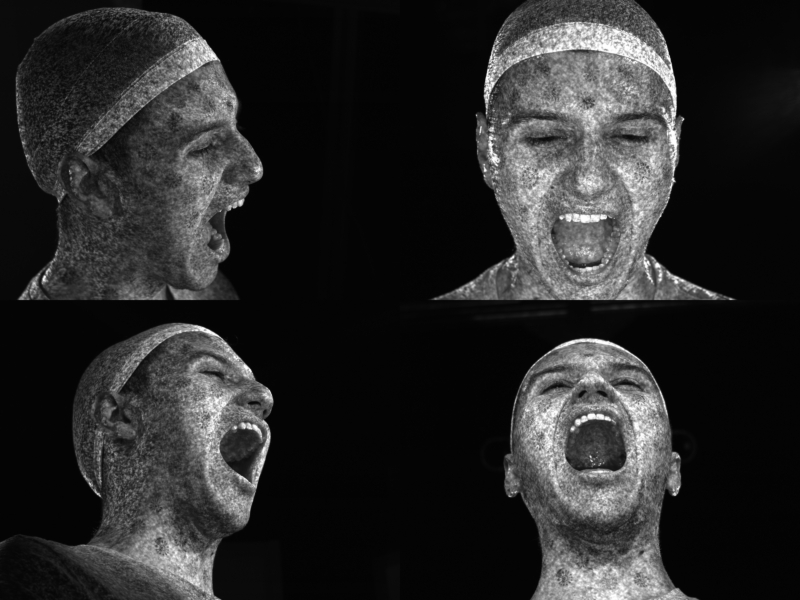} &
            \includegraphics[width=\imgsize, clip]{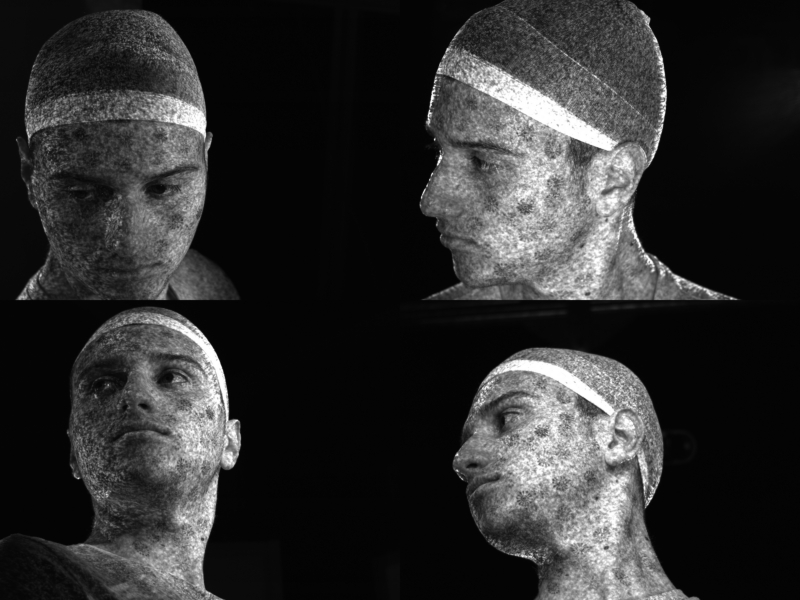} &
            \includegraphics[width=\imgsize, clip]{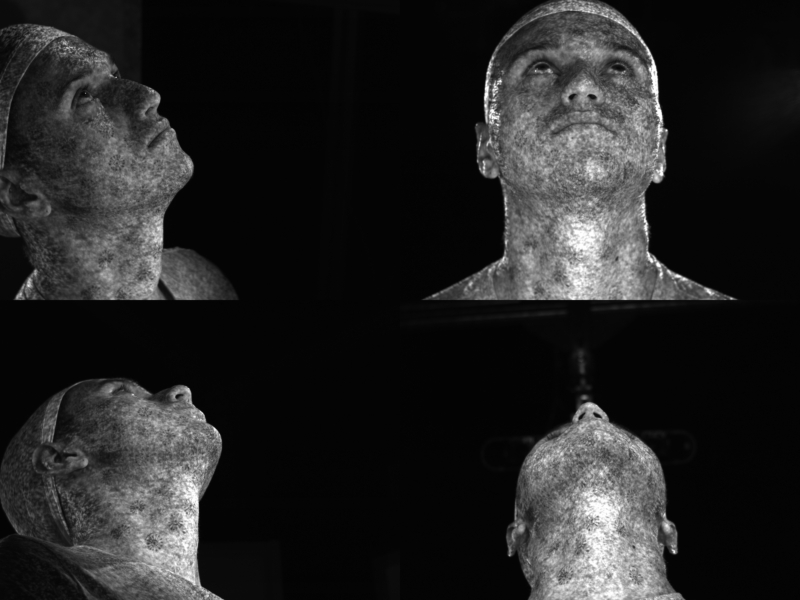} &
            \includegraphics[width=\imgsize, clip]{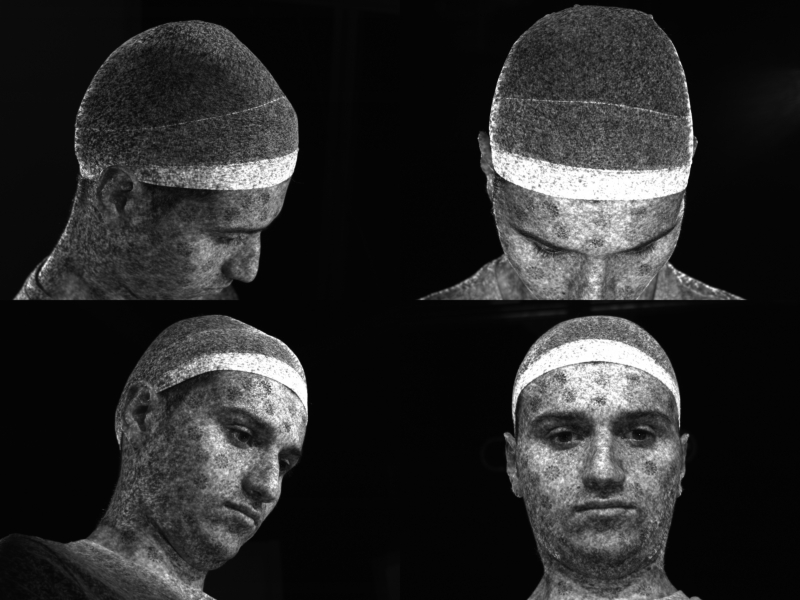} \\               
            \includegraphics[width=\meshsize, clip, trim={\meshimgcropleft} {\meshimgcroplower} {\meshimgcropright} {\meshimgcropupper}]{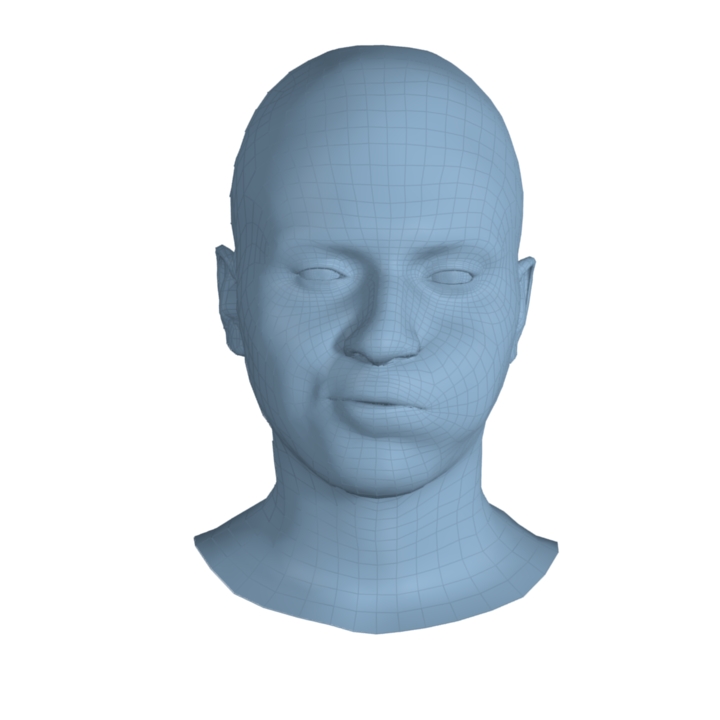} & 
            \includegraphics[width=\meshsize, clip, trim={\meshimgcropleft} {\meshimgcroplower} {\meshimgcropright} {\meshimgcropupper}]{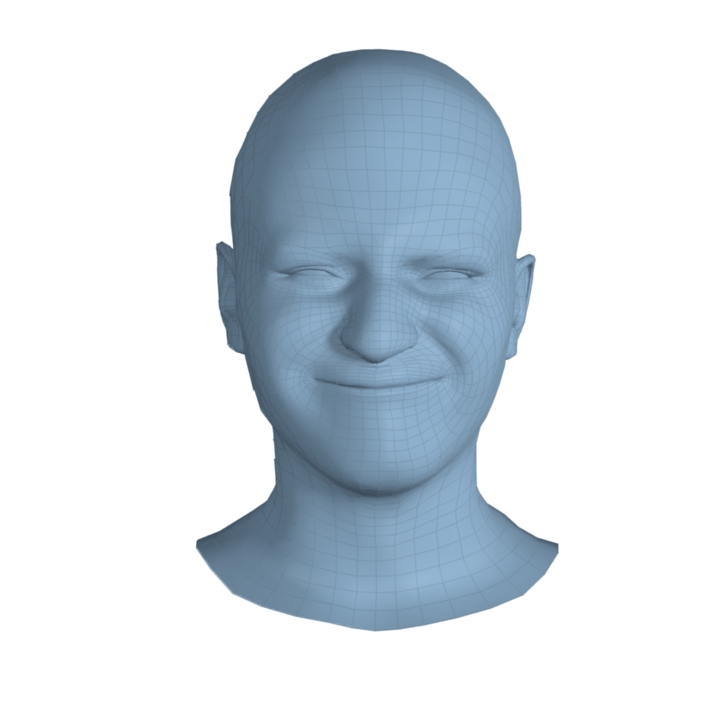} & 
            \includegraphics[width=\meshsize, clip, trim={\meshimgcropleft} {\meshimgcroplower} {\meshimgcropright} {\meshimgcropupper}]{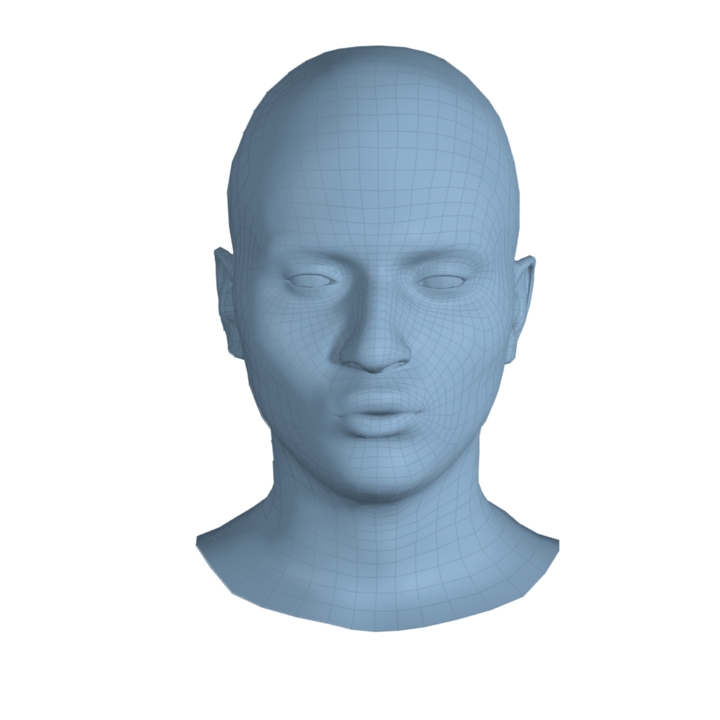} & 
            \includegraphics[width=\meshsize, clip, trim={\meshimgcropleft} {\meshimgcroplower} {\meshimgcropright} {\meshimgcropupper}]{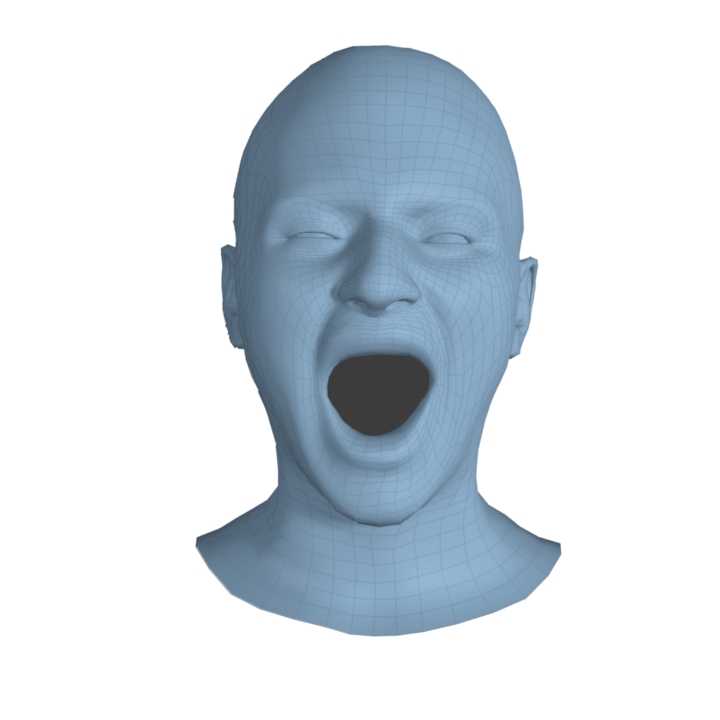} & 
            \includegraphics[width=\meshsize, clip, trim={\meshimgcropleft} {\meshimgcroplower} {\meshimgcropright} {\meshimgcropupper}]{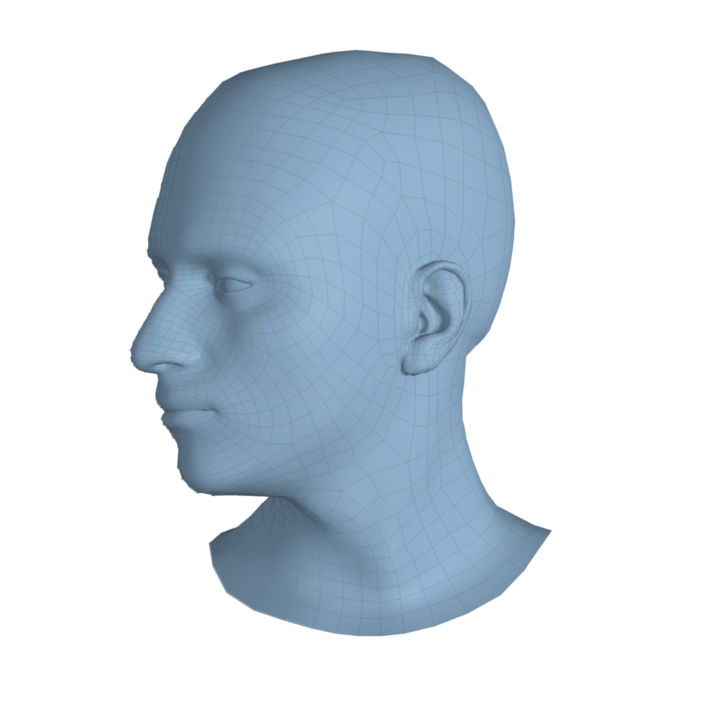} &             
            \includegraphics[width=\meshsize, clip, trim={\meshimgcropleft} {\meshimgcroplower} {\meshimgcropright} {\meshimgcropupper}]{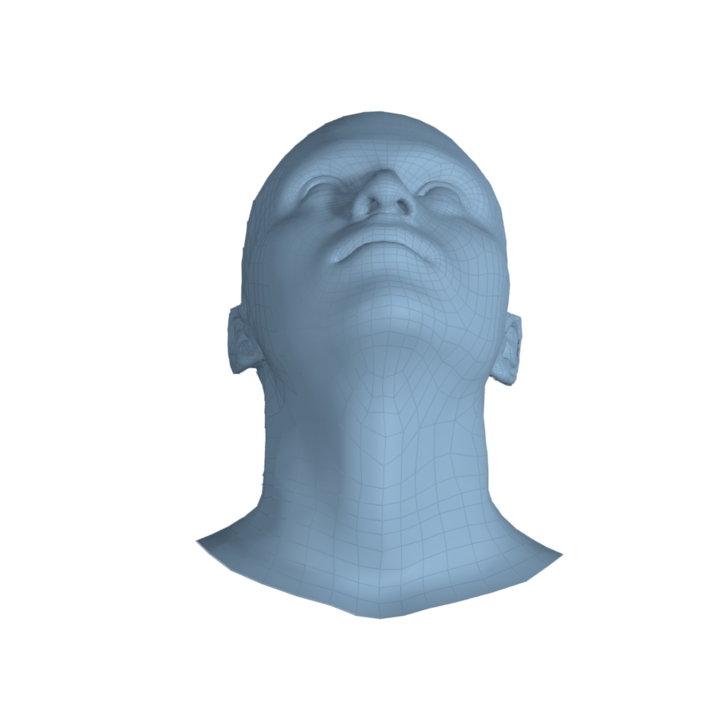} & 
            \includegraphics[width=\meshsize, clip, trim={\meshimgcropleft} {\meshimgcroplower} {\meshimgcropright} {\meshimgcropupper}]{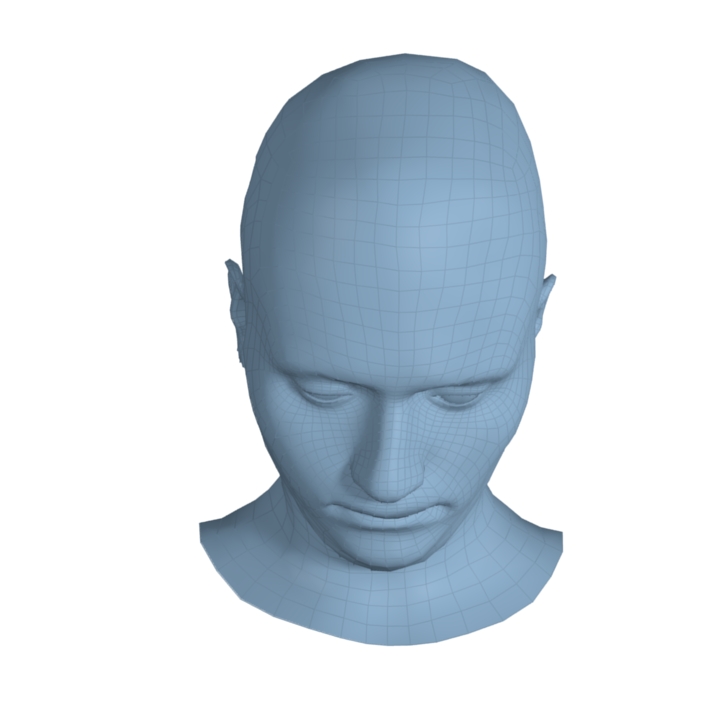}
        \end{tabular}
    \end{minipage}
    \vspace{-0.1in}
    \captionof{figure}
    {
        Given calibrated multi-view images (top: 4 of 16 views; contrast enhanced for visualization), \modelname directly infers 3D head meshes in dense semantic correspondence (bottom) in about 0.3 seconds. 
        \modelname reconstructs heads with varying expressions (left) and head poses (right) for subjects unseen during training.
        Applied to multi-view video input, the frame-by-frame inferred meshes are temporally coherent, making them directly applicable to full-head performance-capture applications. 
        See Sup.~Mat.~for the video output. 
    }
    \label{fig:teaser}
    \vspace{8pt}
}]

\begin{abstract}

Existing methods for capturing datasets of 3D heads in dense semantic correspondence are slow and commonly address the problem in two separate steps; \ac{MVS} reconstruction followed by non-rigid registration.
To simplify this process, we introduce \modelnamelong to directly infer 3D heads in dense correspondence from calibrated multi-view images.
Registering datasets of 3D scans typically requires manual parameter tuning to find the right balance between accurately fitting the scans' surfaces and being robust to scanning noise and outliers. 
Instead, we propose to jointly register a 3D head dataset while training \modelname. 
Specifically, during training, we minimize a geometric loss commonly used for surface registration, effectively leveraging \modelname as a regularizer.
Our multi-view head inference builds on a volumetric feature representation that samples and fuses features from each view using camera calibration information.
To account for partial occlusions and a large capture volume that enables head movements, we use view- and surface-aware feature fusion, and a spatial transformer-based head localization module, respectively. 
We use raw \ac{MVS} scans as supervision during training, but, once trained, \modelname directly predicts 3D heads in dense correspondence without requiring scans.
Predicting one head takes about $0.3$ seconds with a median reconstruction error of 0.26 mm, 64\% lower than the current state-of-the-art. 
This enables the efficient capture of large datasets containing multiple people and diverse facial motions.
Code, model, and data are publicly available at \url{https://tempeh.is.tue.mpg.de}.
\end{abstract}

\newcommand{\scalar}[1]{\lowercase{#1}}
\renewcommand{\vector}[1]{\textbf{\lowercase{#1}}}
\renewcommand{\matrix}[1]{\textbf{\uppercase{#1}}}
\newcommand{\tensor}[1]{\mathcal{\uppercase{#1}}}
\newcommand{\misc}[1]{\uppercase{#1}}

\newcommand{\image}{\tensor{I}}
\newcommand{\imagewidth}{\scalar{w}}
\newcommand{\imageheight}{\scalar{h}}
\newcommand{\calibration}{\misc{C}}
\newcommand{\camcenter}{\vector{o}}
\newcommand{\numviews}{\scalar{k}}

\newcommand{\gmo}{\rho}
\newcommand{\vertex}{\vector{v}}
\newcommand{\vertices}{\matrix{V}}
\newcommand{\normal}{\vector{n}}
\newcommand{\vertexvisibility}{\delta}

\newcommand{\scan}{\misc{S}}
\newcommand{\scanpoint}{\vector{s}}

\newcommand{\registration}{\misc{T}}
\newcommand{\registrationverts}{\vertices^{t}}
\newcommand{\registrationvertex}{\vertex^{t}}
\newcommand{\registrationedge}{\vector{e}^{t}}

\newcommand{\mesh}{\misc{M}}
\newcommand{\meshpoint}{\vector{m}}
\newcommand{\meshvertex}{\vertex^m}
\newcommand{\meshedge}{\vector{e}^m}

\newcommand{\reconmesh}{\mesh_r}
\newcommand{\reconverts}{\vertices^r}
\newcommand{\reconfaces}{\matrix{T}}
\newcommand{\numreconverts}{\scalar{n}_v}
\newcommand{\numreconfaces}{\scalar{n}_f}
\newcommand{\numreconedges}{\scalar{n}_e}
\newcommand{\reconvertex}{\vertex^r}
\newcommand{\reconedge}{\vector{e}^r}

\newcommand{\reconmeshcoarse}{\mesh_c}
\newcommand{\reconvertscoarse}{\vertices^c}
\newcommand{\reconvertexcoarse}{\vertex^c}
\newcommand{\reconnormalscoarse}{\normal^c}
\newcommand{\numreconvertscoarse}{\scalar{n}_v}

\newcommand{\samplegridcoarse}{\tensor{G}_c}
\newcommand{\griddimcoarse}{\scalar{d}_c}
\newcommand{\samplegrids}{\tensor{G}_r}
\newcommand{\griddim}{\scalar{d}_r}
\newcommand{\gridpoint}{\vector{p}}

\newcommand{\featureimage}{\tensor{F}}
\newcommand{\featurevector}{\vector{f}}
\newcommand{\featuredim}{\scalar{d}_f}
\newcommand{\featuremeancoarse}{\boldsymbol{\mu}_c}
\newcommand{\featurestdevcoarse}{\boldsymbol{\sigma}_c^2}
\newcommand{\featuremean}{\boldsymbol{\mu}_r}
\newcommand{\featurestdev}{\boldsymbol{\sigma}_r^2}
\newcommand{\featurevweights}{\eta}

\newcommand{\viewdir}{\vector{d}}
\newcommand{\featurevolumecoarse}{\tensor{Q}_c}
\newcommand{\featurevolume}{\tensor{Q}_r}

\newcommand{\volumefeature}{\vector{f}}

\newcommand{\probabilityvolume}{\tensor{P}}

\newcommand{\featurenet}{\misc{F}_{\text{img}}}
\newcommand{\transformer}{\misc{F}_{\text{loc}}}
\newcommand{\reconnetcoarse}{\misc{F}_{\text{rec}}}
\newcommand{\reconnet}{\misc{F}_{\text{ref}}}

\newcommand{\scale}{\vector{s}}
\newcommand{\translation}{\vector{t}}
\newcommand{\rotation}{\vector{r}}

\newcommand{\vtovweights}{\omega}
\newcommand{\edgeweights}{\gamma}

\section{Introduction}
\label{sec:introduction}

Capturing large datasets containing 3D heads of multiple people with varying facial expressions and head poses is a key enabler for modeling and synthesizing realistic head avatars. %
Typically, building such datasets is done in two steps: unstructured 3D scans are captured with a calibrated multi-view stereo (\ac{MVS}) system, followed by a non-rigid registration step to unify the mesh topology \cite{Egger2020_Survey}. 
This two-stage process has major drawbacks. 
\ac{MVS} reconstruction requires cameras with strongly overlapping views and the resulting scans frequently contain holes and noise.
Registering a template mesh to these scans typically involves manual parameter tuning to balance the trade-off between accurately fitting the scan's surface and being robust to scan artifacts.
Both stages are computationally expensive, each taking several minutes per scan.
For professional captures, both steps are augmented with manual clean-up to enhance the quality of the output meshes \cite{Alexander2009,Seymour2017_MeetMike}.
Such manual editing is infeasible for large-scale captures ($\gg 10$K scans). 

Instead, we advocate for a more practical setting that directly predicts 3D heads in dense  correspondence from calibrated multi-view images, effectively bypassing the \ac{MVS} step. 
We achieve this with \modelnamelong, which quickly ($\sim0.3$ seconds per head on a NVIDIA A100-SXM GPU) infers accurate 3D heads ($\sim~0.26$~mm median error) in correspondence, without manual user input.

While several methods exist that directly recover 3D faces in correspondence from calibrated multi-view images, they have high computational cost and require careful selection of optimization parameters per capture subject \cite{Bradley2010, Riviere2020, Beeler2011, Fyffe2017}. These remain major obstacles for large-scale data captures. 
A few learning-based methods directly regress parameters of a \ac{3DMM} \cite{Wu2019_MVF_Net} or iteratively refine \ac{3DMM} meshes from multi-view images \cite{Bai2020}. 
As shown by Li et al.~\cite{Li2021_ToFu}, this \ac{3DMM} dependency constrains the quality and expressiveness of these methods. 

The recent ToFu \cite{Li2021_ToFu} method goes beyond these \ac{3DMM}-based approaches with a volumetric feature sampling framework to infer face meshes from calibrated multi-view images. 
While demonstrating high-quality predictions, ToFu has several limitations. 
(a) The training is fully-supervised with paired data of multi-view images and high-quality registered meshes; creating such data requires extensive manual input.
(b) Only the face region is predicted; ears, neck, and the back of the head are manually completed in an additional fitting step. 
(c) Self-occlusions in scanner setups designed to capture the entire head result in mediocre predictions due to the na\"ive feature aggregation strategy that ignores the surface visibility.
(d) Only a small capture volume is supported and increasing the size of the capture volume to cover head movements reduces the accuracy. 

\modelname adapts ToFu's volumetric feature sampling framework but goes beyond it in several ways: 
(a) %
The training requires no manually curated data as we jointly optimize \modelname's weights and register the raw scans. 
Obtaining the clean, registered meshes required by ToFu is a key practical hurdle. 
\modelname learns from raw scans and is robust to their noise and missing data.
This is done by directly minimizing the point-to-surface distance between scans and predicted meshes. 
(b) At run time the entire head is inferred from images alone and includes the ears, neck, and back of the head. 
(c) The feature aggregation accounts for surface visibility. 
(d) A spatial transformer module \cite{Jaderberg2015_SpatialTransformer} localizes the head in the feature volume to only sample regions relevant for prediction, improving the accuracy. 

In summary, \modelname is the first framework to accurately capture the entire head from multi-view images at near interactive rates. 
During training, \modelname jointly learns to predict 3D heads from multi-view images, and registers unstructured scans. 
Once trained, it only requires calibrated camera input and it generalizes to diverse extreme expressions and head poses for subjects unseen during training (see Fig.~\ref{fig:teaser}). 
\modelname is trained and evaluated on a dynamic 3D head dataset of 95 subjects, each performing 28 facial motions, totalling about 600K 3D head meshes.
The registered dataset meshes, raw images, camera calibrations, trained model, and training code are publicly available.

\section{Related work}
\label{sec:related_work}

\paragraph{Scan registration:}
Registering unstructured 3D scans to a common mesh topology has been extensively studied over the last two decades since the work of Blanz and Vetter~\cite{BlanzVetter1999}.
For a comprehensive overview, we refer the reader to the survey of Egger et al.~\cite{Egger2020_Survey}. 
Most prior methods 
non-rigidly deform a template mesh \cite{BlanzVetter1999,Booth2016,Cao2014_FW,Yang2020_FaceScape,Li2009,Salazar2014,Li2017_FLAME,Passalis2011,Ji2021,Wu2018} with a generalization of the rigid \ac{ICP} algorithm \cite{BeslMcKay1992}.
Existing methods mostly register 3D scans of faces in a neutral expression \cite{BlanzVetter1999,Passalis2011,Booth2016}, a few static expressions \cite{Amberg2008,Salazar2014}, or faces in motion \cite{Abrevaya2018}.
Bolkart and Wuhrer~\cite{BolkartWuhrer2015_groupwise} and Zhang et al.~\cite{Zhang2016} jointly register static datasets of 3D faces while building a \ac{3DMM}.
Only a few methods consider more than just the face by registering scans of entire heads \cite{Li2017_FLAME,Yang2020_FaceScape,Dai2017,Dai2020}.
All these methods have in common that they are optimization-based, which makes them slow and sensitive to scans with noise or holes.
Dealing with this to obtain accurate results requires manual parameter tuning. 
Few learning-based methods exist to directly go from a scan to a registered mesh \cite{Liu2019,Bahri2021,Zheng2022_ImFace}.
While registering scans faster than previous optimization-based methods, these methods only register tightly cropped faces (i.e., no neck, ears, back of the head, etc.), they require facial landmarks, and  scan noise can negatively impact the reconstructed meshes. 
\modelname takes inspiration from these scan registration methods by minimizing similar objective functions (namely, point-to-surface distance and an edge-based surface regularization \cite{Li2017_FLAME}).
However, in contrast to these registration methods, we jointly register the training scans and train \modelname. 
Once trained, \modelname requires no scans as input, but, instead, directly infers 3D heads in correspondence from multi-view images.

\paragraph{Image-based reconstruction:}
Undoubtedly, monocular images or videos have drawn the largest focus as input to 3D face reconstruction methods and we refer to recent surveys for a thorough overview \cite{Morales2021,Zollhofer2018_Survey}. 
Most single-image-based 3D face reconstruction methods either fit some \ac{3DMM} to an image \cite{BlanzVetter1999,AldrianSmith2013,Bas2017fitting,Thies2016,Gecer2019,Wood2022,Ploumpis2020} in an analysis-by-synthesis fashion or regress the parameters of a \ac{3DMM} \cite{Tewari2017_MoFA,Tewari2018_250Hz,Deng2019,Kim2018,Feng2022_TRUST}.  
Reconstructing 3D from monocular images or videos is an ill-posed problem, as many 3D reconstructions give the same image when projected to 2D. 
To improve reconstruction accuracy, existing methods leverage multi-image constraints (e.g., same identity across images) \cite{AnhTran2017,Feng2021_DECA,Sanyal2019_RingNet,Danecek2022_EMOCA,Filntisis2022_SPECTRE} or multi-view constraints \cite{Shang2020} during training. 
Others use paired image-3D data for training, obtained by fitting a \ac{3DMM} to images \cite{Martyniuk2022_DAD3DHeads, Feng2018_PRNet, Guler2017, Jackson2017,Wei2019} or by sampling a \ac{3DMM} to generate synthetic data \cite{Dou2017, Richardson2016, Genova2018}. 
The state-of-the-art in single-image face reconstruction leverages a face recognition network trained on large amounts of image data, combined with supervised training from paired 2D-3D data obtained by registering 3D scans \cite{Zielonka2022_MICA}.
At test time, these methods reconstruct 3D faces from a single image without prior knowledge about the camera, image resolution, lighting, etc., making the problem ill-posed.  
The ambiguity between 3D face shape, camera, and distance to the camera  limits the metric accuracy of their 3D reconstructions \cite{BasSmith2019}. 

Instead, several methods focus on a more constrained scenario by reconstructing 3D faces from collections of images instead of just a single image. 
Such methods optimize the parameters of a parametric model \cite{Wood2022} or non-rigidly deform a template mesh \cite{Bai2020,Garrido2016,KemelmacherSeitz2011,Kemelmacher2013,Roth2015,Suwajanakorn2014} to fit multiple images of one subject. 
Learning-based methods independently regress \ac{3DMM} parameters from multiple images of the same subject and fuse the identity shape parameters to obtain coherent identity shape parameters per subject \cite{Ramon2019,Tewari2019_FML,DouKakadiaris2018,Wu2019_MVF_Net}. 
As camera intrinsics are unknown for these image collections, the ambiguity between (unknown) focal length and object scale 
means that the reconstructed faces are not metrically accurate.
Further, the approximate assumption about the camera (typically weak perspective projection) and the ambiguity of identity and expression-dependent shape result in reconstruction errors for each image.
This limits the overall reconstruction accuracy when integrating the erroneous results across images (see Li et al.~\cite{Li2021_ToFu} for comparisons). 
\modelname instead leverages camera calibration information to reconstruct metrically accurate 3D heads, and it is designed for a multi-view setup (i.e., multiple time-synchronized images per expression) to disambiguate identity and expression shape variations. 

Few methods directly reconstruct 3D faces from calibrated multi-view images. 
Among these, optimization-based approaches \cite{Bradley2010,Riviere2020,Beeler2011,Fyffe2017} to date achieve the most impressive results with fine-scale geometric details, but at the cost of being computationally slow, and requiring carefully tuned parameters per subject. 
As these methods are tailored towards specific custom capture setups, they cannot be directly applied to our off-the-shelf active stereo system.
The recent ToFu method~\cite{Li2021_ToFu} directly predicts 3D faces in correspondence, parameter-free at test time, and at near interactive rates.
However, ToFu is trained fully-supervised from high-quality registered 3D faces, which are difficult to obtain and limit the applicability of ToFu to new capture setups. 
Further, its na\"ive multi-view feature integration neglects surface visibility, limiting  generalization to $360^{\circ}$ capture setups as, e.g., features from the back of the head contribute to reconstructing face vertices and vice versa. 
Additionally, it assumes the capture volume tightly encapsulates the face, leading to mediocre results for larger capture volumes, required for covering moving heads. 
ToFu is therefore only able to capture tightly cropped faces, while the rest of the head is completed in a manual post-processing step. 
We build on top of ToFu's design but to overcome its limitations, (1) we train our model directly from raw scans, which makes it easier to adapt to new capture setups, (2) our feature integration considers surface properties and visibility, and (3) we localize the head in the feature volume with a spatial transformer. 
These changes improve the reconstruction quality and enable us to infer entire heads.

\paragraph{Multi-view stereo:}
\ac{MVS} capture systems are commonly used to reconstruct 3D faces \cite{Beeler2011,Ma2007,Ghosh2011,Goesele2006}. 
While reconstructing high-quality geometry, these methods are computationally expensive due to the pairwise feature matching across views. 
Recent learning-based methods \cite{Gu2020, Im2019_DPSNet, Kar2017, Sitzmann2019_DeepVoxels, Yao2018} reduce this computational cost but lose accuracy. 
For an overview of learning-based \ac{MVS} methods, see the survey of Wang et al.~\cite{Wang2021}.
All these methods have in common that the reconstructed geometry is unstructured, while our goal is to reconstruct meshes in correspondence. 
However, we use a commercial \ac{MVS} method to generate unstructured 3D head scans for our training data, and use these scans as a supervision signal. 
Once trained, \modelname directly predicts head meshes in  correspondence from multi-view images without requiring \ac{MVS} scans.

\section{Method}
\label{sec:method}

\begin{figure*}[ht]
	\centering
 	\includegraphics[width=0.99\linewidth]{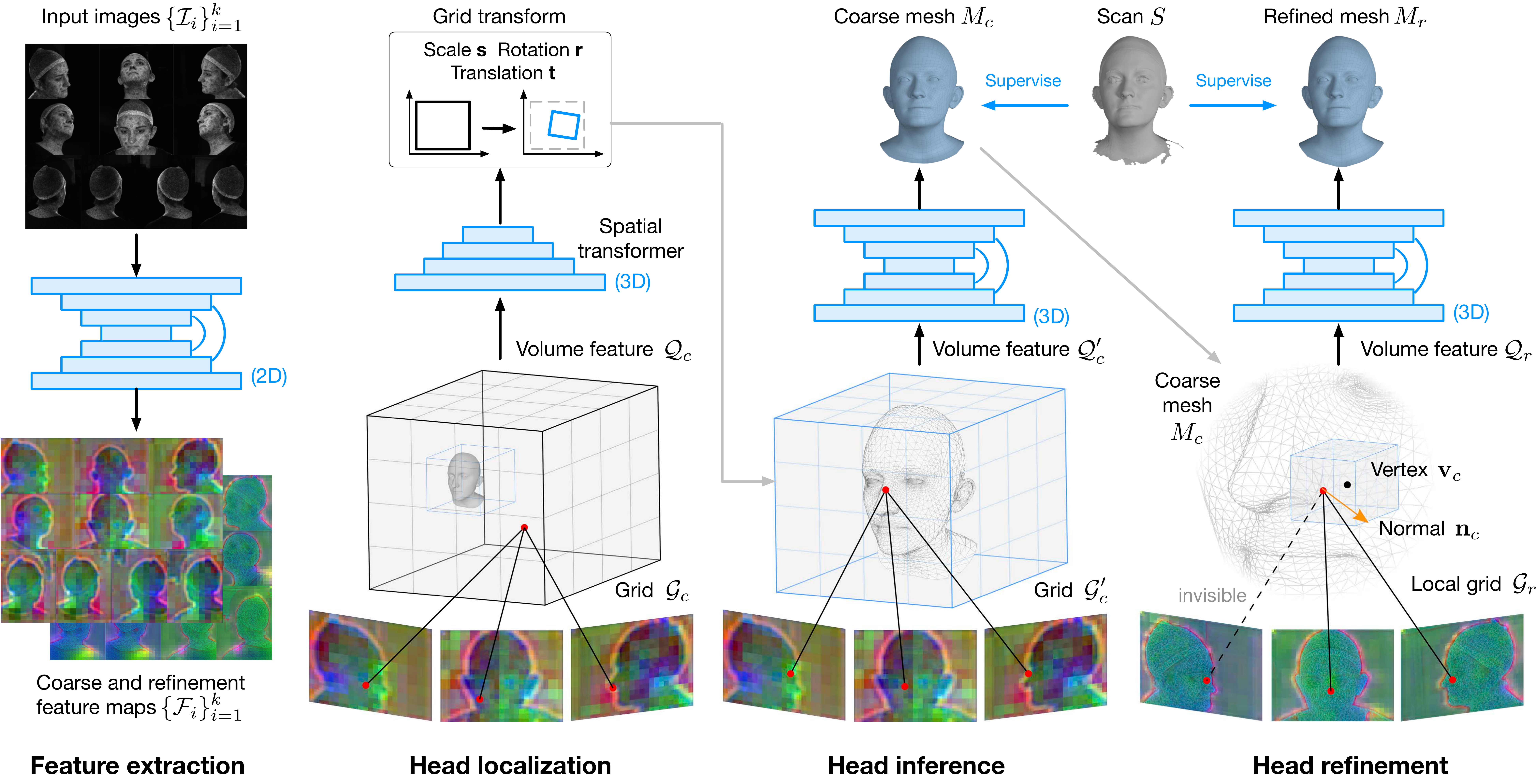}  %
	\caption{
	    \textbf{Overview}.
	    \modelname predicts a high-quality registered head mesh in two stages.
	    The coarse stage builds a feature volume from the feature maps, localizes the moving head in this volume with a spatial transformer, and then infers an intermediate 3D head mesh from the localized feature volume. 
	    The refinement stage updates each vertex location by sampling features locally, fusing these features view- and surface-aware, and predicting the updated vertex location from the local feature volume. 
	    During training, raw \ac{MVS} scans are used as supervision. 
	    Once trained, \modelname directly predicts head meshes from calibrated multi-view images without requiring scans as input. 
	}
	\label{fig:overview}
\end{figure*}

Given sets of images $\{\image_i \in \mathbb{R}^{\imagewidth \times \imageheight \times 3}\}_{i=1}^{\numviews}$ and camera calibrations $\{\calibration_i\}_{i=1}^{\numviews}$ (i.e., camera intrinsics, extrinsics, and radial distortion parameters) of a multi-view capture system with $\numviews$ views, \modelname infers a head mesh $\reconmesh :=(\reconverts, \reconfaces)$ with vertices $\reconverts \in \mathbb{R}^{\numreconverts \times 3}$ and faces $\reconfaces \in \mathbb{R}^{\numreconfaces \times 3}$.
Here, $\numreconverts$ and $\numreconfaces$ are the number of vertices and faces, respectively. 
Note that $\numreconverts$ and $\reconfaces$ are fixed, as all meshes output for different sets of input images are in dense  correspondence. 

As outlined in Fig.~\ref{fig:overview}, \modelname consists of two stages, a coarse head inference stage, which outputs an intermediate 3D head $\reconmeshcoarse:=(\reconvertscoarse, \reconfaces)$ with vertices $\reconvertscoarse \in \mathbb{R}^{\numreconverts \times 3}$, followed by a refinement stage that updates all vertex locations and outputs the final mesh $\reconmesh$. 
This two stage process allows us to leverage the surface properties of $\reconmeshcoarse$ (i.e., vertex position, visibility and normals) in the second stage for multi-view feature aggregation and vertex refinement. 

\begin{figure}[ht]
	\centering
	\includegraphics[width=0.99\linewidth]{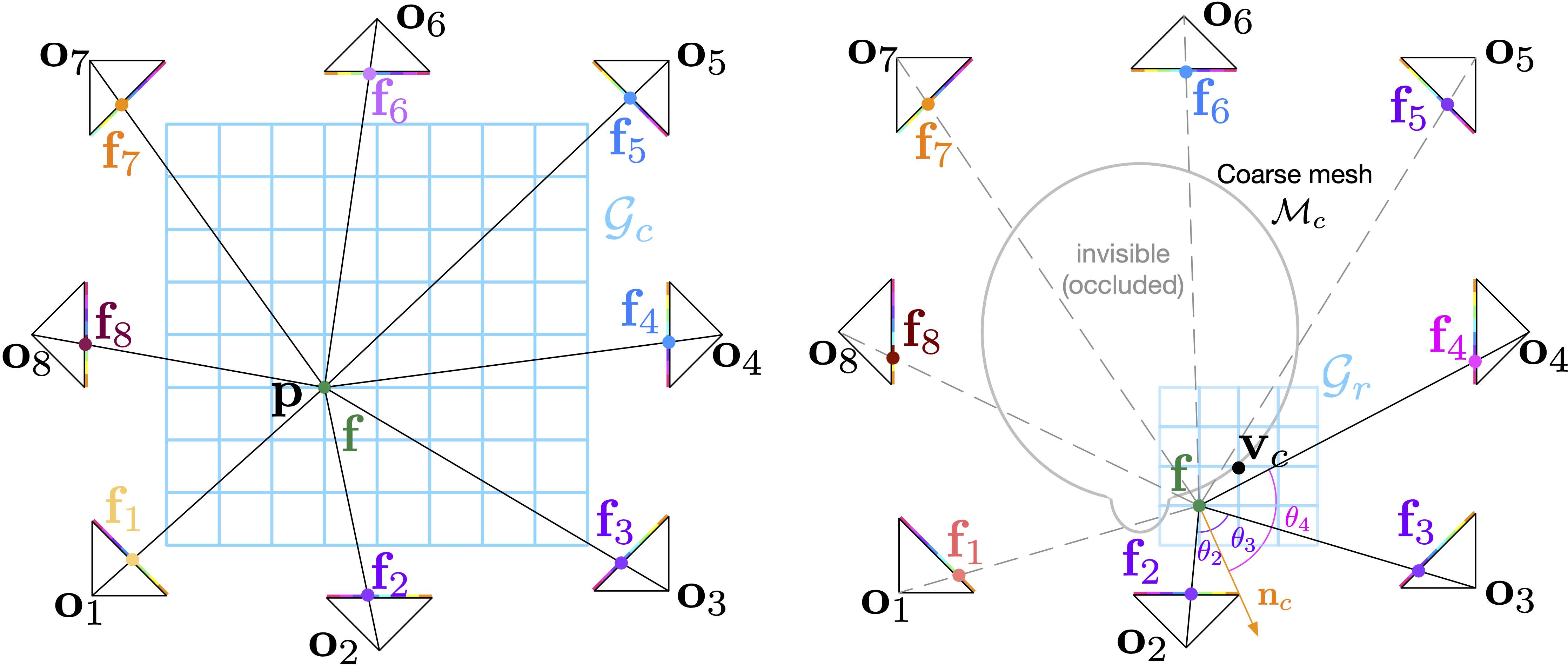} 
	\caption{\textbf{Volumetric feature sampling illustrated in 2D}. 
	While the coarse stage (left) directly fuses the feature vectors of all views, the refinement stage (right) performs a surface-aware feature fusion by leveraging the mesh estimated in the coarse stage. 
	Specifically, image features for occluded views (dashed lines), and views with large angle between the surface normal and viewing direction are assigned a low importance weight, while non-occluded views with low angle receive a higher weight.
	}
	\label{fig:volume_sample}
\end{figure}

\subsection{Coarse head prediction}

\paragraph{Feature extraction:}
Each image $\image_i$ is processed by a shared convolutional U-Net $\featurenet(\image_i) \rightarrow \featureimage_i \in \mathbb{R}^{\imagewidth \times \imageheight \times \featuredim}$ to extract 2D feature maps $\featureimage_i$ with $\featuredim$ feature channels. 

\paragraph{Volumetric feature sampling:}
Following Li et al.~\cite{Li2021_ToFu}, all 2D feature maps are then unprojected and fused into a feature cube (see Fig.~\ref{fig:volume_sample} left).
Specifically, for a point $\gridpoint \in \mathbb{R}^3$ and view $i$, a feature vector $\featurevector_i \in \mathbb{R}^{\featuredim}$ is obtained by perspective projection $\Pi(.)$ and bilinear sampling as $\featureimage_i(\Pi(\gridpoint, \calibration_i)) \rightarrow \featurevector_i$.
The view feature vectors are then fused by computing the mean $\featuremeancoarse = \frac{1}{\numviews} \sum_{i=1}^{\numviews} \featurevector_i$ and variance $\featurestdevcoarse = \frac{1}{\numviews} \sum_{i=1}^{\numviews} \featurevector_i^2-\featuremeancoarse^2$ (with a slight abuse of notation, where $\featurevector_i^2$ and $\featuremeancoarse^2$ are the element-wise squares) across all views, and concatenated to the feature vector $\volumefeature = (\featuremeancoarse \oplus \featurestdevcoarse) \in \mathbb{R}^{2\featuredim}$, where $\oplus$ denotes the vector concatenation.
This procedure is performed for every point in a 3D sampling grid $\samplegridcoarse \in \mathbb{R}^{\griddimcoarse \times \griddimcoarse \times \griddimcoarse \times 3}$ that covers the entire capture volume with $\griddimcoarse$ samples per dimension, to obtain a feature cube $\featurevolumecoarse \in \mathbb{R}^{\griddimcoarse \times \griddimcoarse \times \griddimcoarse \times 2\featuredim}$.

\paragraph{Head localization:}
While the head to be captured only occupies a small subspace of the capture volume, the feature cube $\featurevolumecoarse$ covers the entire space. 
This wastes learning capability due to the lower resolution of the feature cube in the relevant head region. 
Training a network to directly recover a 3D head from $\featurevolumecoarse$, as done in ToFu, therefore reduces the reconstruction accuracy. 
However, $\featurevolumecoarse$ contains information about the location and scale of the head.
We leverage this with a trainable spatial transformer $\transformer(\featurevolumecoarse) \rightarrow \{\scale, \rotation, \translation\}$ that predicts scale $\scale \in \mathbb{R}^3$, rotation $\rotation \in \mathbb{R}^6$~\cite{Zhou2019}, and translation $\translation \in \mathbb{R}^3$ to localize the head in $\featurevolumecoarse$.
Transforming every point of $\samplegridcoarse$ with $\{\scale, \rotation, \translation \}$ then gives a head localized feature grid $\samplegridcoarse^{'}$.
We then perform the volumetric feature sampling for $\samplegridcoarse^{'}$ to obtain a localized feature cube $\featurevolumecoarse^{'}$.

\paragraph{Head inference:}
The $\numreconvertscoarse$ vertices $\reconvertexcoarse_i$ of $\reconmeshcoarse$ are then predicted from $\featurevolumecoarse^{'}$ following Li et al.~\cite{Li2021_ToFu}.
For this, a convolutional 3D U-Net $\reconnetcoarse(\featurevolumecoarse^{'}) \rightarrow \probabilityvolume$ with a Softmax function applied to the output of the final layer's output across the first three dimensions predicts a probability volume $\probabilityvolume \in \mathbb{R}^{\griddimcoarse \times \griddimcoarse \times \griddimcoarse \times \numreconvertscoarse}$.
In $\probabilityvolume$, each of the $\numreconvertscoarse$ channels encodes the probability distribution $\probabilityvolume_i$ of the 3D location for one vertex $\reconvertexcoarse_i$. 
Each $\reconvertexcoarse_i$ is then computed as the element-wise product of $\probabilityvolume_i$ and the grid $\samplegridcoarse^{'}$, followed by a sum over the grid dimensions. 
Formally, $\reconvertexcoarse_i = \sum_{j,k,l=1}^{\griddimcoarse} (\probabilityvolume_i \odot \samplegridcoarse^{'})_{jkl}$, where $\odot$ denotes the Hadamard product, broadcasted across the fourth dimension of $\samplegridcoarse^{'}$ (i.e., the xyz-channels), and $(.)_{jkl}$ is the $jkl$-th tensor element.

\subsection{Geometry refinement}

\paragraph{Volumetric surface-aware feature fusion:}
Following the coarse feature sampling stage, for a point $\gridpoint \in \mathbb{R}^3$ and view $i$, a feature vector $\featurevector_i \in \mathbb{R}^{\featuredim}$ is obtained from the i-th view 2D feature map by perspective projection and bilinear sampling.
Different from the coarse stage, which uses one big sampling grid, the refinement stage defines, for every vertex $\reconvertexcoarse$, a small 3D sampling grid $\samplegrids \in \mathbb{R}^{\griddim \times \griddim \times \griddim \times 3}$, centered at $\reconvertexcoarse$.
For simplicity, we omit the vertex index and describe the feature sampling for one particular vertex.

ToFu \cite{Li2021_ToFu} na\"ively fuses all $\featurevector_i$ with equal importance across views, regardless of the visibility of point $\gridpoint$ in the $i$-th view. 
Instead, we use a surface-aware feature aggregation (see Fig.~\ref{fig:volume_sample} right) in form of a weighted mean 
\begin{equation}
	\featuremean = \frac{1}{\sum_{i=1}^{\numviews} \featurevweights_i} \sum \limits_{i=1}^{\numviews} \featurevweights_i \featurevector_i,
\end{equation}
and weighted variance
\begin{equation}
	\featurestdev = \frac{1}{\sum_{i=1}^{\numviews} \featurevweights_i} \sum \limits_{i=1}^{\numviews} \featurevweights_i \featurevector_i^2 - \featuremean^2,
\end{equation}
where $\featurevweights_i$ is a view- and surface-dependent weight that depends on the coarse mesh vertex $\reconvertexcoarse$, its corresponding vertex normal $\reconnormalscoarse$, and the camera location $\camcenter_i$ (in $\calibration_i$) of view $i$. 
The weight is computed as
\begin{equation}
	\featurevweights_i = 
	\text{Softplus}(
	    \vertexvisibility_i
	    \cdot \text{cos}~\theta_i
	),
\end{equation}
where $\vertexvisibility_i \in \left\{0,1\right\}$ is the visibility of $\reconvertexcoarse$ from the camera centered at $\camcenter_i$ (i.e., $0$ invisible, $1$ visible), and $\text{cos}~\theta_i = {<\reconnormalscoarse, \viewdir_i>}$ measures the angle between the surface normal $\reconnormalscoarse$ and the negative viewing direction $\viewdir_i$, where $\viewdir_i = (\camcenter_i - \gridpoint)/{\left\| \camcenter_i - \gridpoint \right\|}$.
The Softplus enforces positivity of the weight to get non-zero gradients w.r.t. features of all views. 

Finally, we compute the fused feature vector as $\volumefeature = (\featuremean \oplus \featurestdev \oplus \vertex) \in \mathbb{R}^{2\featuredim+3}$, where $\vertex \in \mathbb{R}^3$ is the vertex corresponding to $\reconvertexcoarse$ of a fixed template mesh, and $\oplus$ denotes the concatenation operation.
Adding $\vertex$ acts as identifier of the vertex to be processed to the refinement network. 
Performing the volumetric feature sampling for all points in $\samplegrids$ gives a local feature cube $\featurevolume$ for every vertex $\reconvertexcoarse$. 

\paragraph{Mesh refinement:}
Similar to the coarse face reconstruction, a convolutional 3D U-Net $\reconnet(\featurevolume) \rightarrow \probabilityvolume$, with a Softmax function applied to the final layer's output, predicts a probability volume $\probabilityvolume \in \mathbb{R}^{\griddim \times \griddim \times \griddim}$.
Note that different from the coarse stage, $\probabilityvolume$ encodes the probability distribution of the 3D location of one vertex $\reconvertexcoarse$. 
The final vertex location $\reconvertex$ is then reconstructed as $\sum_{j,k,l=1}^{\griddim} (\probabilityvolume \odot \samplegrids)_{jkl}$.

\subsection{Loss functions}

\paragraph{Surface distance:}
\modelname reconstructs meshes $\mesh \in \{ \reconmeshcoarse, \reconmesh \}$ (i.e., either coarse mesh $\reconmeshcoarse$ or refined mesh $\reconmesh$) in correspondence, which must closely resemble the raw training scans $\scan$.
To enforce this, the training of \modelname minimizes the point-to-surface distance, given as
\begin{equation}
    E_{\text{s2m}} = \lambda_{\text{s2m}} \frac{1}{\left| \scan \right|} \sum \limits_{\scanpoint \in \scan} \gmo \left( \min_{\meshpoint \in \mesh} \left\| \scanpoint - \meshpoint \right\|_2^2 \right),
    \label{eq:s2m}
\end{equation}
which measures the distances between points $\scanpoint \in \mathbb{R}^3$ on the surface of $\scan$ and their closest points $\meshpoint \in \mathbb{R}^3$ on the surface of $\mesh$.
The Geman-McClure~\cite{GemanMcClure1987} robust penalty function $\gmo(.)$ provides robustness to noise and outliers in the scans, the weight $\lambda_{\text{s2m}} \in \mathbb{R}^{+}_{0}$ controls the impact of the loss.

\paragraph{Surface regularization:}
Directly optimizing Eq.~\ref{eq:s2m} results in poor registrations with overlapping and self-intersecting faces. 
To prevent this, we add a relative edge regularization
\begin{equation}
    E_{\text{reg}} = \lambda_{\text{reg}} \frac{1}{\numreconedges} \sum \limits_{i=1}^{\numreconedges} \edgeweights_i \left\| \meshedge_i - \registrationedge_i \right\|_2^2,
\end{equation}
which penalizes the difference between the 3D edge vectors $\meshedge$ of $\mesh \in \{ \reconmeshcoarse, \reconmesh \}$ and the corresponding edge vectors $\registrationedge$ of the reference registration $\registration$. 
To account for varying scan quality in different face regions (e.g., eyes, lips, and hair regions are often more noisy), pre-defined edge weights $\edgeweights \in \mathbb{R}^{+}_{0}$ control the amount of regularization per face region, $\lambda_{\text{reg}} \in \mathbb{R}^{+}_{0}$ weights the overall regularization. 

\paragraph{Registration error:}
The distance to the registration $\registration$ is minimized as vertex-to-vertex distance, specifically as
\begin{equation}
    E_{\text{v2v}} = \frac{1}{\numreconverts} \sum \limits_{i=1}^{\numreconverts} \vtovweights_i \left\| \meshvertex_i - \registrationvertex_i \right\|_2^2,
    \label{eq:v2v}
\end{equation}
where $\meshvertex$ and $\registrationvertex$ are vertices of $\mesh \in \{ \reconmeshcoarse, \reconmesh \}$ and $\registration$, respectively. 
The binary weights $\vtovweights_i \in \left\{0,1\right\}$ control the impact of individual vertices. 
Note that $E_{\text{v2v}}$ is only used to pre-train the coarse reconstruction stage for the full head (i.e., $\vtovweights_i = 1~\forall i$). 
Afterwards, %
it only serves to regularize the eyeball location, with $\vtovweights_i = 1$ for all eyeball vertices, and $\vtovweights_i = 0$ otherwise. 

\section{Implementation details}
\label{sec:implementation}

\paragraph{Capture setup:}
We use a multi-camera active stereo capture system (3dMD LLC, Atlanta) with eight pairs of gray-scale stereo cameras, and eight color cameras.
The capture system is calibrated, providing parameters for camera extrinsics, intrinsics, and radial distortions. 
Each camera captures images of resolution $1600\times1200$ at 60 fps, where the color cameras are time synchronized with light LED panels, and the stereo cameras are synchronized with random speckle pattern projectors. 
The system uses a \ac{MVS} method to reconstruct unstructured 3D scans at 60 fps, each with about 110K vertices, from the stereo images.

\paragraph{Data capture:}
We collect a multi-view 3D head dataset, referred to as FaMoS (Facial Motion across Subjects), from 95 subjects with our capture setup. 
Each subject performs 28 motion sequences, containing six prototypical expressions (i.e., Anger, Disgust, Fear, Happiness, Sadness, and Surprise), two head rotations (left/right and up/down), and diverse facial motions, including extreme and asymmetric expressions. 
All subjects wear a hair net. %
In total, FaMoS contains around 600K frames (i.e., $\sim225$ frames per sequence) of calibrated multi-view images and corresponding 3D head scans. 
See the Sup. Mat. for additional details.

\paragraph{Training data:}
We train \modelname on data of 78 FaMoS subjects (70 subjects for training, 8 for validation). 
For training, we randomly sample 40 frames per expression sequence, and 120 frames per head rotation sequence (88K frames in total).
For validation, we sample 5 frames per sequence (1,118 frames in total).
For each frame, we compute a reference registration in FLAME mesh topology with the fully-automatic registration method of Li et al.~\cite{Li2017_FLAME}.

\paragraph{Test data:}
We qualitatively and quantitatively evaluate \modelname on all 28 sequences of 15 FaMoS subjects, disjoint from the training subjects. 
For each sequence, we randomly sample 20 frames; in total $8,350$ frames.

\paragraph{Implementation details:}
\modelname is implemented in PyTorch~\cite{Paszke2019PyTorchAI}, and optimized with AdamW~\cite{Loshchilov2019_AdamW} with a learning rate of $1e-4$ for the parameters of the head localization networks, and $1e-3$ for all other parameters.
The volumetric feature sampling is adapted from ToFu~\cite{Li2021_ToFu}.
The differentiable distance between scans and reconstructed mesh is based on Kaolin~\cite{KaolinLibrary}, random points are sampled in each scan's surface during training with Trimesh~\cite{Trimesh}.

\modelname takes the 16 gray-scale stereo images as input, as they are time synchronized with the 3D scans used as training supervision, and it outputs meshes in FLAME mesh topology \cite{Li2017_FLAME} with $\numreconverts=5023$ vertices and $\numreconfaces=9976$ faces. 
The feature network $\featurenet$ is a U-Net \cite{Ronneberger2015_UNet} with a ResNet34 \cite{He2016_ResNet} backbone.
The head localization network  $\transformer$ consists of two 3D convolutional layers and a fully-connected layer, each with ReLU activations, followed by a final linear layer. 
The reconstruction networks $\reconnetcoarse$ and $\reconnet$ are 3D U-Nets \cite{Iskakov2019}, with five and three down- and upsampling layers, respectively.
See the Sup.~Mat.~for details on the model architecture and computational requirements.

\paragraph{Parameter settings:}
First, the coarse reconstruction stage is trained for 300K iterations on the reference registrations with $\vtovweights_i = 1$ for all vertices, and without surface distance or regularization ($\lambda_{\text{s2m}} = \lambda_{\text{reg}} = 0$).
Then, we train the coarse reconstruction stage with surface distance weight $\lambda_{\text{s2m}} = 10.0$ and surface regularization weight $\lambda_{\text{reg}} = 1.0$ for 300K %
iterations. 
Directly optimizing  $E_{\text{s2m}}$ and $E_{\text{reg}}$ can result in dislocated eyeballs during training, as the eyeballs are parts separated from the head, and $E_{\text{reg}}$ is translation invariant. 
To prevent this, we regularize the eyeball locations with $\vtovweights_i = 1$ for all eyeball vertices, for all other vertices, we set $\vtovweights_i = 0$. 
We train the refinement stage for 150K iterations with $\lambda_{\text{s2m}} = 10.0$ and $\lambda_{\text{reg}} = 0.3$.  %
Following ToFu, the dimensions of the feature grids are $\griddimcoarse=32$ (coarse stage) and $\griddim = 8$ (refinement stage), and the number of features is $\featuredim=8$. 
To reduce the memory consumption, we downsample the input images by a factor of $8$ for the coarse training (i.e., $\imagewidth=200, \imageheight=150$), and by a factor of $4$ for the refinement step (i.e., $\imagewidth=400, \imageheight=300$).
We use a batch size of $2$ for the entire training.

\section{Evaluation}
\label{sec:evaluation}

\newcommand{\qualcolmargin}{0.0005\linewidth}

\newcommand{\qualimgsize}{0.17\linewidth}
\newcommand{\qualmeshsize}{0.12\linewidth}
\newcommand{\qualimgcropleft}{70}
\newcommand{\qualimgcroplower}{70}
\newcommand{\qualimgcropright}{120}
\newcommand{\qualimgcropupper}{40}

\begin{figure*}[ht]
    \centering
    \begin{tabular}{@{}c@{\hskip \qualcolmargin}c@{\hskip \qualcolmargin}c@{\hskip \qualcolmargin}c@{\hskip \qualcolmargin}r@{}}     
        \includegraphics[width=\qualimgsize, clip]{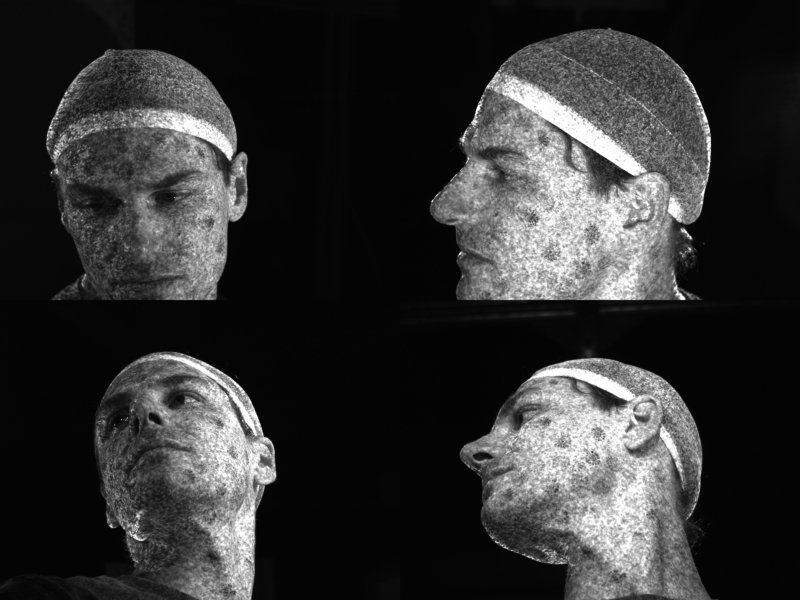}          
        &
        \includegraphics[width=\qualmeshsize, clip, trim={\qualimgcropleft} {\qualimgcroplower} {\qualimgcropright} {\qualimgcropupper}]{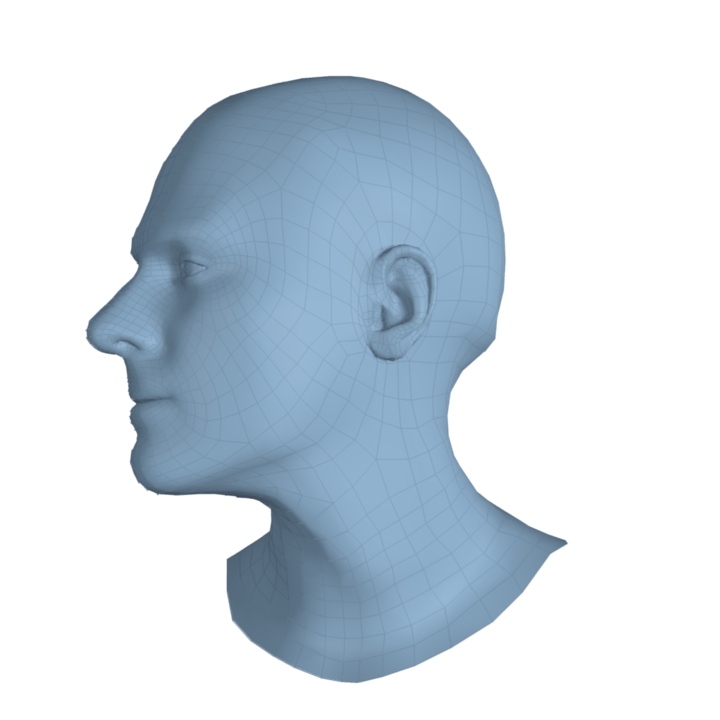}
        \includegraphics[width=\qualmeshsize, clip, trim={\qualimgcropleft} {\qualimgcroplower} {\qualimgcropright} {\qualimgcropupper}]{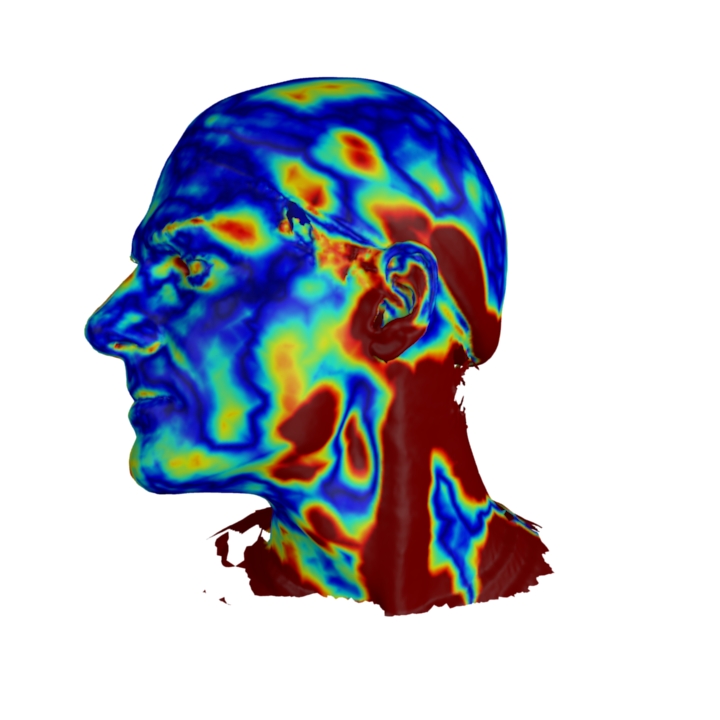}  
        & 
        \includegraphics[width=\qualmeshsize, clip, trim={\qualimgcropleft} {\qualimgcroplower} {\qualimgcropright} {\qualimgcropupper}]{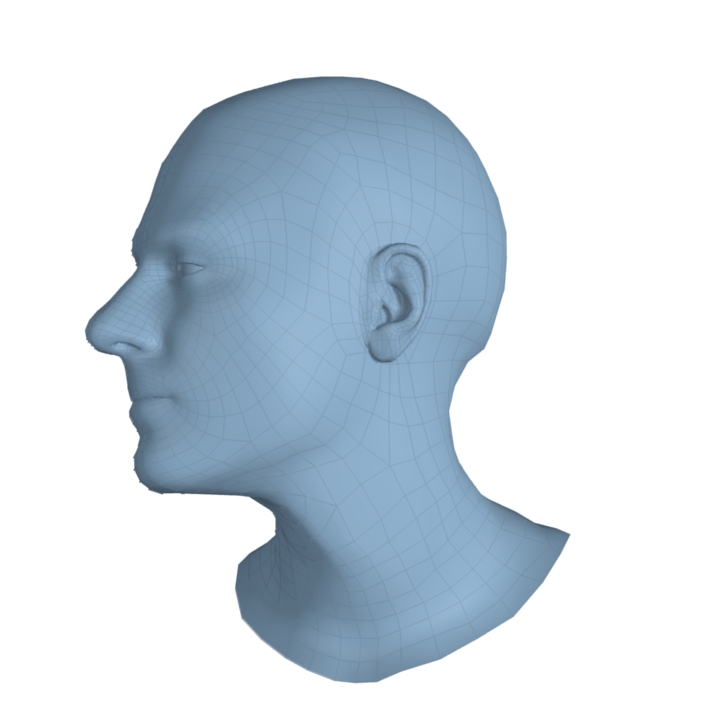}
        \includegraphics[width=\qualmeshsize, clip, trim={\qualimgcropleft} {\qualimgcroplower} {\qualimgcropright} {\qualimgcropupper}]{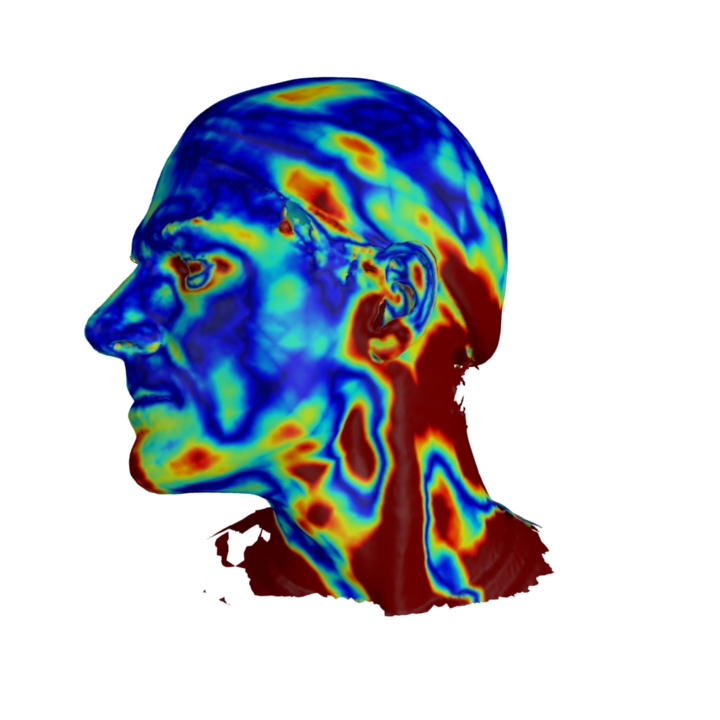}      
        &
        \includegraphics[width=\qualmeshsize, clip, trim={\qualimgcropleft} {\qualimgcroplower} {\qualimgcropright} {\qualimgcropupper}]{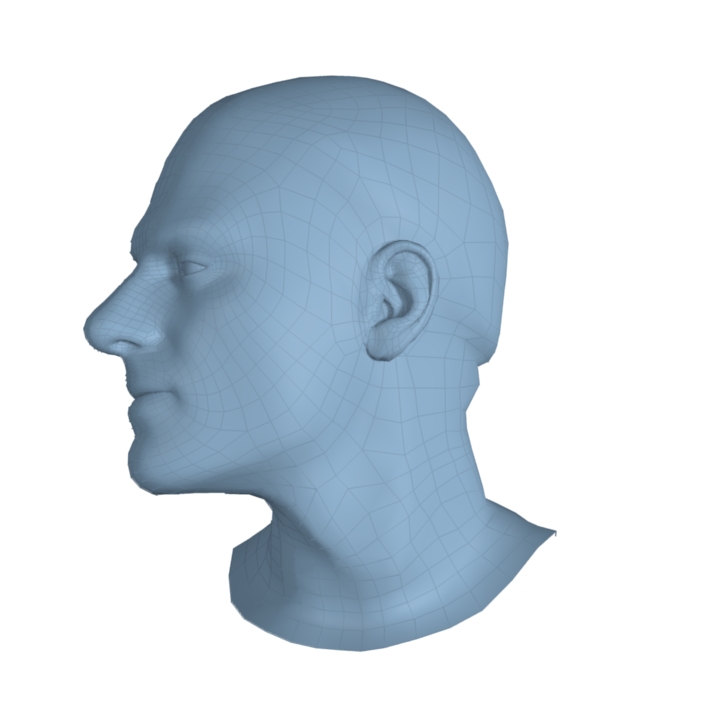}
        \includegraphics[width=\qualmeshsize, clip, trim={\qualimgcropleft} {\qualimgcroplower} {\qualimgcropright} {\qualimgcropupper}]{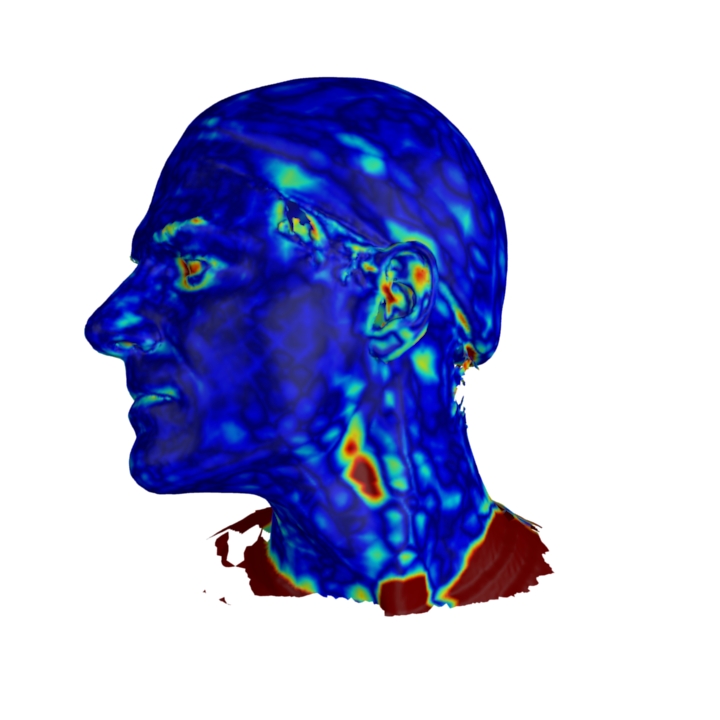}
        & %
        \includegraphics[width=0.05\linewidth]{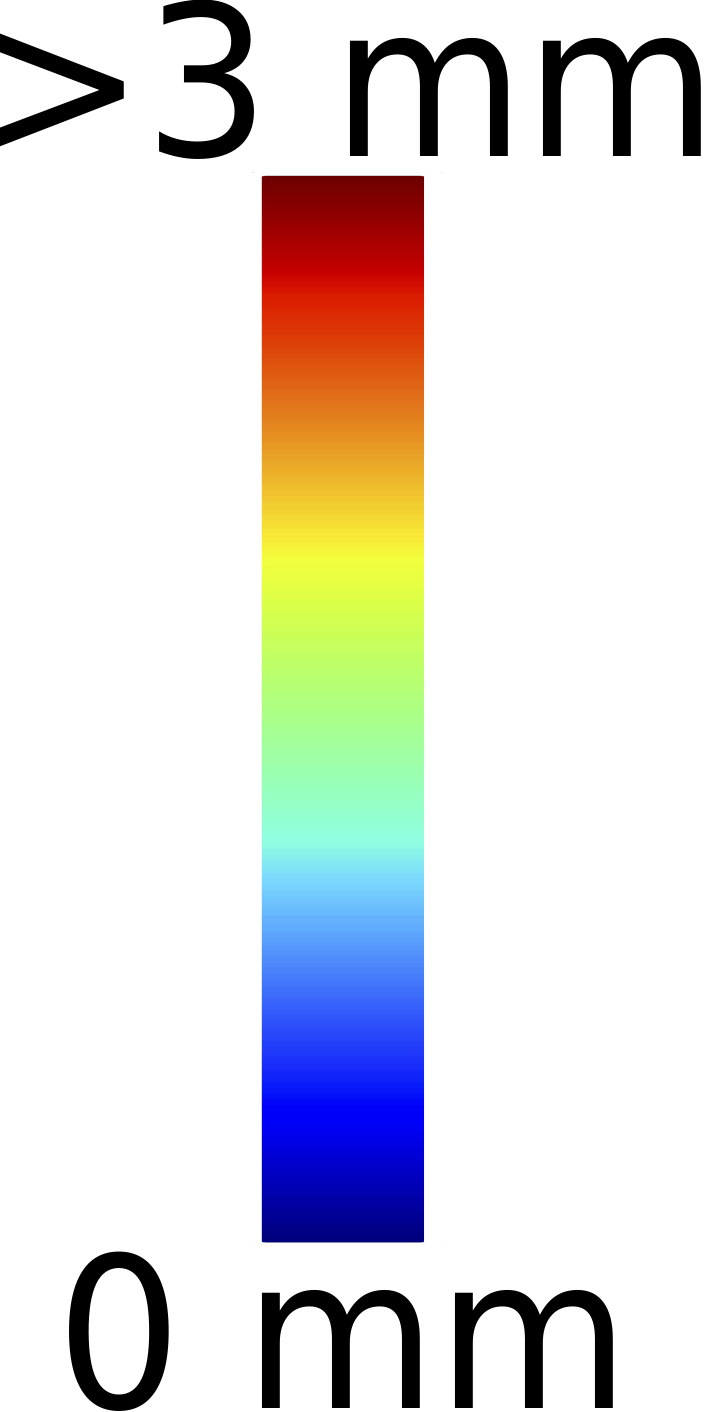}
        \\
        \includegraphics[width=\qualimgsize, clip]{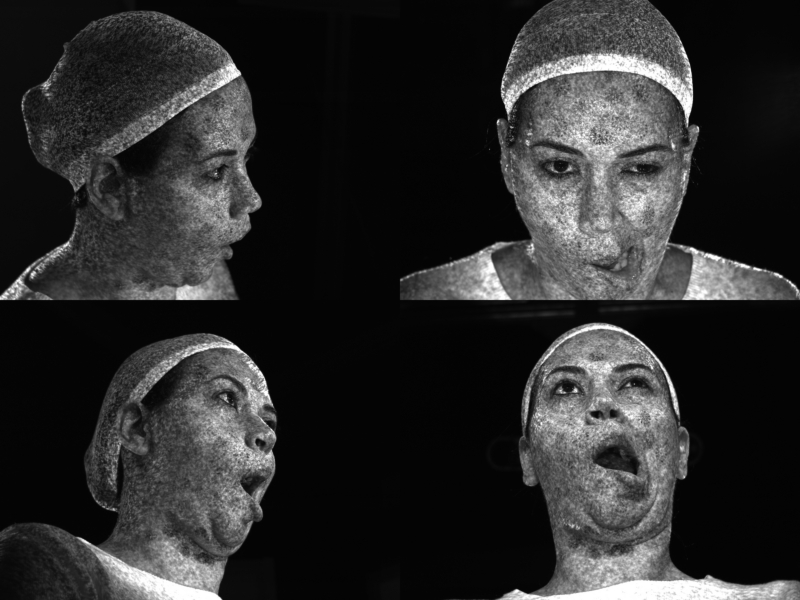}         
        &
        \includegraphics[width=\qualmeshsize, clip, trim={\qualimgcropleft} {\qualimgcroplower} {\qualimgcropright} {\qualimgcropupper}]{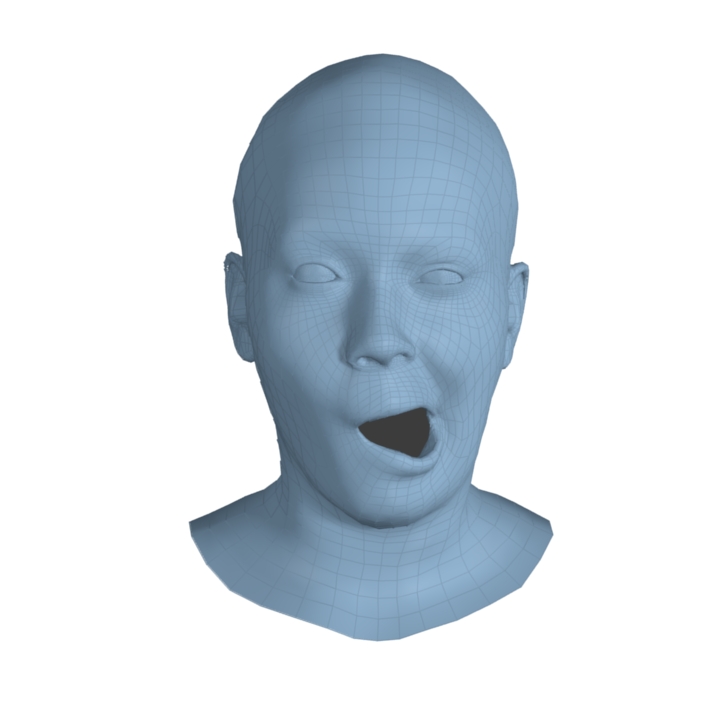}
        \includegraphics[width=\qualmeshsize, clip, trim={\qualimgcropleft} {\qualimgcroplower} {\qualimgcropright} {\qualimgcropupper}]{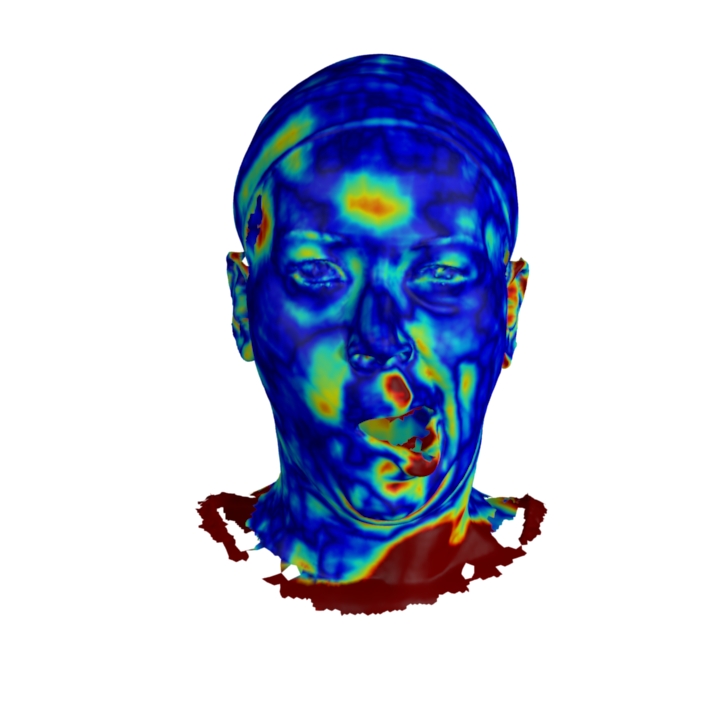}  
        & 
        \includegraphics[width=\qualmeshsize, clip, trim={\qualimgcropleft} {\qualimgcroplower} {\qualimgcropright} {\qualimgcropupper}]{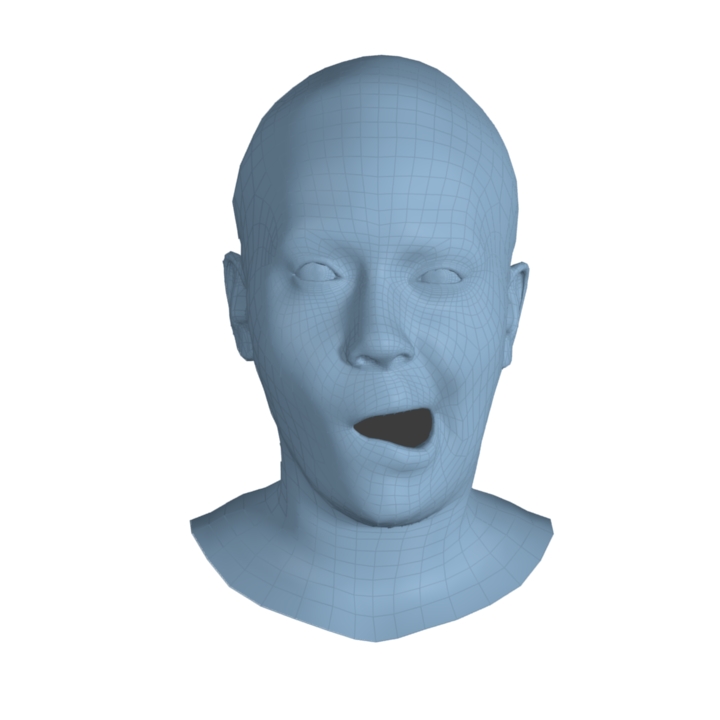}
        \includegraphics[width=\qualmeshsize, clip, trim={\qualimgcropleft} {\qualimgcroplower} {\qualimgcropright} {\qualimgcropupper}]{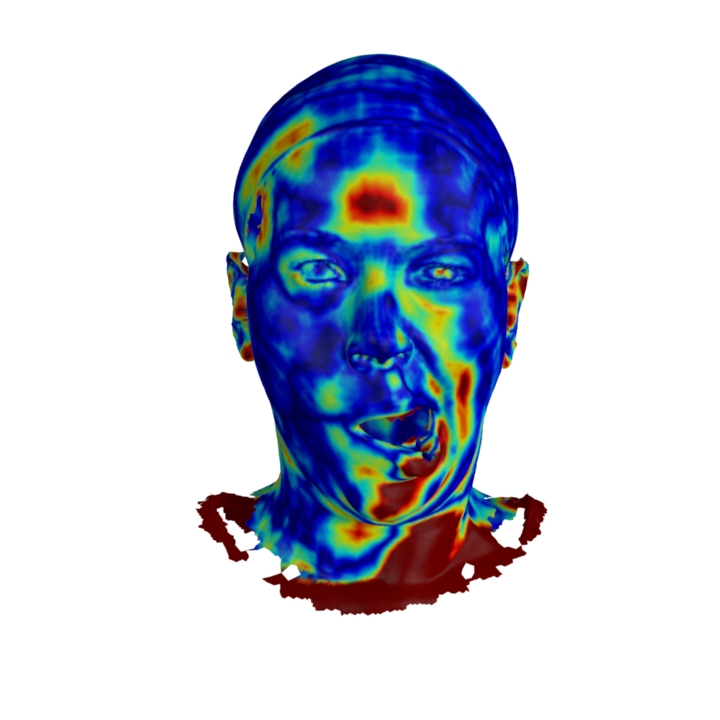}      
        &
        \includegraphics[width=\qualmeshsize, clip, trim={\qualimgcropleft} {\qualimgcroplower} {\qualimgcropright} {\qualimgcropupper}]{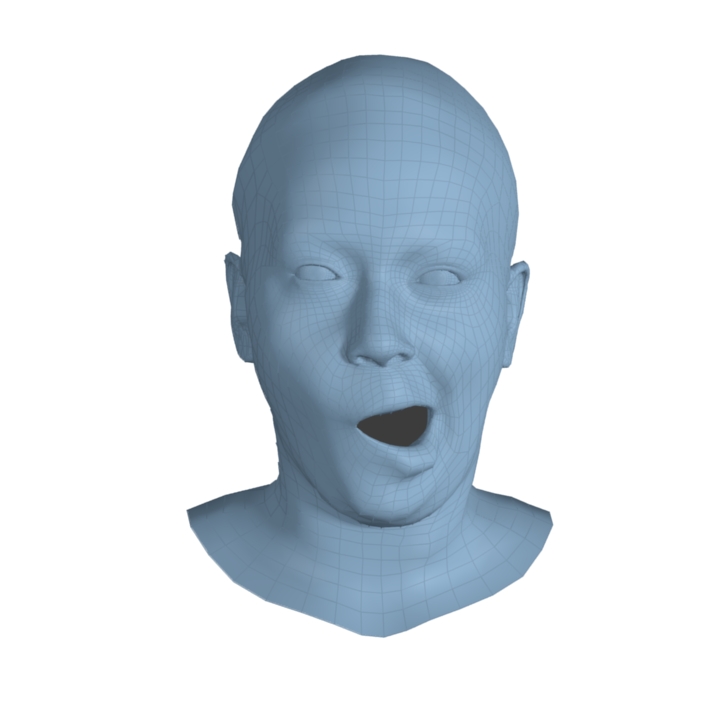}
        \includegraphics[width=\qualmeshsize, clip, trim={\qualimgcropleft} {\qualimgcroplower} {\qualimgcropright} {\qualimgcropupper}]{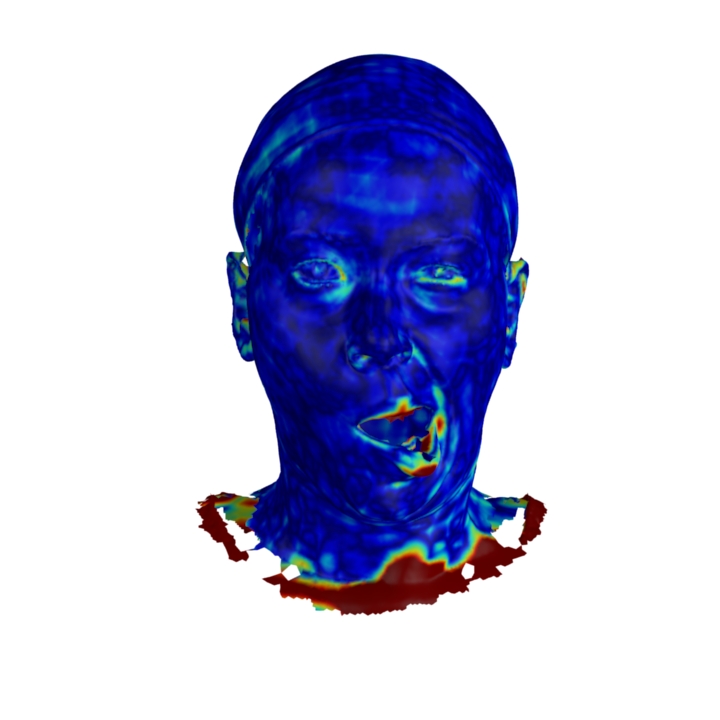}    
        & %
        \includegraphics[width=0.05\linewidth]{images/qualitative_results/color_map_3mm.pdf}
        \\                     
         \includegraphics[width=\qualimgsize, clip]{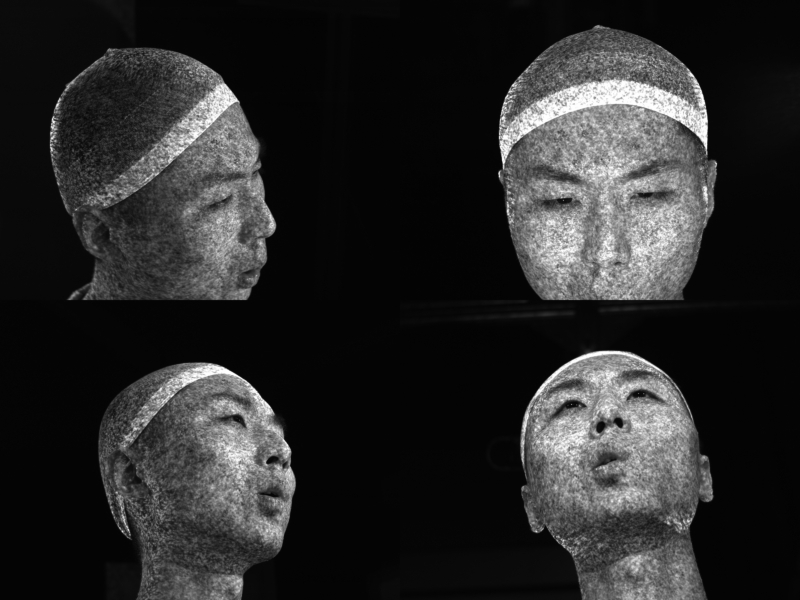}           
         &
         \includegraphics[width=\qualmeshsize, clip, trim={\qualimgcropleft} {\qualimgcroplower} {\qualimgcropright} {\qualimgcropupper}]{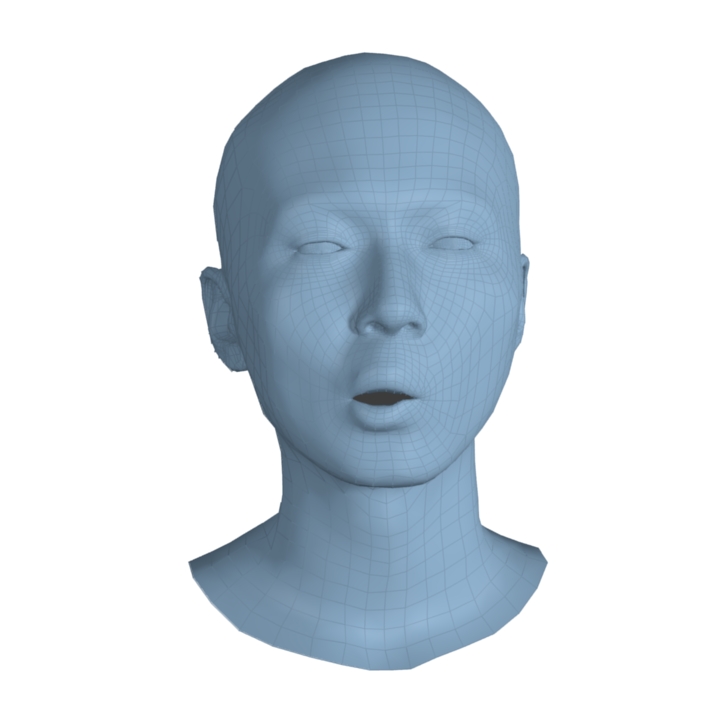}
        \includegraphics[width=\qualmeshsize, clip, trim={\qualimgcropleft} {\qualimgcroplower} {\qualimgcropright} {\qualimgcropupper}]{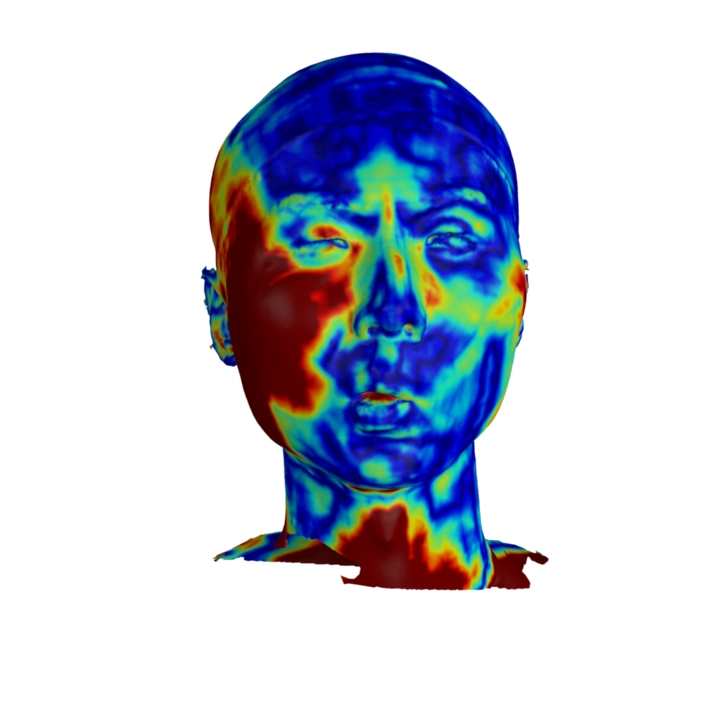}  
         & 
        \includegraphics[width=\qualmeshsize, clip, trim={\qualimgcropleft} {\qualimgcroplower} {\qualimgcropright} {\qualimgcropupper}]{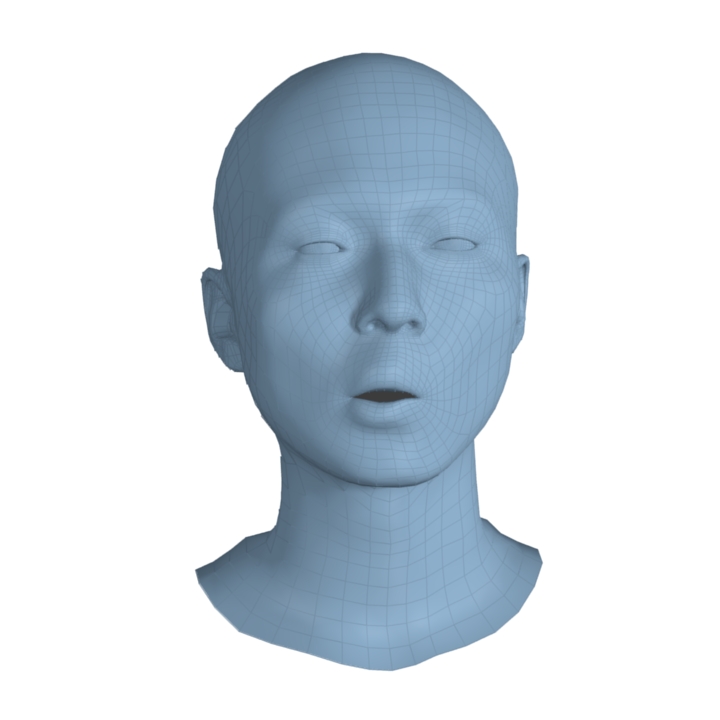}
        \includegraphics[width=\qualmeshsize, clip, trim={\qualimgcropleft} {\qualimgcroplower} {\qualimgcropright} {\qualimgcropupper}]{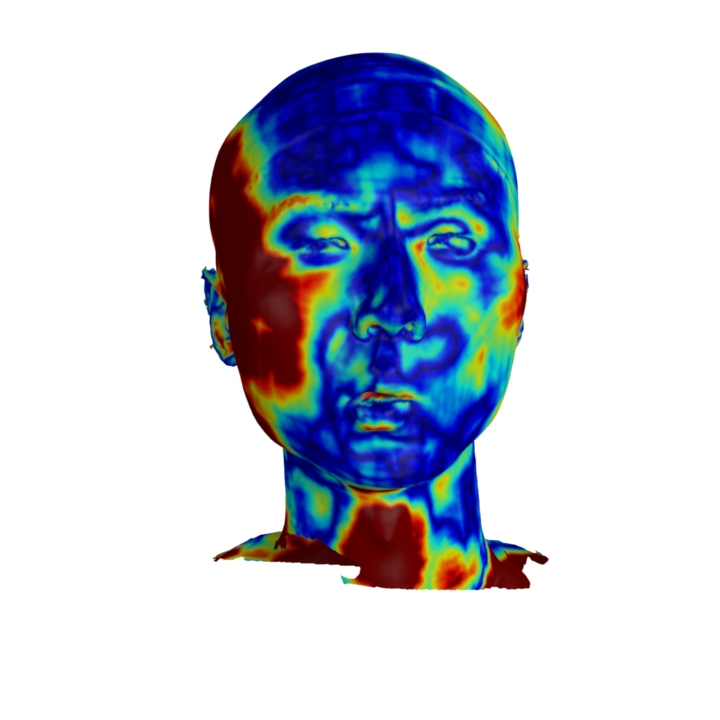}      
        &
        \includegraphics[width=\qualmeshsize, clip, trim={\qualimgcropleft} {\qualimgcroplower} {\qualimgcropright} {\qualimgcropupper}]{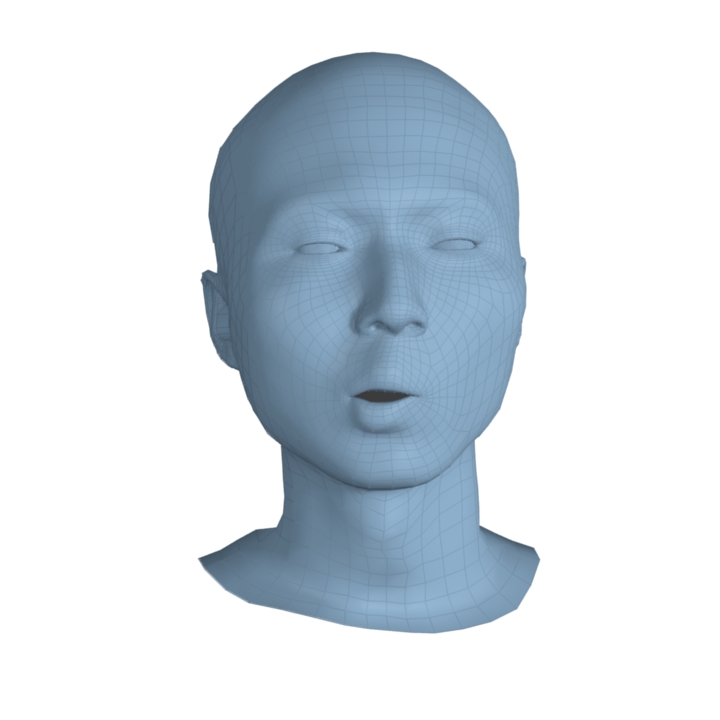}
        \includegraphics[width=\qualmeshsize, clip, trim={\qualimgcropleft} {\qualimgcroplower} {\qualimgcropright} {\qualimgcropupper}]{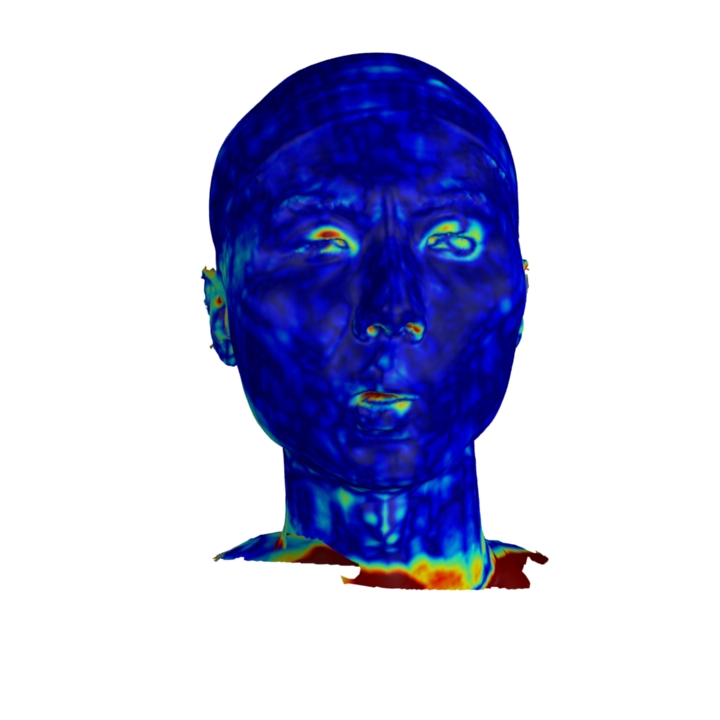}    
        & %
        \includegraphics[width=0.05\linewidth]{images/qualitative_results/color_map_3mm.pdf}
        \\          
        Input (4 of 16 views) & ToFu & ToFu+ & Ours
    \end{tabular}
	\caption{\textbf{Qualitative evaluation}.
	Comparison to ToFu \cite{Li2021_ToFu} and ToFu+ on FaMoS test samples with varying expressions and head poses for subjects not present during training. 
	For each method, we show the predicted mesh (left) and the color coded point-to-surface distance (right) between the reference scan and the predicted mesh as a heatmap on the scan's surface (red means $\geq$ 3 millimeter). 
	}
    \label{fig:qualitative_eval}
\end{figure*}

\begin{table*}[ht]
    \centering
    \resizebox{0.9\textwidth}{!}{

\begin{tabular}{l|ccc|ccc|ccc|ccc}
    \toprule
    & \multicolumn{3}{c|}{\bf Complete head} & \multicolumn{3}{c|}{\bf Face} & \multicolumn{3}{c|}{\bf Scalp} & \multicolumn{3}{c}{\bf Neck}  \\    
    Method & Median $\downarrow$ & Mean $\downarrow$ & Std $\downarrow$ & Median $\downarrow$ & Mean $\downarrow$ & Std $\downarrow$ & Median $\downarrow$ & Mean $\downarrow$ & Std $\downarrow$ & Median $\downarrow$ & Mean $\downarrow$ & Std $\downarrow$     \\
    \midrule
    3DMM regressor & 9.42 & 12.06 & 10.11 & 8.74 & 10.38 & 7.91 & 9.19 & 11.80 & 9.74 & 13.02 & 16.45 & 13.29 \\ %
    ToFu      & 0.72 & 1.44 & 2.72 & 0.63 & 0.93 & 1.50 & 0.62 & 1.46 & 3.27 & 1.30 & 2.39 & 3.24 \\ %
    ToFu+     & 0.82 & 1.59 & 2.84 & 0.68 & 1.00 & 1.52 & 0.73 & 1.59 & 3.16 & 1.50 & 2.77 & 3.82 \\ %
    Ours      & \textbf{0.26} & \textbf{0.51} & \textbf{1.22} & \textbf{0.21} & \textbf{0.34} & \textbf{1.22} & \textbf{0.26} & \textbf{0.41} & \textbf{0.66} & \textbf{0.38} & \textbf{0.95} & \textbf{1.91}  \\ %
    \bottomrule
\end{tabular}

}
	\caption{\textbf{Quantitative evaluation}. 
	Reconstruction error for varied head regions (FaMoS test set).
	We compare to a 3DMM regressor, ToFu \cite{Li2021_ToFu} and ToFu without hierarchical architecture (ToFu+), all trained to predict entire heads on the \modelname training data.
	Errors in mm.}
	\label{tab:qualitative_eval}
\end{table*}

\newcommand{\ablcolmargin}{0.0007\linewidth}

\newcommand{\ablimgsize}{0.17\linewidth}
\newcommand{\ablmeshsize}{0.11\linewidth}
\newcommand{\ablimgcropleft}{80}
\newcommand{\ablimgcroplower}{70}
\newcommand{\ablimgcropright}{120}
\newcommand{\ablimgcropupper}{40}

\begin{figure*}[ht]
    \centering
    \begin{tabular}{@{}c@{\hskip \ablcolmargin}c@{\hskip \ablcolmargin}c@{\hskip \ablcolmargin}c@{\hskip \ablcolmargin}r@{}}
        \includegraphics[width=\ablimgsize, clip]{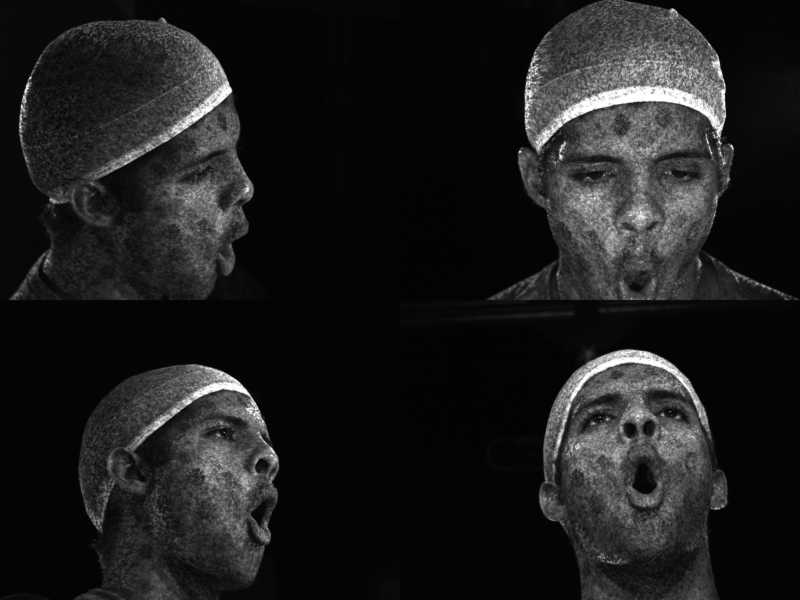}          
        & %
        \includegraphics[width=\ablmeshsize, clip, trim={\ablimgcropleft} {\ablimgcroplower} {\ablimgcropright} {\ablimgcropupper}]{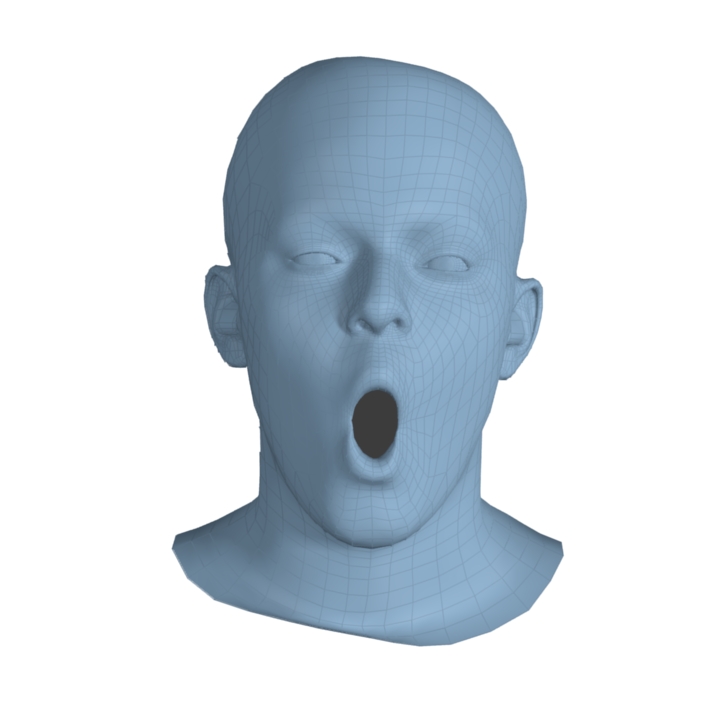}
        \includegraphics[width=\ablmeshsize, clip, trim={\ablimgcropleft} {\ablimgcroplower} {\ablimgcropright} {\ablimgcropupper}]{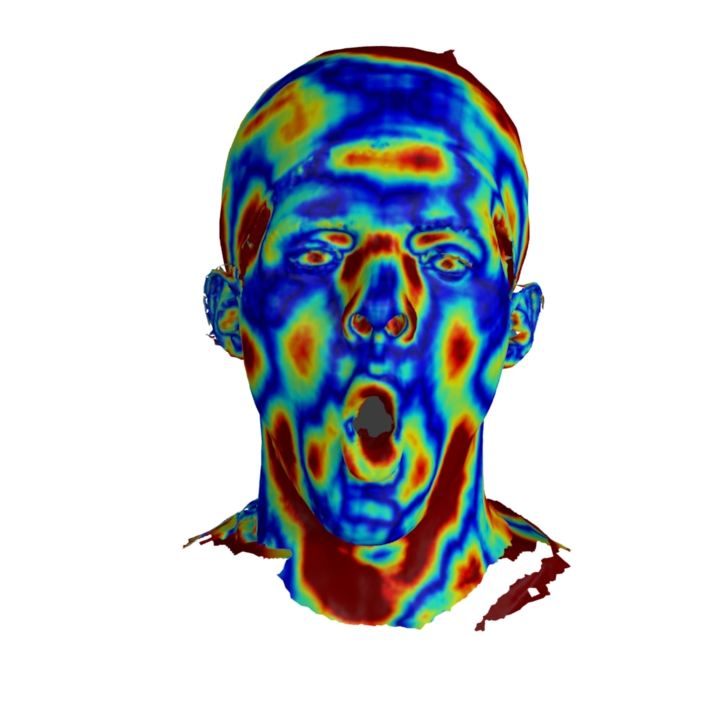}
        & %
        \includegraphics[width=\ablmeshsize, clip, trim={\ablimgcropleft} {\ablimgcroplower} {\ablimgcropright} {\ablimgcropupper}]{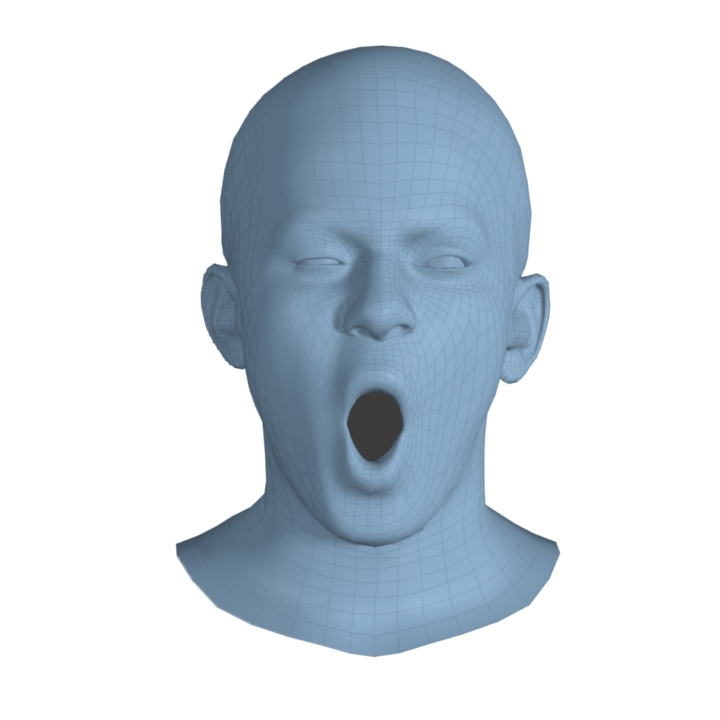}
        \includegraphics[width=\ablmeshsize, clip, trim={\ablimgcropleft} {\ablimgcroplower} {\ablimgcropright} {\ablimgcropupper}]{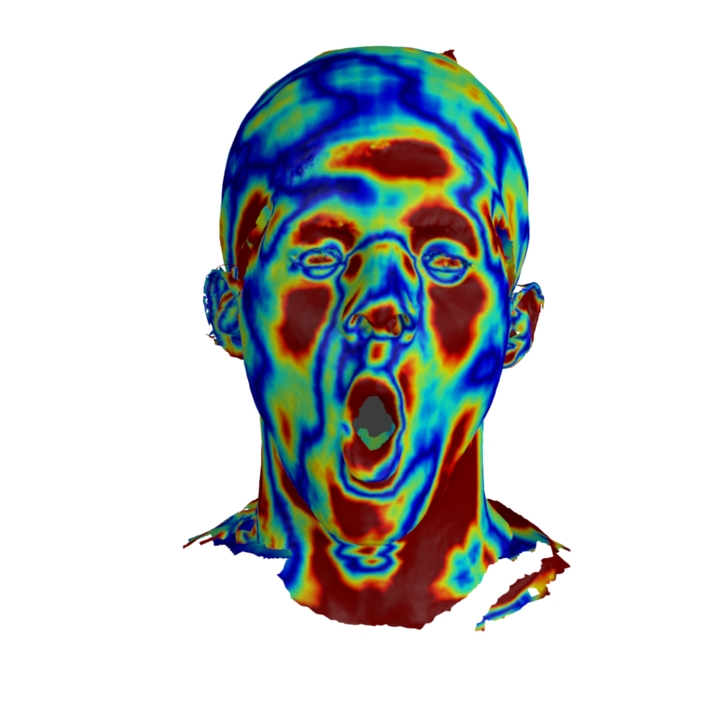}
        & %
        \includegraphics[width=\ablmeshsize, clip, trim={\ablimgcropleft} {\ablimgcroplower} {\ablimgcropright} {\ablimgcropupper}]{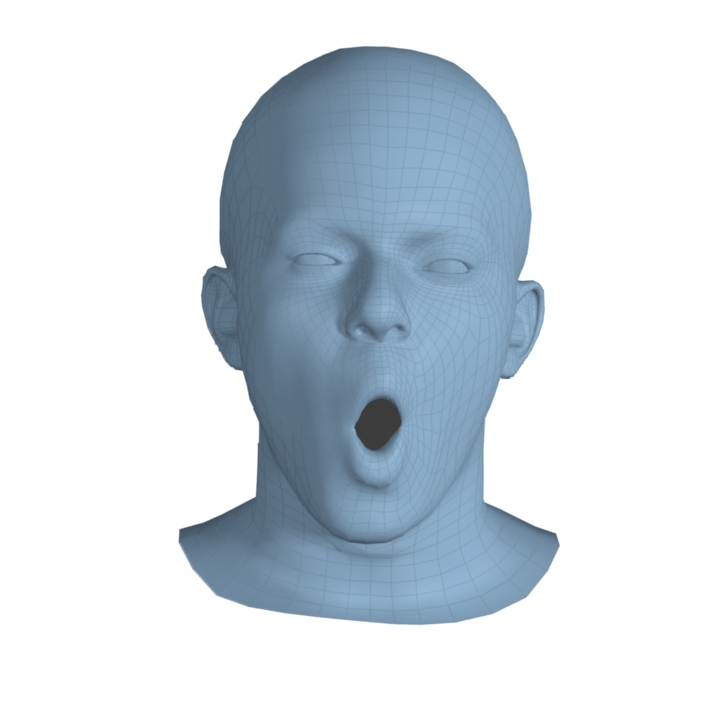}
        \includegraphics[width=\ablmeshsize, clip, trim={\ablimgcropleft} {\ablimgcroplower} {\ablimgcropright} {\ablimgcropupper}]{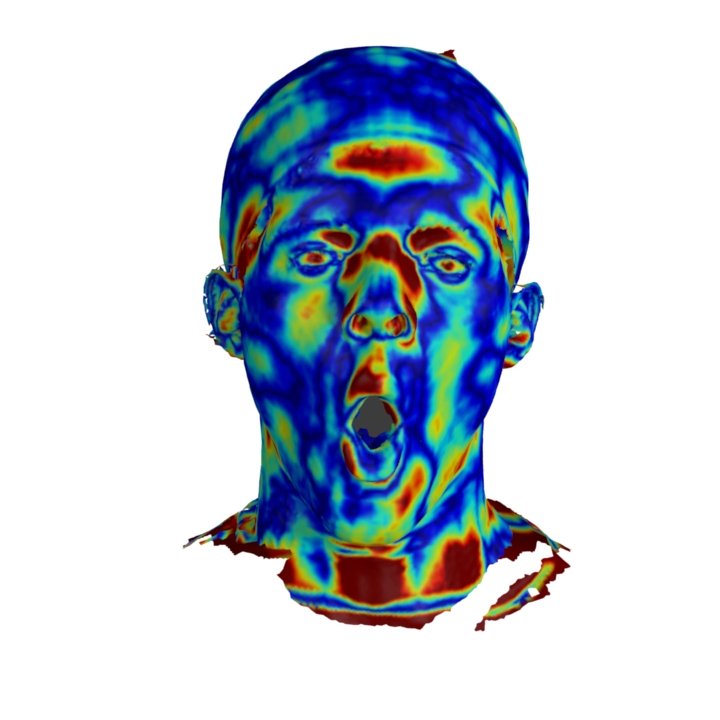}       
        & 
        \includegraphics[width=0.05\linewidth]{images/qualitative_results/color_map_3mm.pdf}
        \\
        Input (4 of 16 views) & Coarse w/o s2m & Coarse w/o head localization & Ours (coarse)
        \\ %
        \includegraphics[width=\ablmeshsize, clip, trim={\ablimgcropleft} {\ablimgcroplower} {\ablimgcropright} {\ablimgcropupper}]{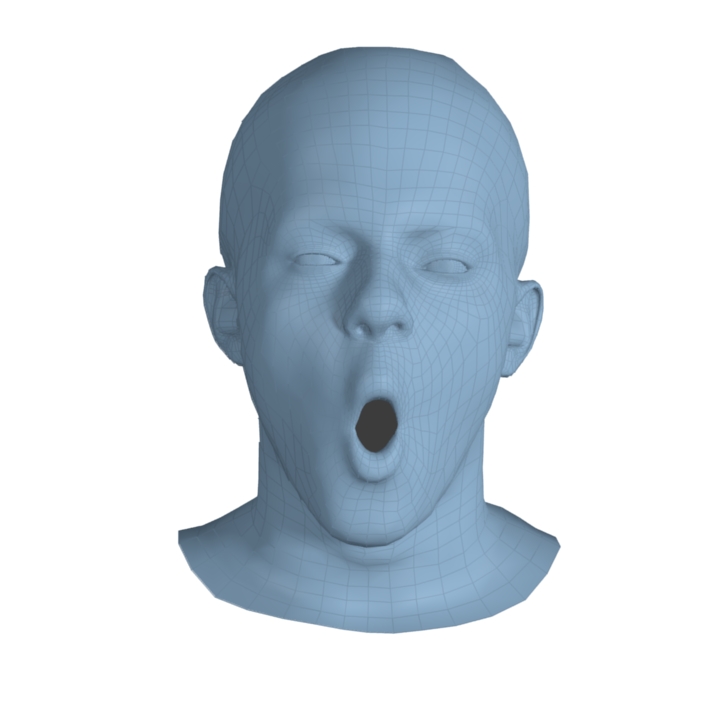}
        \includegraphics[width=\ablmeshsize, clip, trim={\ablimgcropleft} {\ablimgcroplower} {\ablimgcropright} {\ablimgcropupper}]{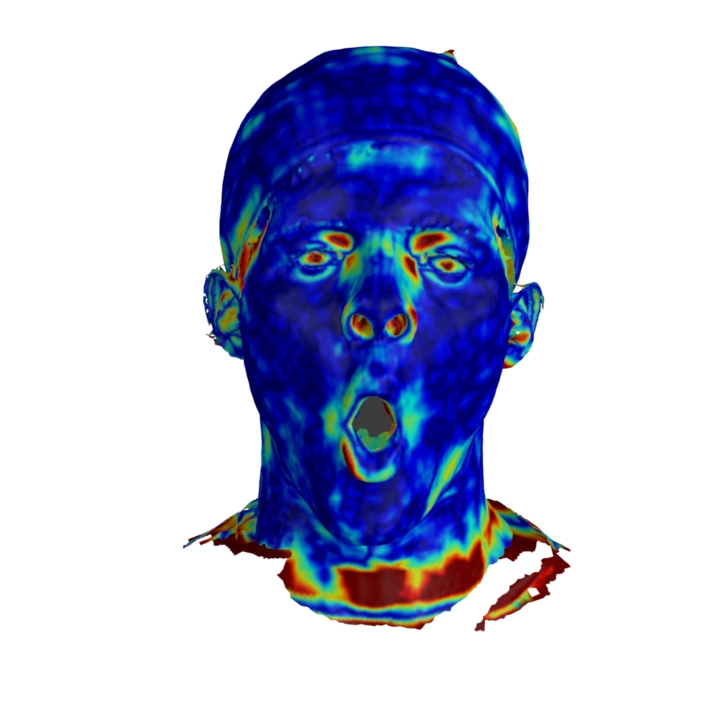}        
        & %
        \includegraphics[width=\ablmeshsize, clip, trim={\ablimgcropleft} {\ablimgcroplower} {\ablimgcropright} {\ablimgcropupper}]{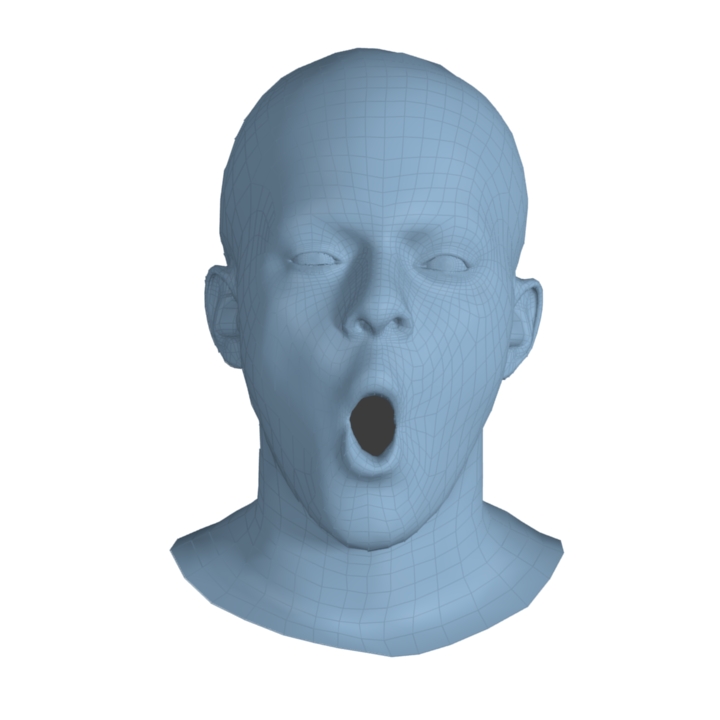}
        \includegraphics[width=\ablmeshsize, clip, trim={\ablimgcropleft} {\ablimgcroplower} {\ablimgcropright} {\ablimgcropupper}]{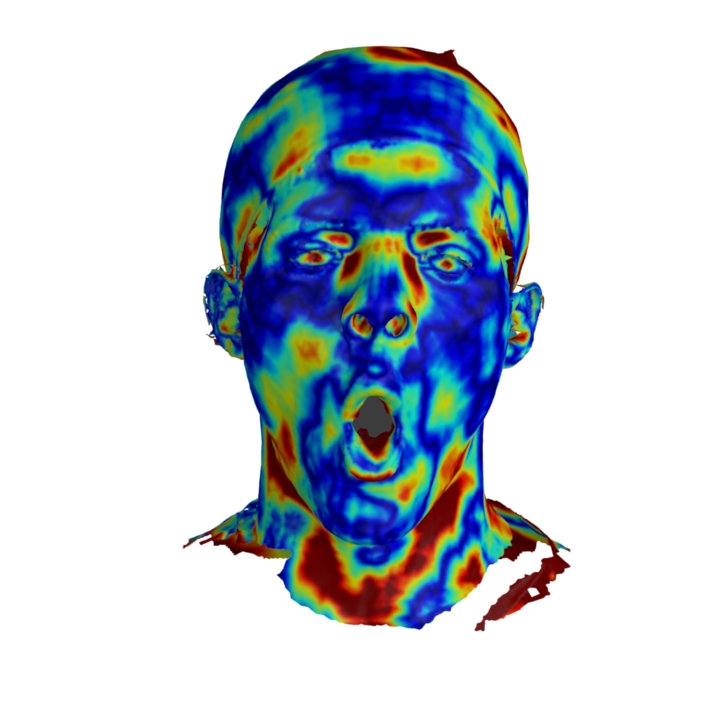}        
        & %
        \includegraphics[width=\ablmeshsize, clip, trim={\ablimgcropleft} {\ablimgcroplower} {\ablimgcropright} {\ablimgcropupper}]{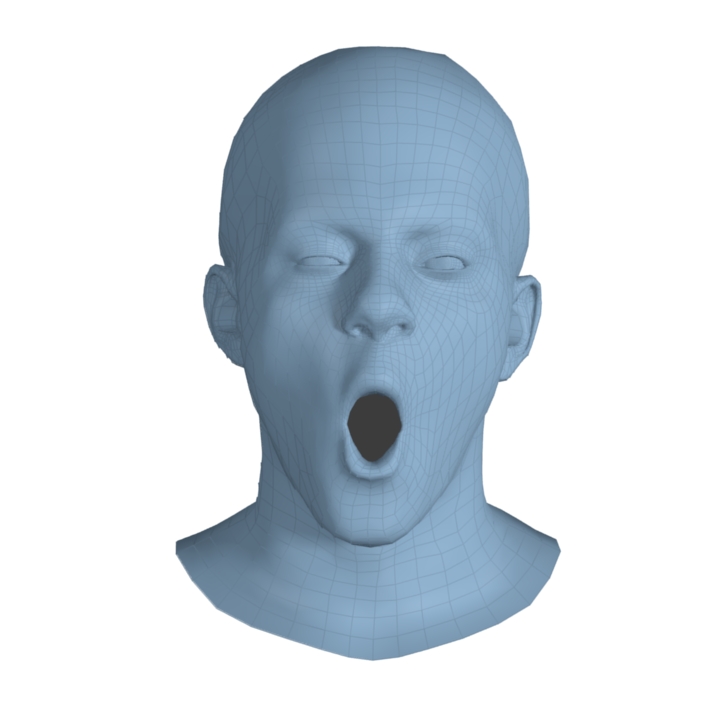}
        \includegraphics[width=\ablmeshsize, clip, trim={\ablimgcropleft} {\ablimgcroplower} {\ablimgcropright} {\ablimgcropupper}]{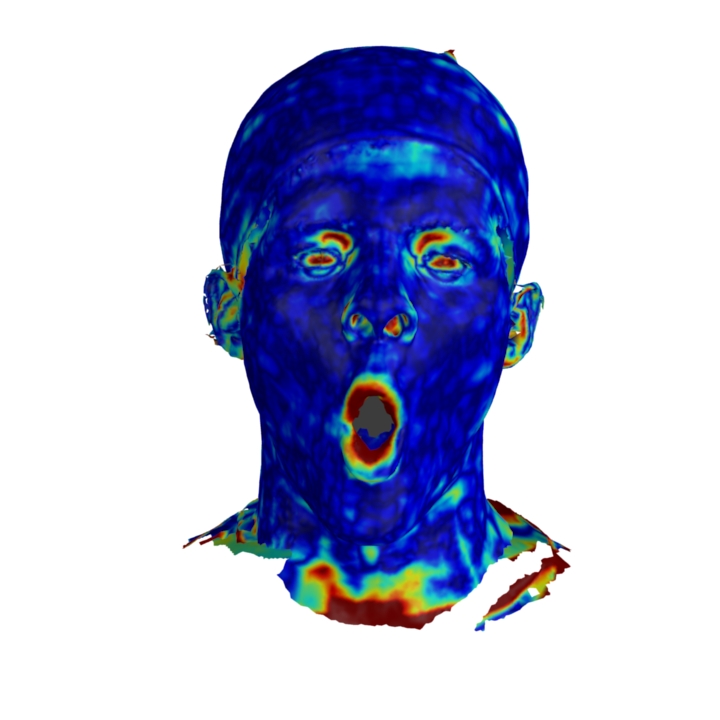}      
        & %
        \includegraphics[width=\ablmeshsize, clip, trim={\ablimgcropleft} {\ablimgcroplower} {\ablimgcropright} {\ablimgcropupper}]{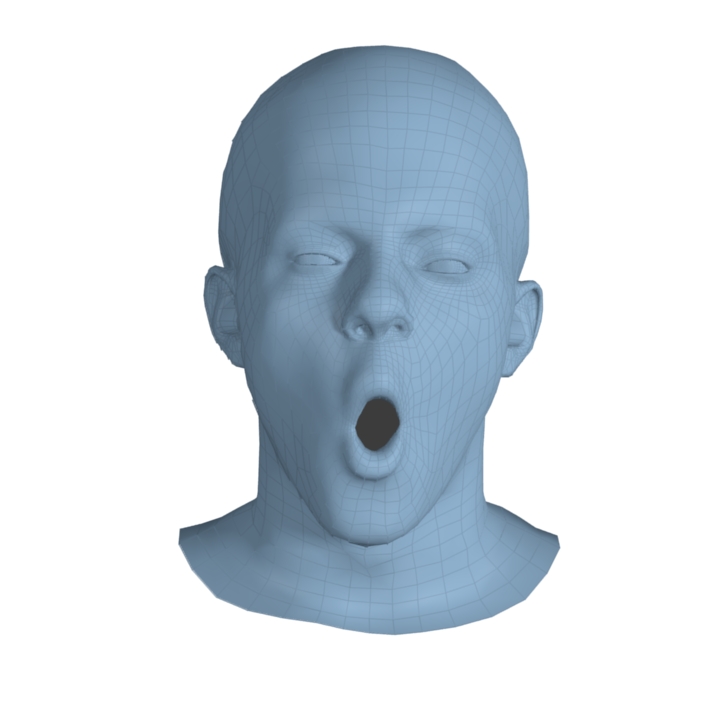}        
        \includegraphics[width=\ablmeshsize, clip, trim={\ablimgcropleft} {\ablimgcroplower} {\ablimgcropright} {\ablimgcropupper}]{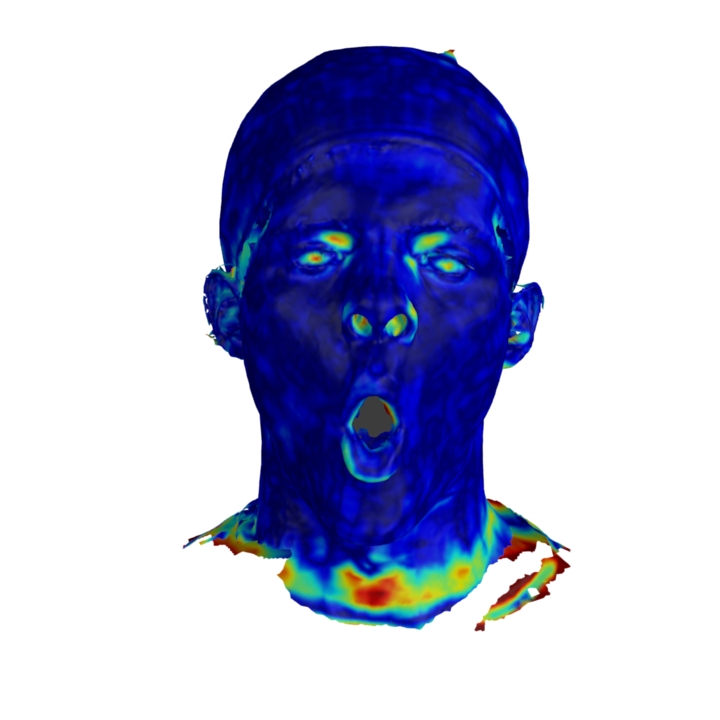}            
        & 
        \includegraphics[width=0.05\linewidth]{images/qualitative_results/color_map_3mm.pdf}      
        \\
        Na\"ive feature fusion & Ours w/o s2m & Ours w/o head localization & Ours              
    \end{tabular}
	\caption{\textbf{Ablation experiments}.
	For each model variant, we show the reconstructed mesh (left) and the color coded point-to-surface distance (right) between the reference scan and the reconstructed mesh as a heatmap on the scan's surface (red means $\geq$ 3 millimeter). 
	}
    \label{fig:ablations}
\end{figure*}

\begin{table*}[ht]
    \centering
    \resizebox{1.0\textwidth}{!}{

\begin{tabular}{l|ccc|ccc|ccc|ccc}
    \toprule
    & \multicolumn{3}{c|}{\bf Complete head} & \multicolumn{3}{c|}{\bf Face} & \multicolumn{3}{c|}{\bf Scalp} & \multicolumn{3}{c}{\bf Neck}  \\    
    Method & Median $\downarrow$ & Mean $\downarrow$ & Std $\downarrow$ & Median $\downarrow$ & Mean $\downarrow$ & Std $\downarrow$ & Median $\downarrow$ & Mean $\downarrow$ & Std $\downarrow$ & Median $\downarrow$ & Mean $\downarrow$ & Std $\downarrow$     \\
    \midrule
    Coarse w/o s2m loss             &  1.15 & 1.85 & 2.76 & 0.91 & 1.21 & 1.56 & 1.22 & 2.13 & 3.44 & 1.75 & 2.60 & 2.91   \\ %
    Coarse w/o head localization    & 1.16 & 1.60 & 1.79 & 1.12 & 1.43 & 1.61 & 1.00 & 1.33 & 1.26 & 1.69 & 2.48 & 2.65 \\ %
    Ours coarse                     & \textbf{0.71} & \textbf{1.11} & \textbf{1.56} & \textbf{0.68} & \textbf{0.93} & \textbf{1.40} & \textbf{0.61} & \textbf{0.92} & \textbf{1.18} & \textbf{1.09} & \textbf{1.81} & \textbf{2.28} \\ %
    \hline
    Refinement w/ na\"ive feature fusion  & 0.35 & 0.70 & 1.36 & 0.27 & 0.45 & 1.26 & 0.36 & 0.63 & 0.90 & 0.54 & 1.23 & 2.06  \\ %
    Ours w/o s2m loss  & 0.78 & 1.44 & 2.59 & 0.64 & 0.89 & 1.43 & 0.76 & 1.62 & 3.22 & 1.27 & 2.18 & 2.81 \\ %
    Ours w/o head localization      & 0.28 & 0.58 & 1.32 & 0.22 & 0.39 & 1.24 & 0.28 & 0.44 & \textbf{0.62} & 0.40 & 1.09 & 2.16  \\ %
    Ours color images input      & 0.44 & 0.77 & 1.33 & 0.34 & 0.54 & 1.27 & 0.46 & 0.71 & 0.93 & 0.65 & 1.28 & 1.94  \\ %
    Ours hierarchical   & 0.27 & 0.61 & 1.55 & \textbf{0.21} & 0.36 & 1.24 & 0.27 & 0.62 & 1.54 & \textbf{0.37} & 0.97 & 2.04  \\ %
    Ours      & \textbf{0.26} & \textbf{0.51} & \textbf{1.22} & \textbf{0.21} & \textbf{0.34} & \textbf{1.22} & \textbf{0.26} & \textbf{0.41} & 0.66 & 0.38 & \textbf{0.95} & \textbf{1.91}  \\ %
    \bottomrule
\end{tabular}

}
	\caption{\textbf{Ablation experiments}. 
	Effects of training from registered meshes instead of scans (w/o s2m loss), reconstructing heads from the entire feature volume (w/o head localization), aggregating features without leveraging surface and visibility information (w/ na\"ive feature fusion), using the capture system's 8 color images as input, and using a hierarchical architecture (ours hierarchical).
	Errors in mm.}
	\label{tab:ablations}
\end{table*}

\paragraph{Qualitative evaluation:}
We evaluate the quality of the predicted meshes on the FaMoS test data, and compare it with the predictions of the current state-of-the-art, ToFu \cite{Li2021_ToFu}.
As the publicly available ToFu model\footnote{https://github.com/tianyeli/tofu} does not generalize to our scanner setup and it only predicts faces, we train ToFu on our training data to predict complete heads. 
For this, we first predict a low-resolution head with 341 vertices in the global stage, followed by upsampling and refining the mesh to the final mesh resolution. 
Unlike ToFu, we use no hierarchical architecture. 
To factor out the potential impact of the intermediate mesh resolution, we additionally compare to ToFu without mesh hierarchy, where the global stage directly predicts meshes of the final resolution, followed by refining the mesh in the local stage. 
We refer to this option as ToFu+. 
To compare with a direct image-to-3DMM regressor, we extend the coarse model of 
DECA~\cite{Feng2021_DECA} to a multi-view setting by regressing FLAME~\cite{Li2017_FLAME} parameters from the concatenated view feature vectors, independently reconstructed for each view. 

Figure~\ref{fig:qualitative_eval} shows that \modelname reconstructs 3D heads with the lowest error in the neck region for extreme head poses (Row 1), and with head shapes closer to the reference scans (Rows 2 \& 3). 
See the Sup.~Mat.~for additional results, including 3DMM regressor details and comparisons.

\paragraph{Quantitative evaluation:}
We quantify the accuracy of the predicted 3D heads on the FaMoS test data by computing the point-to-surface distances between the vertices of each reference scan, and their closest points in the predicted 3D head's surface.
To analyze the accuracy in different head regions, we segment each scan into face, scalp, and neck regions (see Sup.~Mat.), and report the reconstruction errors for the entire head and the individual segments. 
Table~\ref{tab:qualitative_eval} shows that \modelname outputs 3D heads with 64\% lower reconstruction error compared to ToFu, and 68\% lower than ToFu+.
See the Sup.~Mat.~for cumulative error plots and additional qualitative comparisons.

Training \modelname minimizes the distance between the predicted heads and \ac{MVS} scans, hence it effectively registers the scans. 
\modelname closely fits the training scans, with a median error of $0.17$~mm (coarse stage: $0.80$~mm).

\paragraph{Ablation experiments:}
To quantify the impact of individual design choices, we train following model variants:
(1) \textit{Coarse w/o s2m loss}: fully supervised training with v2v loss only (Eq.~\ref{eq:v2v}) (i.e., no scan supervision).
(2) \textit{Coarse w/o head localization}: direct prediction of the head mesh from the coarse feature volume without head localization.
(3) \textit{Refinement w/ na\"ive feature fusion}: feature aggregation as mean and variance across views, without surface-aware feature fusion.
(4) \textit{Ours w/o s2m loss}: training of coarse and refinement stage with v2v loss only.
(5) \textit{Ours w/o head localization}: coarse model w/o head localization with refinement stage.
(6) \textit{Ours color images input}:  use of color images as input, instead of gray-scale stereo images.
(7) \textit{Ours hierarchical}: predicting a mesh with 1000 vertices in the coarse stage, followed by upsampling and refinement in the second stage. 
Fig.~\ref{fig:ablations} and Tab.~\ref{tab:ablations} compare different model variants qualitatively and quantitatively. 
We find that both, surface distance loss and head localization are essential for the coarse head inference, as ablating either of them leads to worse performance. 
Further, our model with surface-aware feature fusion and surface distance loss predicts heads with lowest error. 
While the models without head localization (5) or with hierarchical architecture (7) infer heads with comparably low errors, they reconstruct e.g., the lips region with lower fidelity due to the worse expression initialization from the coarse stage (see Fig.~\ref{fig:ablations} and Sup.~Mat.).

\section{Discussion}
\label{sec:discussion}

\paragraph{Reference registrations:}
\modelname uses reference registrations for pre-training and regularization. 
While these registrations are obtained fully automatically \cite{Li2017_FLAME}, their computation is slow and computationally expensive. 
Instead, directly regularizing to a statistical model \cite{Egger2020_Survey} during training by jointly optimizing the statistical model's parameters (i.e., as done for coupled registrations \cite{Li2017_FLAME}) could mitigate the need for registrations.
This, however, adds an additional level of complexity, which goes beyond our current scope.

\paragraph{Registration quality:}
While \modelname's reconstructions well resemble the reference scans, expressions like eye blinks are not well captured. 
This is due to the poor quality of the scans in the eye region, the fast motion of the eyelids, and the absence of a clear signal in the optimized point-to-surface distance (Eq.~\ref{eq:s2m}).
We plan to add an additional eyelid landmark error to improve the eyelid tracking in the future.

\paragraph{Representation:}
Several methods exist to learn deep implicit functions with dense correspondence from scans \cite{Zheng2021_Implicit,Zheng2022_ImFace,LiuLiu2020_Implicit}. 
Replacing \modelname's mesh representation with such implicit functions is an interesting future direction.

\paragraph{Camera calibrations:}
\modelname is designed for lab environments with a carefully calibrated capture system. 
Adapting \modelname to less constrained scenarios with noisy or unknown camera calibration goes beyond the current scope.
\section{Conclusion}
\label{sec:conclusion}

We have presented \modelname, a framework to predict entire 3D heads in dense  correspondence from calibrated multi-view images.
\modelname infers 3D heads with reconstruction accuracy that is 64\% lower than the previous state-of-the-art. 
We achieve this by training \modelname directly from scans, using a spatial transformer head localization module, and surface-aware feature fusion. 
Intuitively, the training from scans overcomes ambiguous correspondence across subjects and imperfect correspondence across expressions.
The head localization enables the coarse stage to handle a large capture volume by focusing on the region of interest, and it provides a better initialization for the geometry refinement. 
The surface-aware feature fusion accounts for self-occlusions.
Due to the inferred geometry accuracy and inference speed, \modelname is useful for applications like multi-view head performance capture. 

\paragraph{Acknowledgement:} 
We thank T. Alexiadis, M. Höschle, Y. Fincan, B. Pellkofer for data capture and IT support, T. McConnell for voice over, and R. Daněček, W. Zielonka, P. Patel, and P. Kulits for proofreading.
\paragraph{Disclosure:} 
\small{\url{https://files.is.tue.mpg.de/tbolkart/disclosure.txt}}

{\small
\balance

}

\newpage
\nobalance
\appendix
\section{Appendix}

\paragraph{Multi-view setup:}
\modelname infers 3D head meshes in correspondence from calibrated multi-view images.
Specifically, we use the eight pairs of gray-scale stereo images of an active stereo camera system as input (see Sec.~\red{4} of the paper for details). 
\begin{figure}[ht]
    \includegraphics[width=1.0\columnwidth]{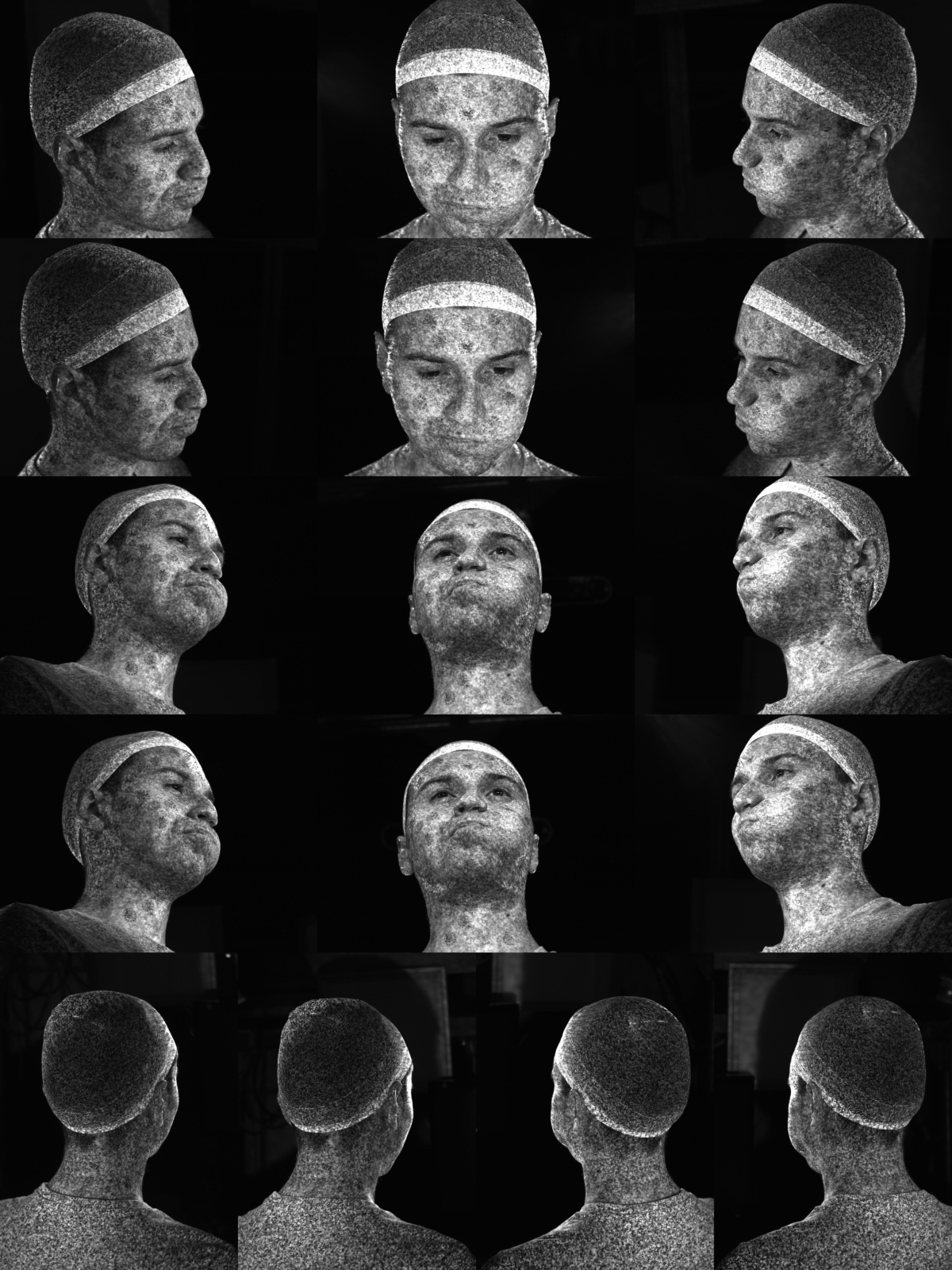}
	\caption{\textbf{Multi-view setup}. The 16 gray-scale stereo images (contrast enhanced for visualization) used as input to \modelname.}
    \label{fig:scannersetup}
\end{figure}
Figure~\ref{fig:scannersetup} shows the 16 images for the first sample of the paper's teaser figure.

\paragraph{Head localization:}
The coarse head prediction stage localizes the head in the feature volume with a learnable spatial transformer. 
Figure~\ref{fig:head_localization} visualizes the spatial dimensions of the localized volume for different predicted heads of the coarse stage. 
This shows that the spatial transformer successfully localizes the head for different subjects in varying head poses. 
\newcommand{\locimgcropleft}{210}
\newcommand{\locimgcroplower}{150}
\newcommand{\locimgcropright}{160}
\newcommand{\locimgcropupper}{210}

\begin{figure}[ht]
    \centering
    \begin{tabular}{@{}c@{}c@{}c@{}c@{}}
        \includegraphics[width=0.25\columnwidth, clip, trim={\locimgcropleft} {\locimgcroplower} {\locimgcropright} {\locimgcropupper}]{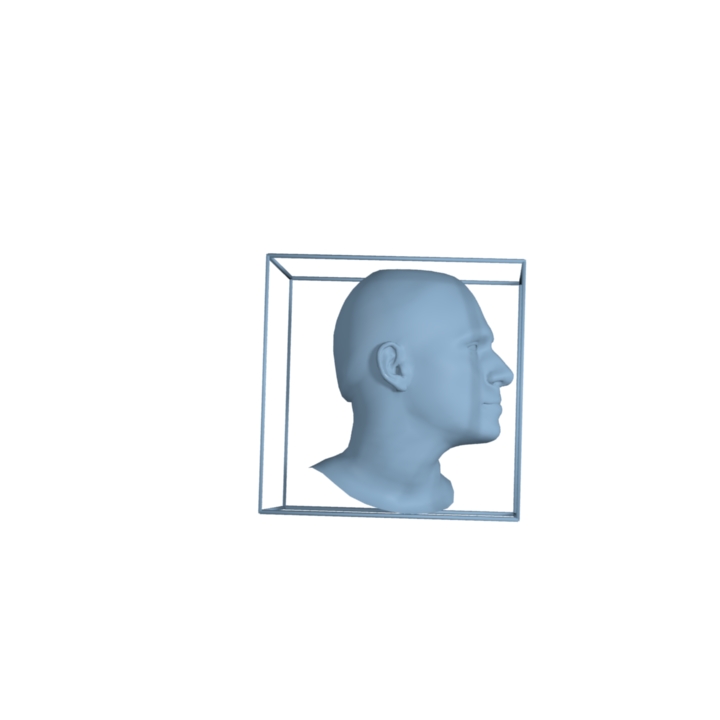} &
        \includegraphics[width=0.25\columnwidth, clip, trim={\locimgcropleft} {\locimgcroplower} {\locimgcropright} {\locimgcropupper}]{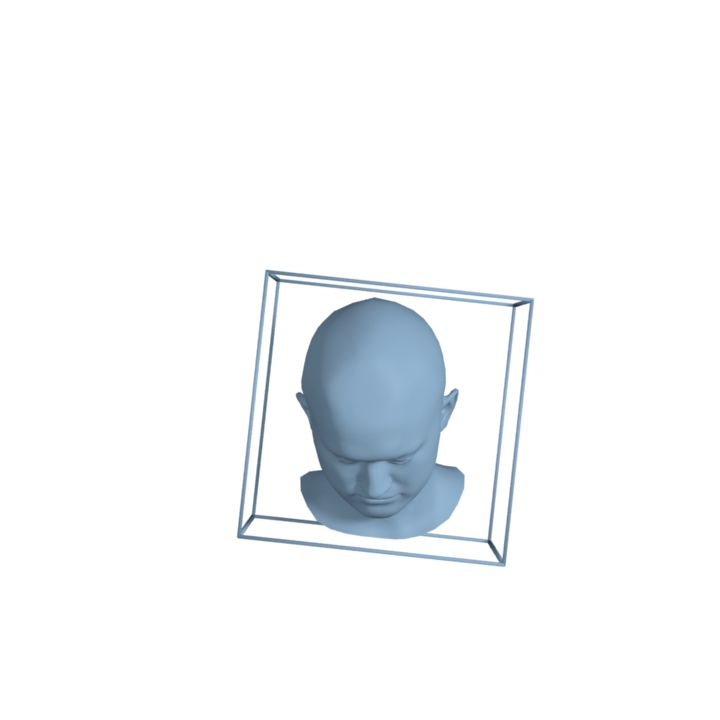} &
        \includegraphics[width=0.25\columnwidth, clip, trim={\locimgcropleft} {\locimgcroplower} {\locimgcropright} {\locimgcropupper}]{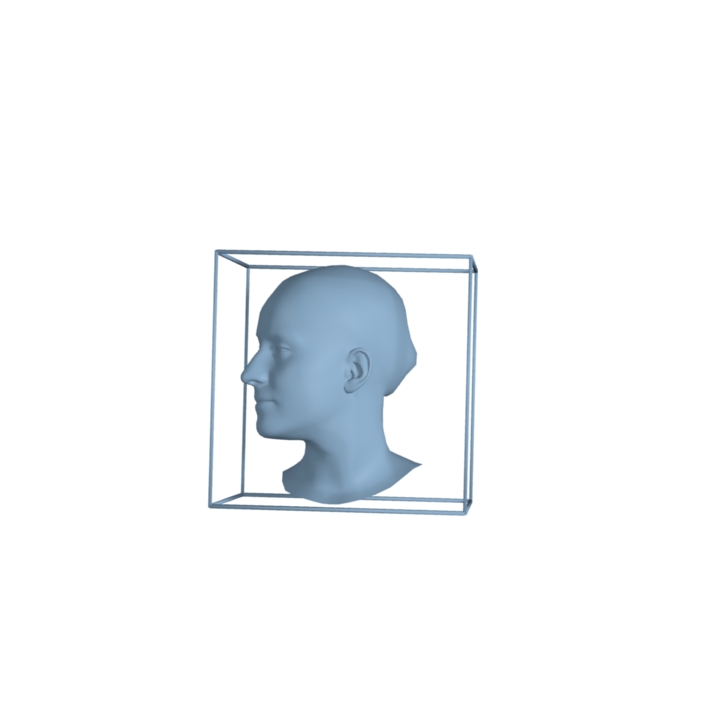} &
        \includegraphics[width=0.25\columnwidth, clip, trim={\locimgcropleft} {\locimgcroplower} {\locimgcropright} {\locimgcropupper}]{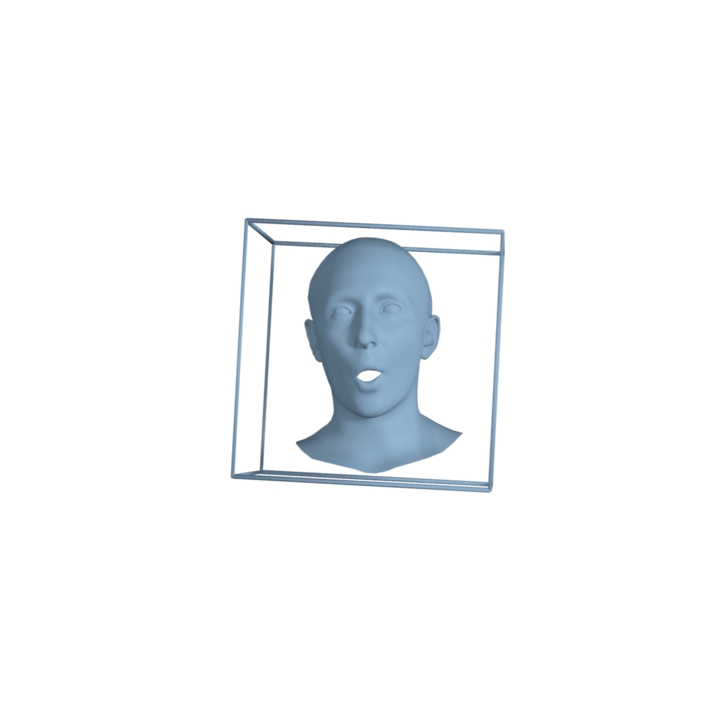}     
    \end{tabular}
	\caption{\textbf{Head localization}. Spatial dimensions of the localized feature volumes (blue box) for different predicted heads, visualized in the same global coordinate system.
	}
    \label{fig:head_localization}
\end{figure}

\paragraph{Error masks:}
To analyze the accuracy in different head regions, reconstruction errors are reported individually for the face, scalp, and neck regions. 
For this, different regions are defined on a FLAME template mesh (see Figure~\ref{fig:error_masks}), and each scan is then segmented based on the distance to the closest points in the surface of the reference registrations.

\newcommand{\errimgsize}{0.24\columnwidth}
\newcommand{\errimgcropleft}{250}
\newcommand{\errimgcroplower}{60}
\newcommand{\errimgcropright}{100}
\newcommand{\errimgcropupper}{60}

\begin{figure}[t]
    \centering
    \includegraphics[width=\errimgsize, clip, trim={\errimgcropleft} {\errimgcroplower} {\errimgcropright} {\errimgcropupper}]{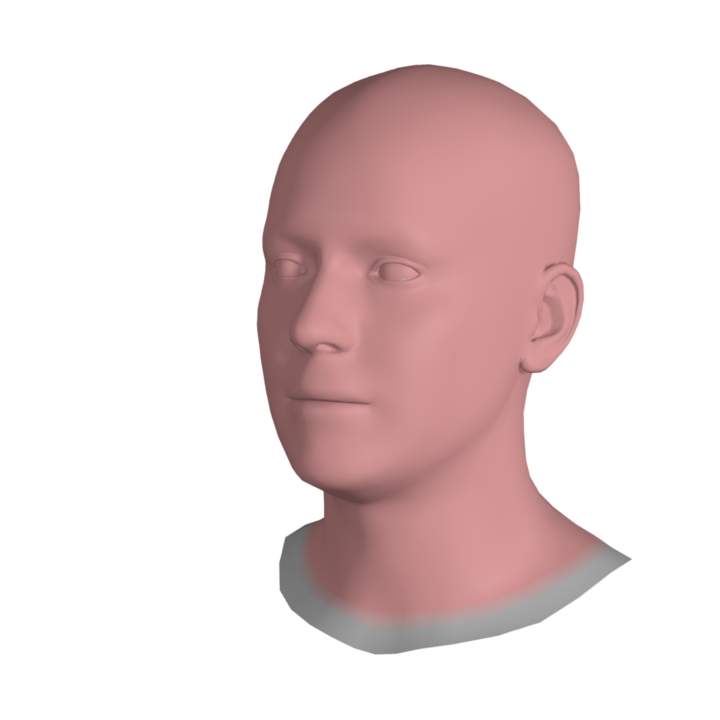}     
    \includegraphics[width=\errimgsize, clip, trim={\errimgcropleft} {\errimgcroplower} {\errimgcropright} {\errimgcropupper}]{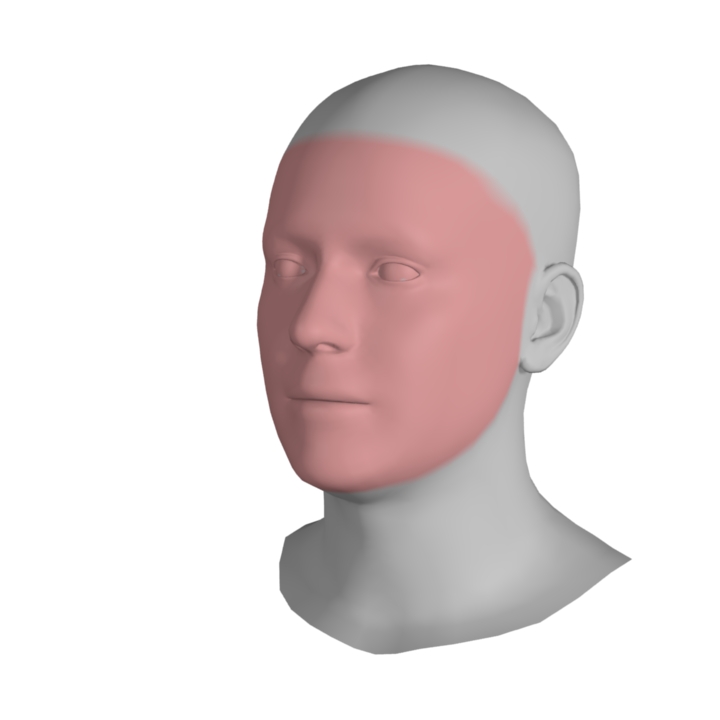} 
    \includegraphics[width=\errimgsize, clip, trim={\errimgcropleft} {\errimgcroplower} {\errimgcropright} {\errimgcropupper}]{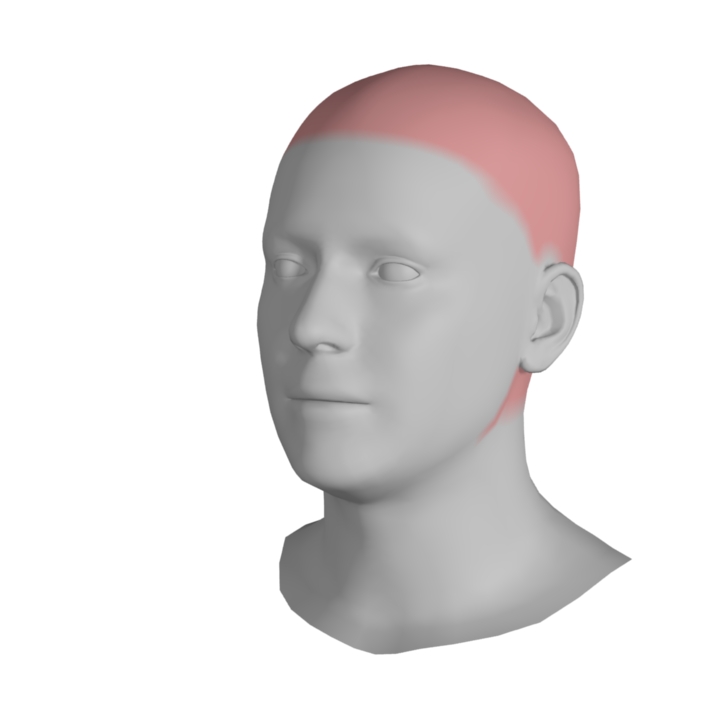} 
	\includegraphics[width=\errimgsize, clip, trim={\errimgcropleft} {\errimgcroplower} {\errimgcropright} {\errimgcropupper}]{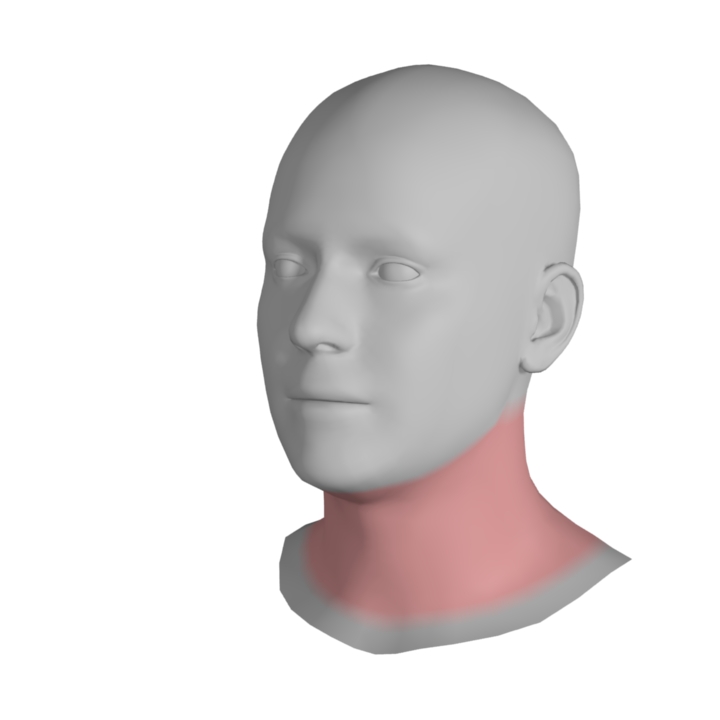} 
	\caption{\textbf{Error masks}. 
	Head regions (red) for quantitative evaluations. 
	From left to right: \textit{complete head}, \textit{face}, \textit{scalp}, and \textit{neck}.}
    \label{fig:error_masks}
\end{figure}

\begin{figure}[ht]
    \centering
    \begin{tabular}{@{}c@{}c@{}c@{}c@{}}
        \includegraphics[width=0.5\columnwidth]{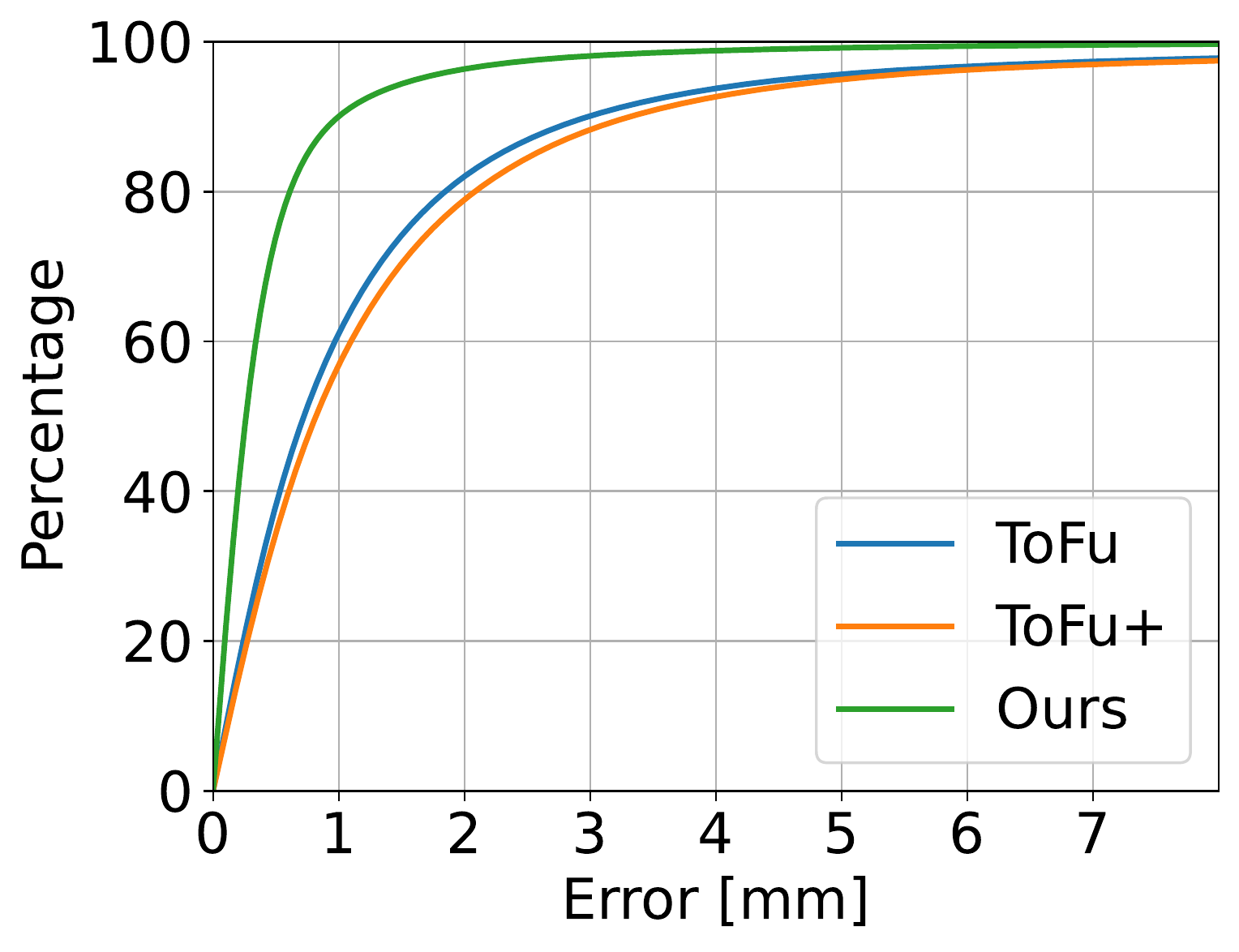} &
        \includegraphics[width=0.5\columnwidth]{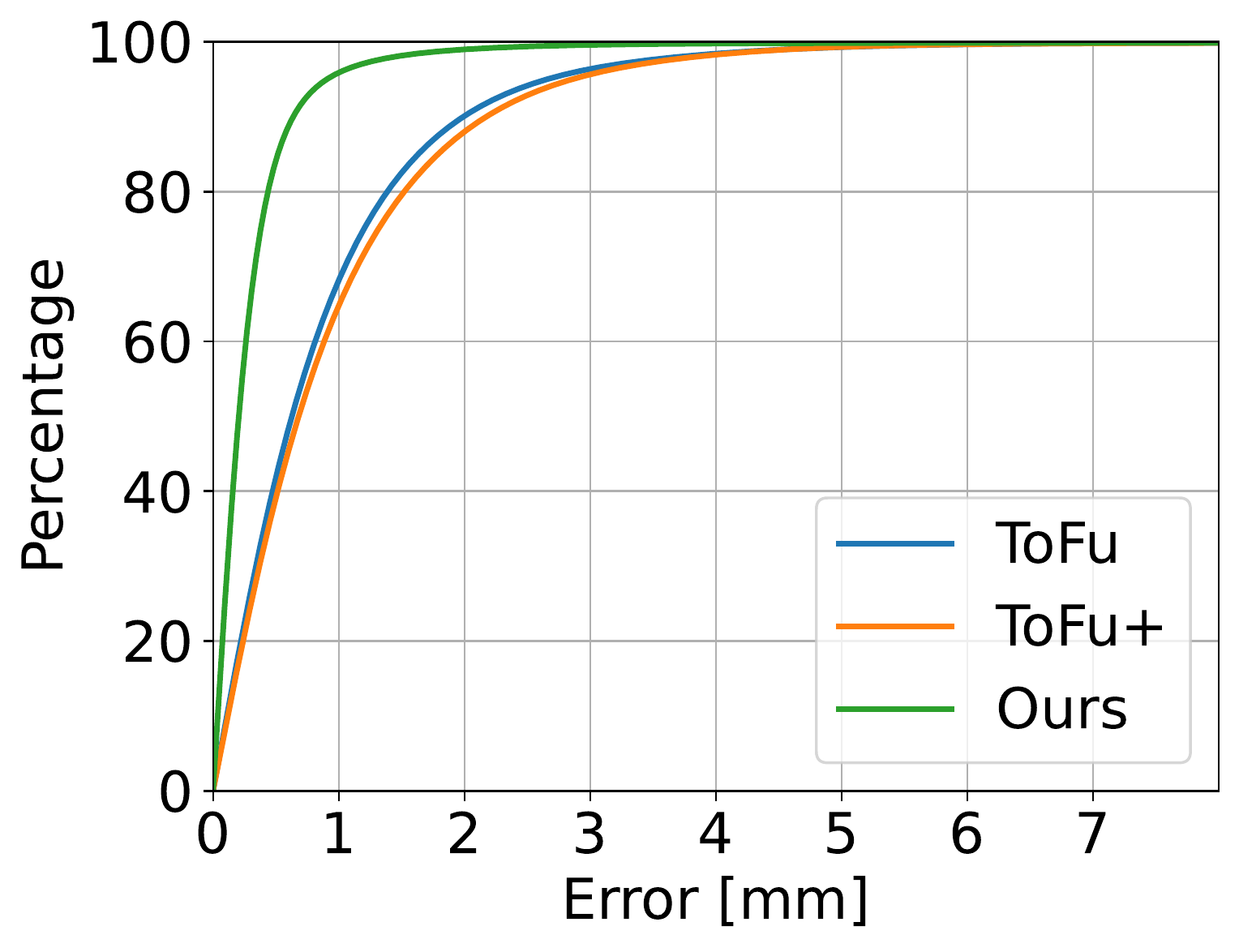} \\
        Complete head & Face \\
        \includegraphics[width=0.5\columnwidth]{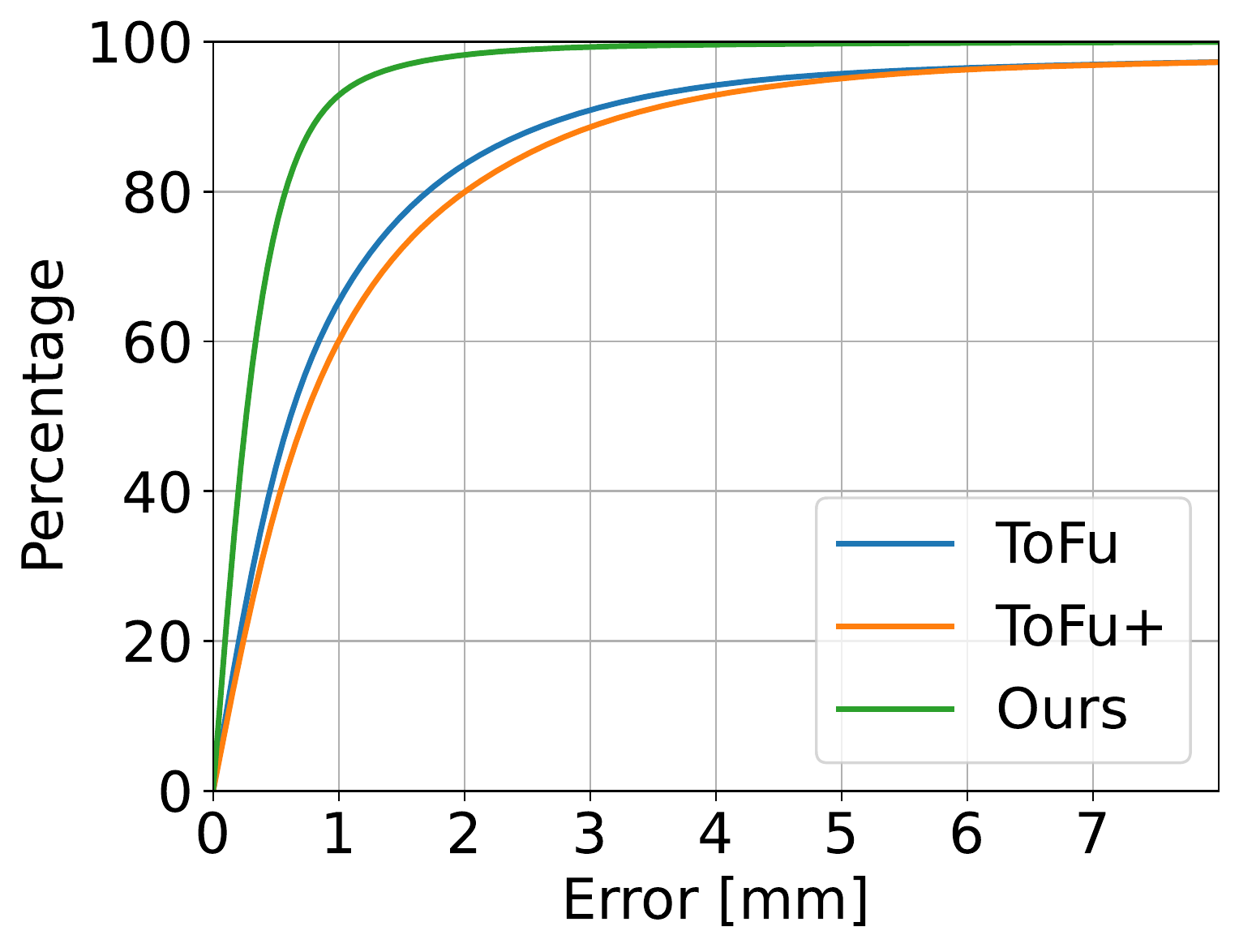} &
        \includegraphics[width=0.5\columnwidth]{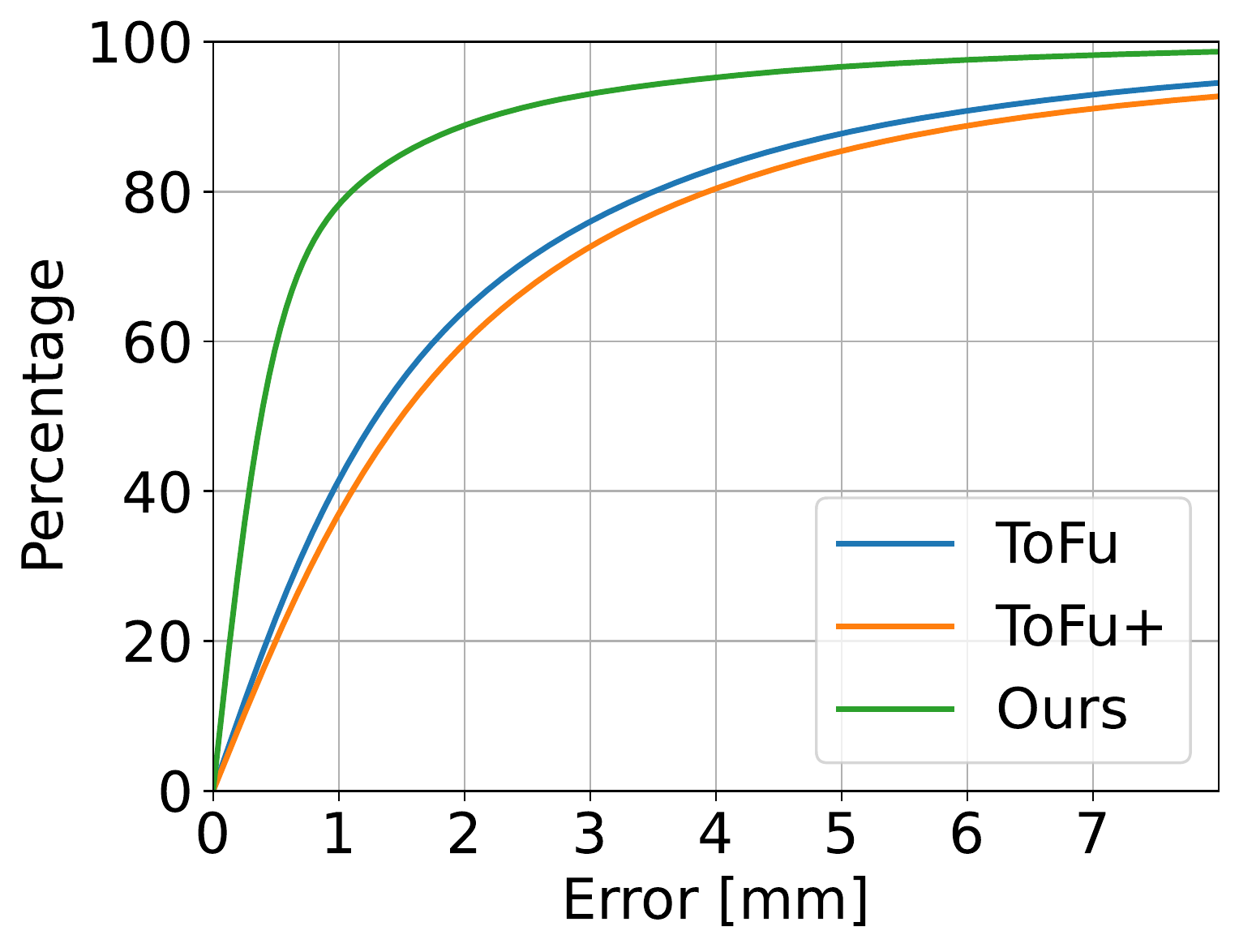} 
        \\
        Scalp & Neck 
    \end{tabular}
	\caption{\textbf{Quantitative evaluation}.
	Cumulative plots of the reconstruction errors on the FaMoS test data. 
	}
    \label{fig:qual_cumulative}
\end{figure}

\paragraph{Test evaluation:}
Figure~\ref{fig:qual_cumulative} provide the cumulative reconstruction errors for the FaMoS test data. 
\modelname predicts heads for the test images (subjects disjoint from the training subjects) with a lower error than previous state-of-the art, ToFu \cite{Li2021_ToFu}, and its variant without mesh hierarchy, ToFu+.

\begin{table*}[ht]
    \resizebox{1.0\textwidth}{!}{

\begin{tabular}{l|ccc|ccc|ccc|ccc}
    \toprule
    & \multicolumn{3}{c|}{\bf Complete head} & \multicolumn{3}{c|}{\bf Face} & \multicolumn{3}{c|}{\bf Scalp} & \multicolumn{3}{c}{\bf Neck}  \\
    Method & Median $\downarrow$ & Mean $\downarrow$ & Std $\downarrow$ & Median $\downarrow$ & Mean $\downarrow$ & Std $\downarrow$ & Median $\downarrow$ & Mean $\downarrow$ & Std $\downarrow$ & Median $\downarrow$ & Mean $\downarrow$ & Std $\downarrow$    \\
    \midrule
    Ours (coarse)   & 0.80 & 1.61 & 3.86 & 0.67 & 0.85 & 1.31 & 0.84 & 2.31 & 5.59 & 1.11 & 1.68 & 2.34  \\ 
    Ours            & 0.17 & 0.30 & 0.97 & 0.14 & 0.23 & 1.10 & 0.16 & 0.24 & 0.39 & 0.24 & 0.53 & 1.46 \\ 
    \bottomrule
\end{tabular}

}
	\caption{\textbf{Registration quality}. 
	Reconstruction errors on the FaMoS training data. All errors are in millimeter.}
	\label{tab:registration_error}
\end{table*}

\begin{figure}[ht]
    \centering
    \begin{tabular}{c@{}c@{}c@{}c}
        \includegraphics[width=0.35\columnwidth]{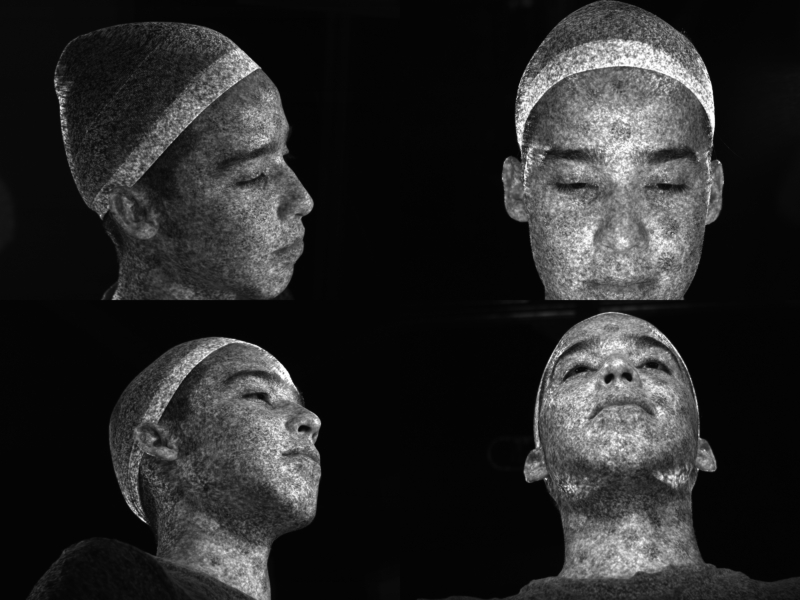}  &
        \includegraphics[width=0.2\columnwidth, clip, trim=100 0 100 0]{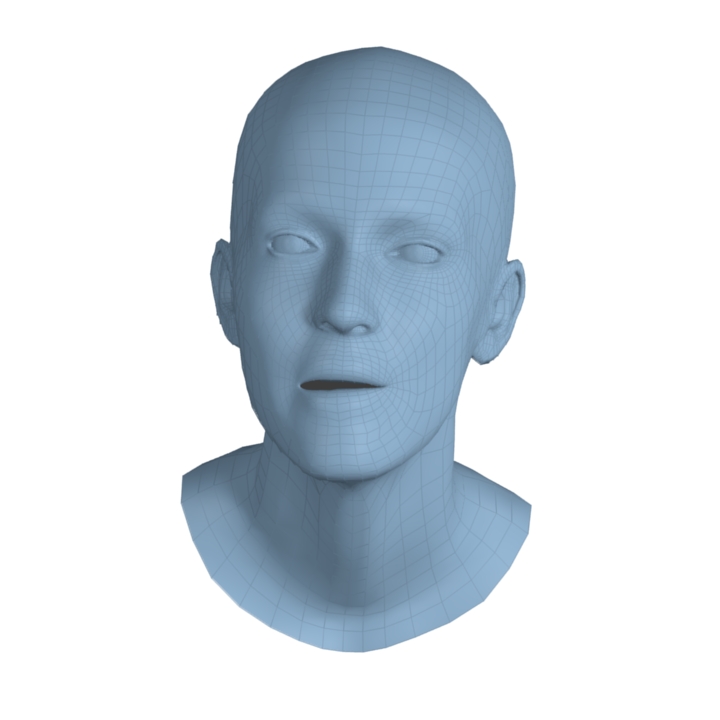} &
        \includegraphics[width=0.2\columnwidth, clip, trim=100 0 100 0]{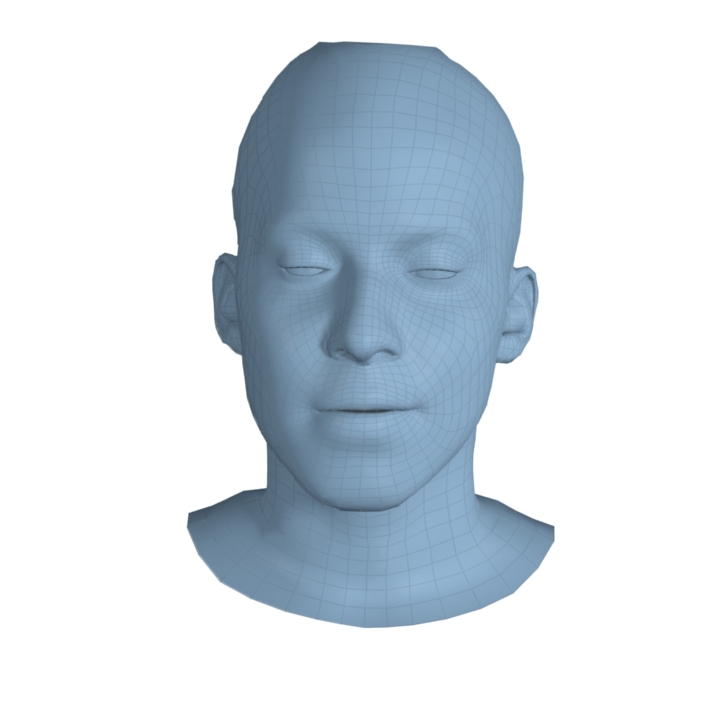} &
        \includegraphics[width=0.2\columnwidth, clip, trim=100 0 100 0]{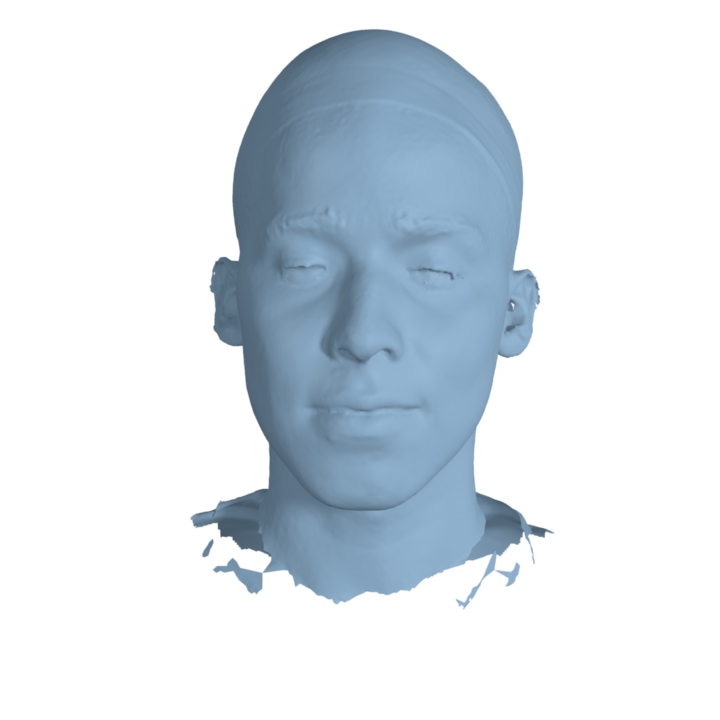}    \\
        Input & Regressor & Ours & Scan
    \end{tabular}
	\caption{\textbf{3DMM regressor comparison}. For multi-view input (left: 4 of 16 views), the 3DMM regressor regressor (second column) is unable to faithfully reconstruct the identity face shape, while \modelname (third column) closely resembles the reference scan (right).
	}
    \label{fig:comparisons}
\end{figure}

\paragraph{3DMM regressor comparisons:}
For the multi-view 3DMM regressor, each image is processed by a shared ResNet152 \cite{He2016_ResNet} to infer a 2048-dimensional feature vector for each view. 
The feature vectors are then fused across all 16 views by concatenating them in a fixed order. 
We experimented with other feature fusion variants such as using the mean across views, or the concatenated mean and variance, but these variants produced inferior results. 
Following DECA \cite{Feng2021_DECA}, a fully-connected layer with ReLU activations outputs a 1024-dimensional feature vector, followed by final linear layer to output FLAME parameters.
We train the 3DMM regressor for 1 Million iterations with a vertex-to-vertex to the reference registrations, with a learning rate of 1e-3.
We found that the 3DMM regressor is unable to reliably reconstruct 3D heads in our setting (see Fig.~\ref{fig:comparisons}). 

\paragraph{Registration quality:}
The training of \modelname minimizes the distance to \ac{MVS} scans, hence it effectively registers the scans. 
For completeness, we also report the registration errors in Table~\ref{tab:registration_error}.
For this, we predict all heads from the training images, and compute the point-to-surface distance for all \ac{MVS} scan points. 

\begin{figure*}[ht]
    \centering
    \begin{tabular}{@{}c@{\hskip \ablcolmargin}c@{\hskip \ablcolmargin}c@{\hskip \ablcolmargin}c@{\hskip \ablcolmargin}r@{}}
        \includegraphics[width=\ablimgsize, clip]{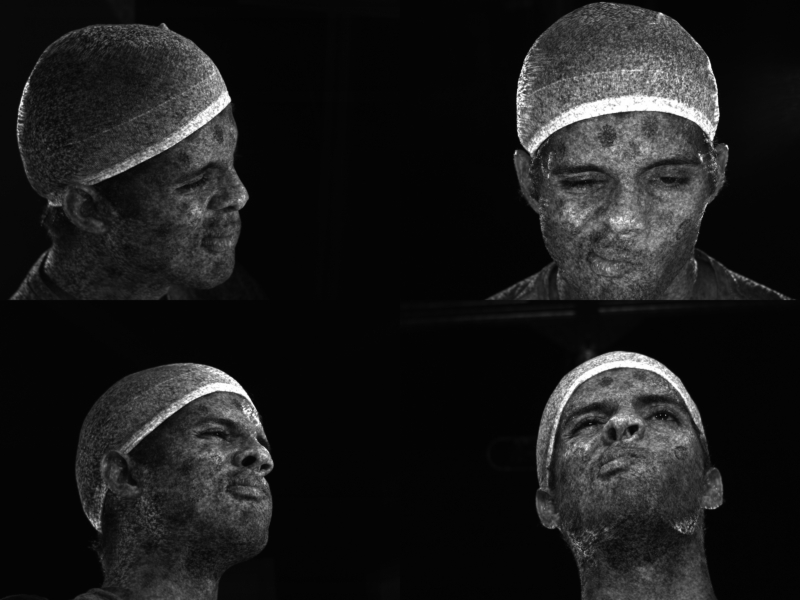} 
        &
        \includegraphics[width=\ablmeshsize, clip, trim={\ablimgcropleft} {\ablimgcroplower} {\ablimgcropright} {\ablimgcropupper}]{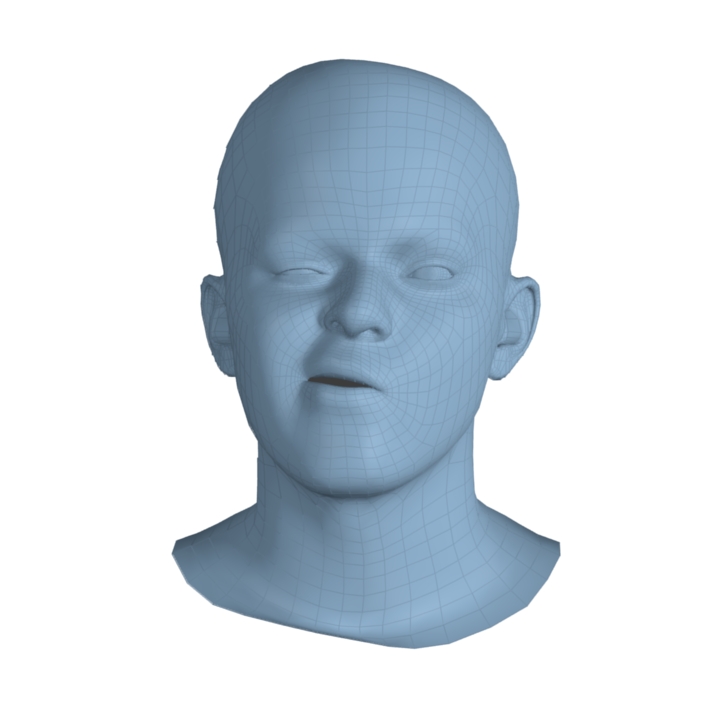}
        \includegraphics[width=\ablmeshsize, clip, trim={\ablimgcropleft} {\ablimgcroplower} {\ablimgcropright} {\ablimgcropupper}]{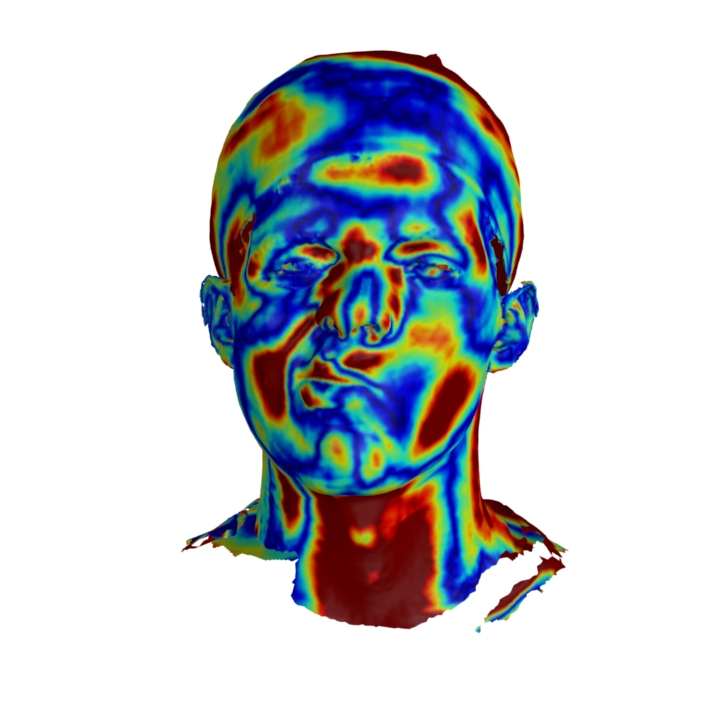}
        &
        \includegraphics[width=\ablmeshsize, clip, trim={\ablimgcropleft} {\ablimgcroplower} {\ablimgcropright} {\ablimgcropupper}]{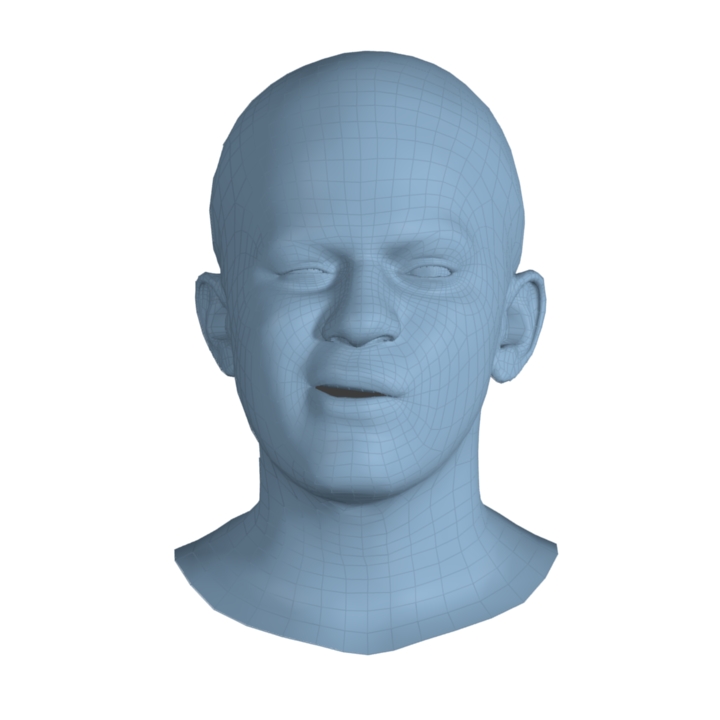}
        \includegraphics[width=\ablmeshsize, clip, trim={\ablimgcropleft} {\ablimgcroplower} {\ablimgcropright} {\ablimgcropupper}]{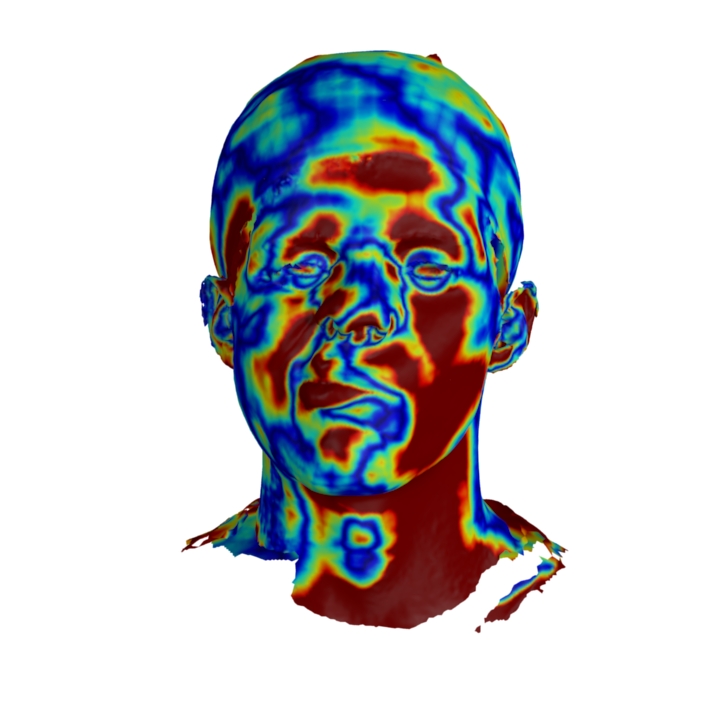}
        &
        \includegraphics[width=\ablmeshsize, clip, trim={\ablimgcropleft} {\ablimgcroplower} {\ablimgcropright} {\ablimgcropupper}]{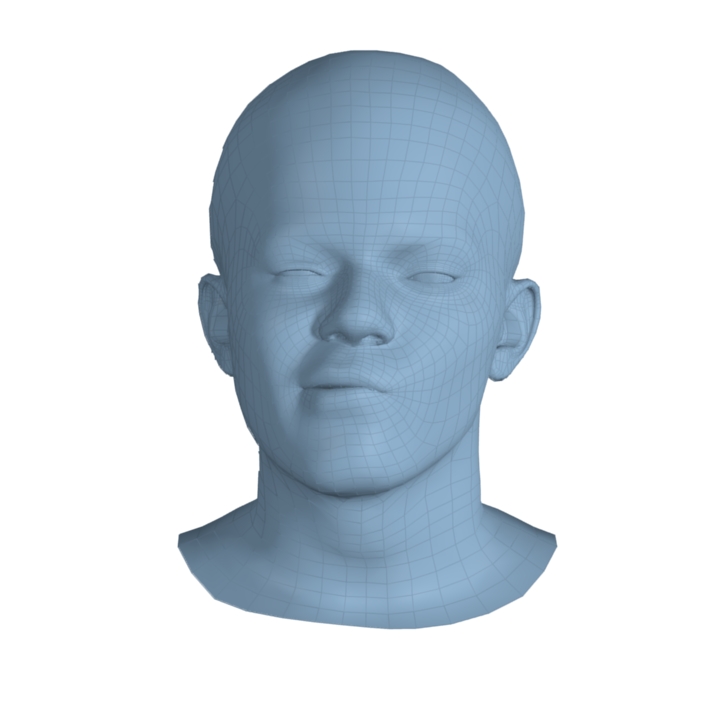}
        \includegraphics[width=\ablmeshsize, clip, trim={\ablimgcropleft} {\ablimgcroplower} {\ablimgcropright} {\ablimgcropupper}]{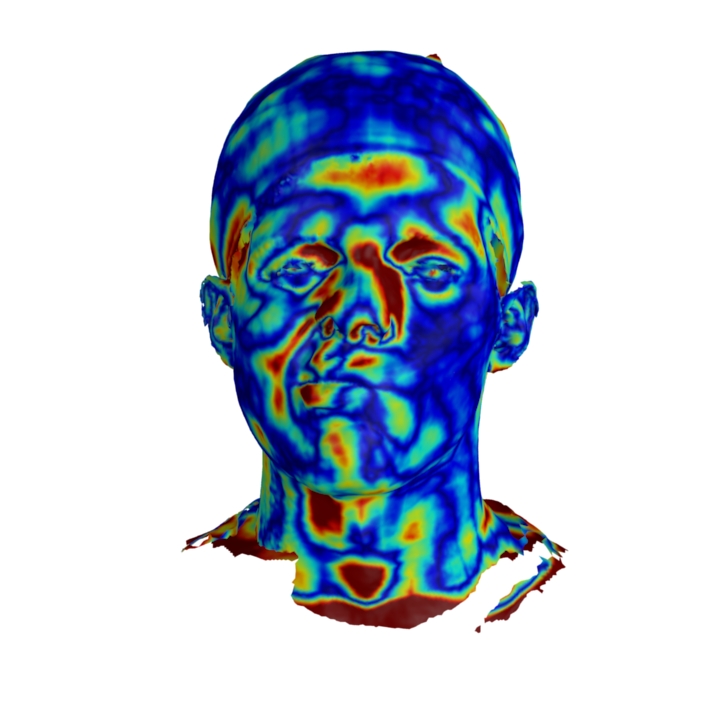}       
        & 
        \includegraphics[width=0.05\linewidth]{images/qualitative_results/color_map_3mm.pdf}
        \\
        Input (4 of 16 views) & Coarse w/o s2m & Coarse w/o head localization & Ours (coarse)
        \\
        \includegraphics[width=\ablmeshsize, clip, trim={\ablimgcropleft} {\ablimgcroplower} {\ablimgcropright} {\ablimgcropupper}]{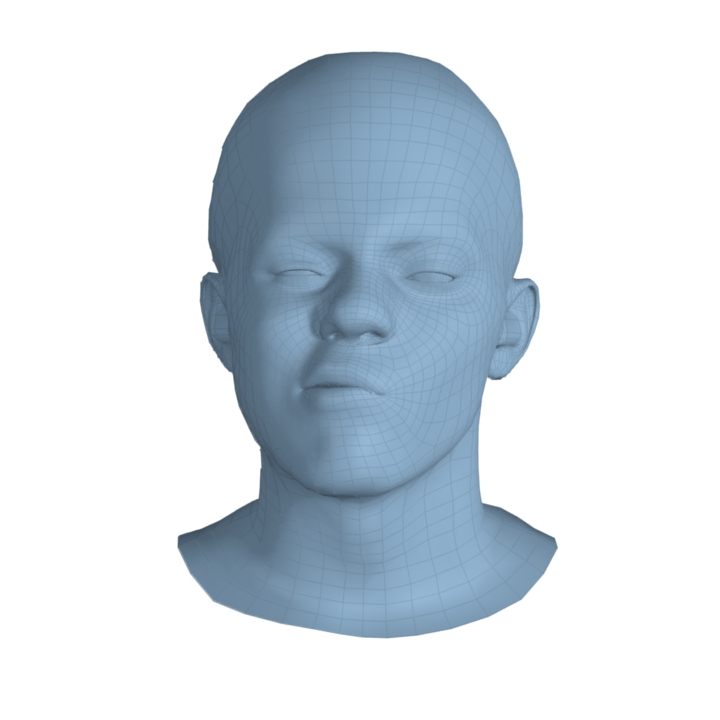}
        \includegraphics[width=\ablmeshsize, clip, trim={\ablimgcropleft} {\ablimgcroplower} {\ablimgcropright} {\ablimgcropupper}]{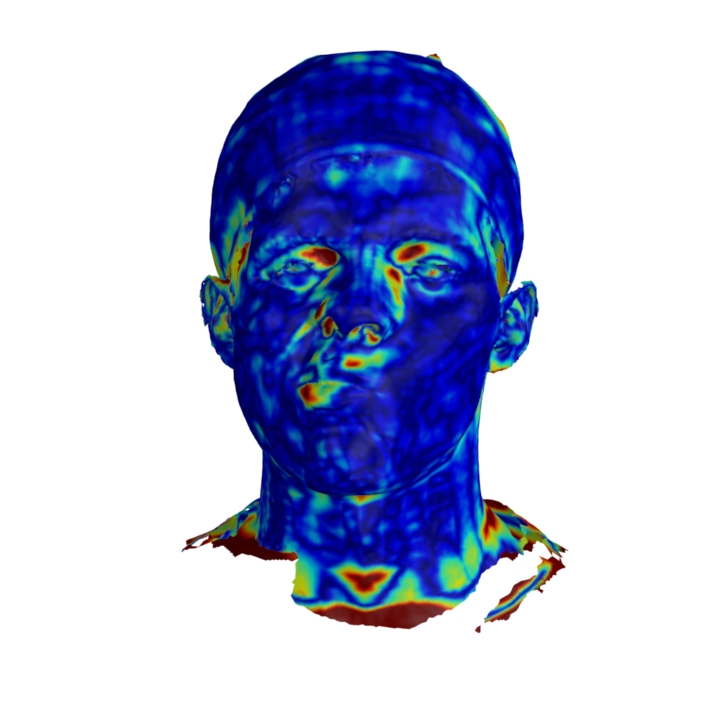}        
        & 
        \includegraphics[width=\ablmeshsize, clip, trim={\ablimgcropleft} {\ablimgcroplower} {\ablimgcropright} {\ablimgcropupper}]{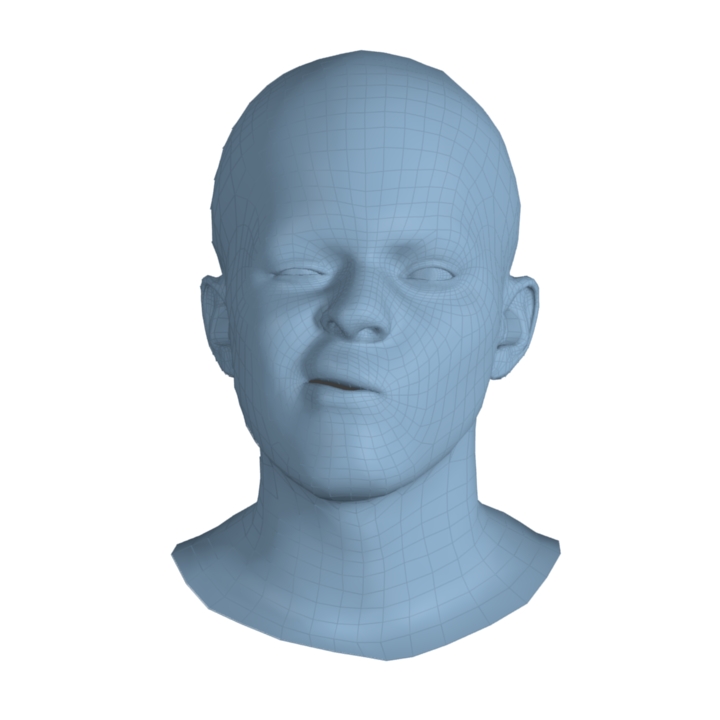}
        \includegraphics[width=\ablmeshsize, clip, trim={\ablimgcropleft} {\ablimgcroplower} {\ablimgcropright} {\ablimgcropupper}]{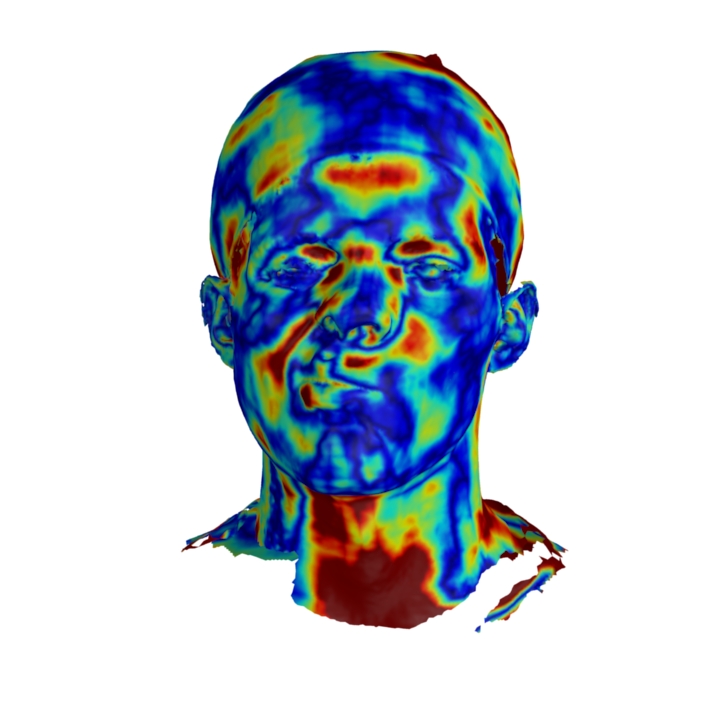}        
        & 
        \includegraphics[width=\ablmeshsize, clip, trim={\ablimgcropleft} {\ablimgcroplower} {\ablimgcropright} {\ablimgcropupper}]{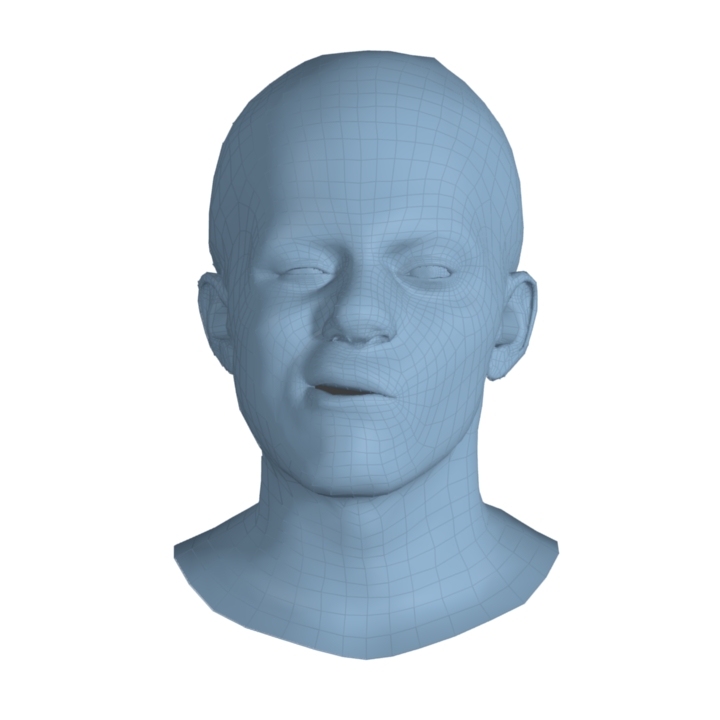}
        \includegraphics[width=\ablmeshsize, clip, trim={\ablimgcropleft} {\ablimgcroplower} {\ablimgcropright} {\ablimgcropupper}]{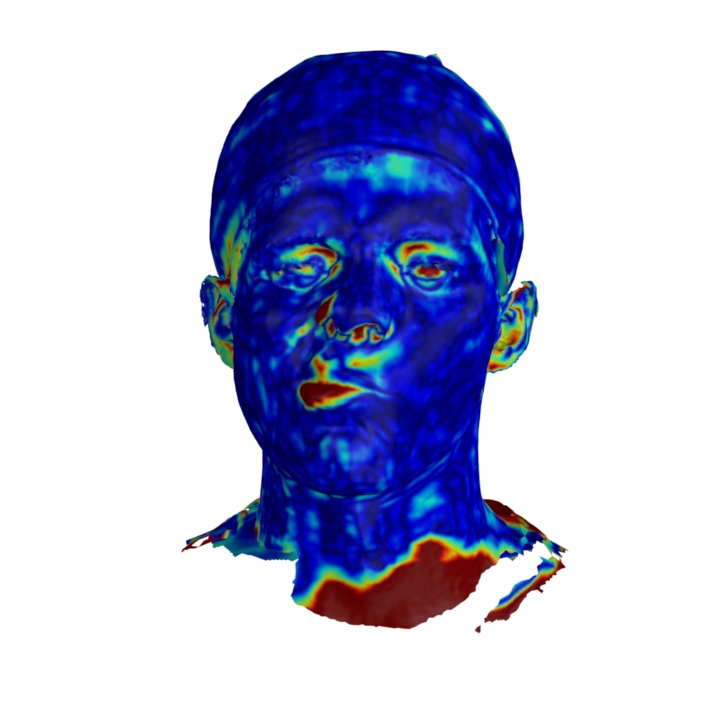}    
        & 
        \includegraphics[width=\ablmeshsize, clip, trim={\ablimgcropleft} {\ablimgcroplower} {\ablimgcropright} {\ablimgcropupper}]{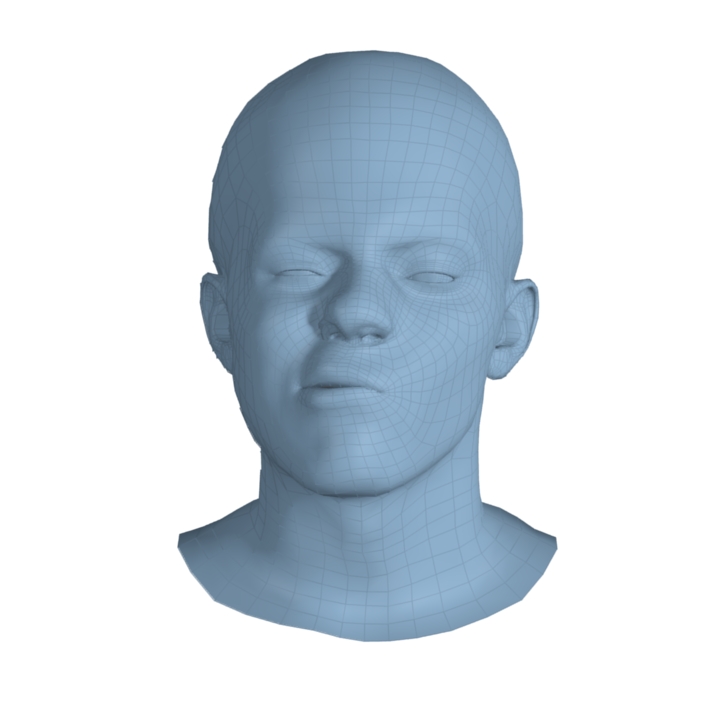}        
        \includegraphics[width=\ablmeshsize, clip, trim={\ablimgcropleft} {\ablimgcroplower} {\ablimgcropright} {\ablimgcropupper}]{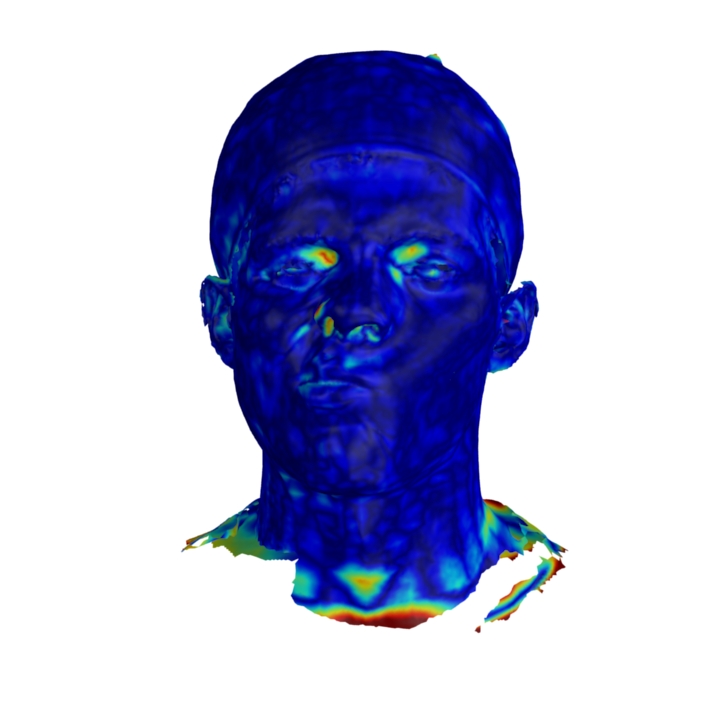}
        & 
        \includegraphics[width=0.05\linewidth]{images/qualitative_results/color_map_3mm.pdf}      
        \\
        Na\"ive feature fusion & Ours w/o s2m & Ours w/o head localization & Ours 
        \\
        \includegraphics[width=\ablmeshsize, clip, trim={\ablimgcropleft} {\ablimgcroplower} {\ablimgcropright} {\ablimgcropupper}]{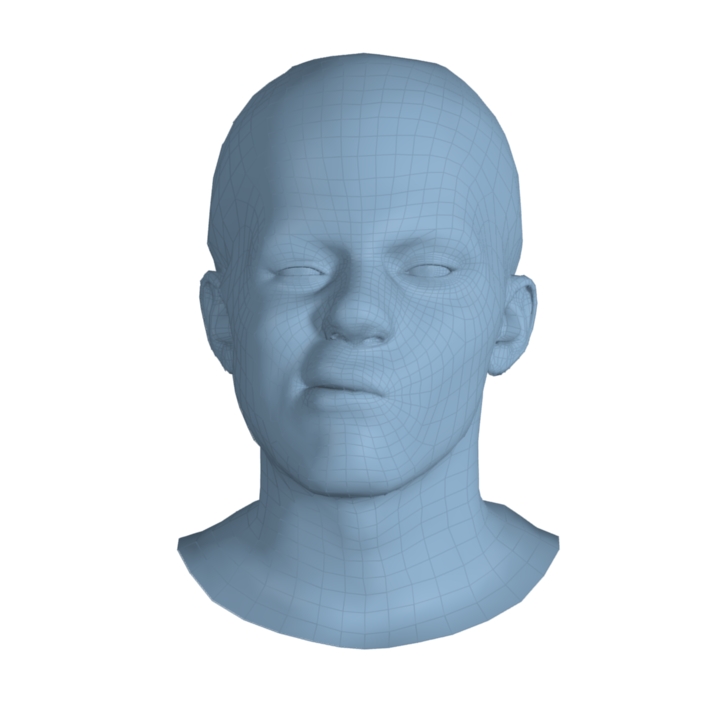}
        \includegraphics[width=\ablmeshsize, clip, trim={\ablimgcropleft} {\ablimgcroplower} {\ablimgcropright} {\ablimgcropupper}]{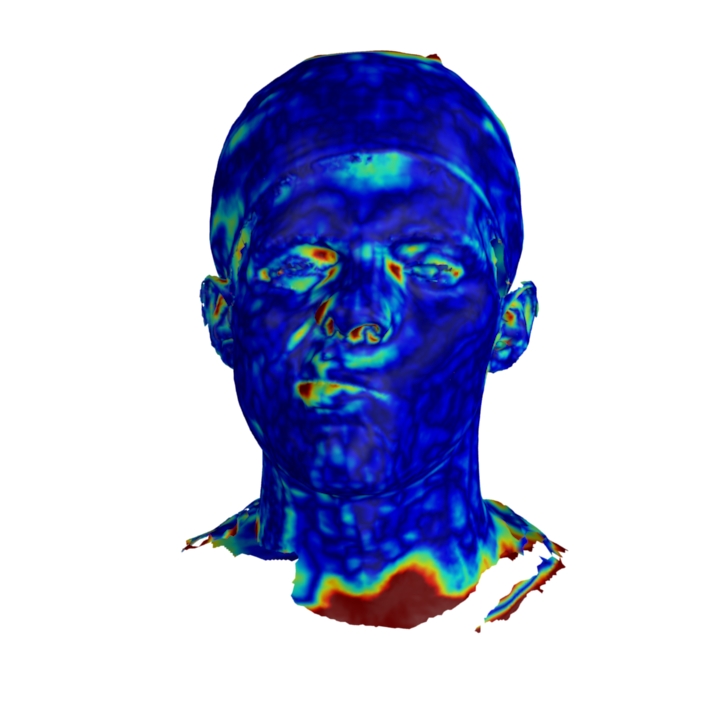}        
        &        
        &
        \includegraphics[width=\ablimgsize]{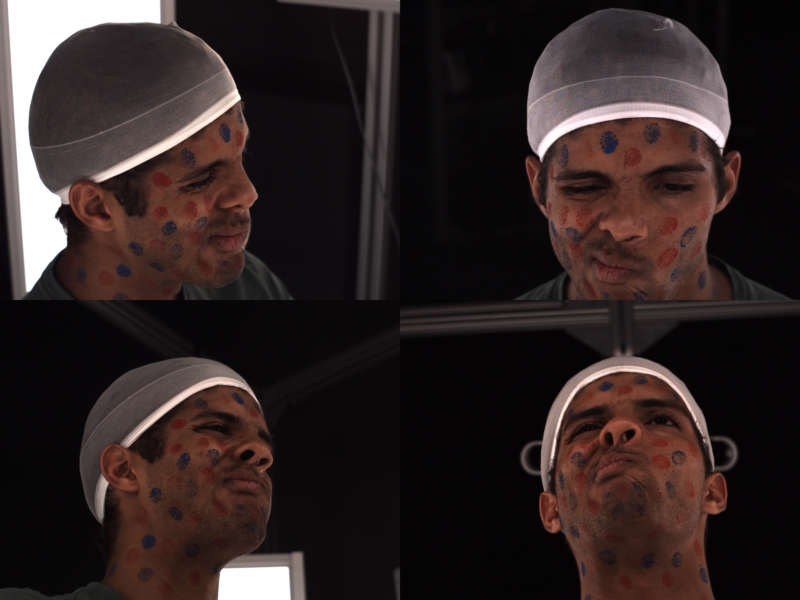}  
        &        
        \includegraphics[width=\ablmeshsize, clip, trim={\ablimgcropleft} {\ablimgcroplower} {\ablimgcropright} {\ablimgcropupper}]{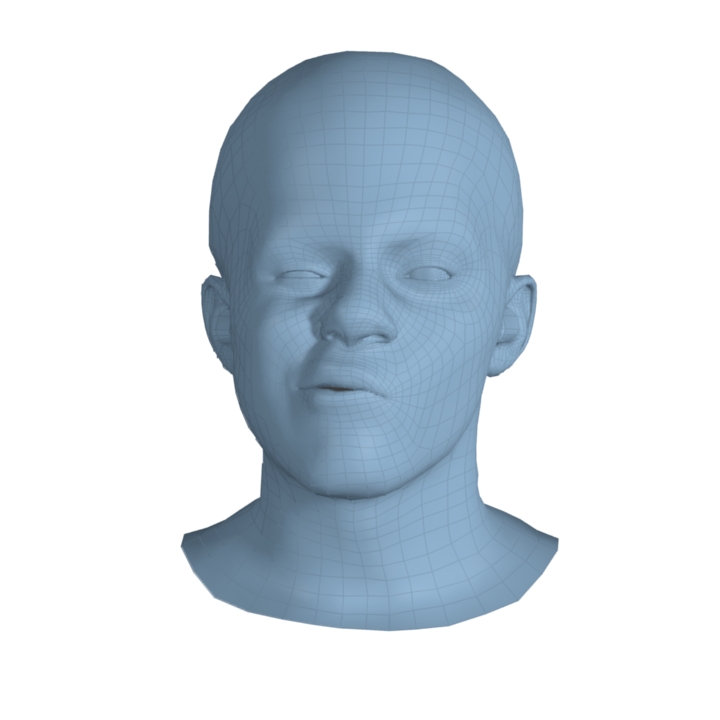}
        \includegraphics[width=\ablmeshsize, clip, trim={\ablimgcropleft} {\ablimgcroplower} {\ablimgcropright} {\ablimgcropupper}]{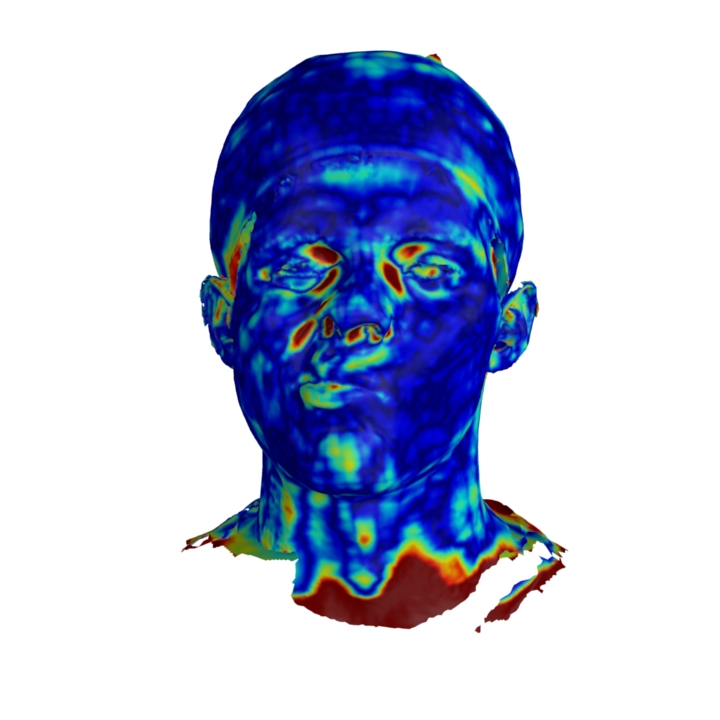}       
        &
        \includegraphics[width=0.05\linewidth]{images/qualitative_results/color_map_3mm.pdf}           
        \\
        Ours hierarchical & & Color input (4 of 8 views) & Ours color
        \\  
    \end{tabular}
	\caption{\textbf{Additional ablation experiments}.
	For each model variant, we show the reconstructed mesh (left) and the color coded point-to-surface distance (right) between reference scan and reconstructed mesh as heatmap on the scan's surface (red means $\geq$ 3 millimeter). 
	}
    \label{fig:ablations2}
\end{figure*}

\paragraph{Ablation experiments:}
Figure~\ref{fig:ablations2} shows additional ablation results.
While the model variants without head localization (Ours w/o head localization) or with a hierarchical architecture (Ours hierarchical) produce reconstructions with low distance to the reference scans, they reconstruct the lip region with lower fidelity than the final model.

\begin{figure}[ht]
    \centering
    \begin{tabular}{c}
        \includegraphics[width=0.35\columnwidth]{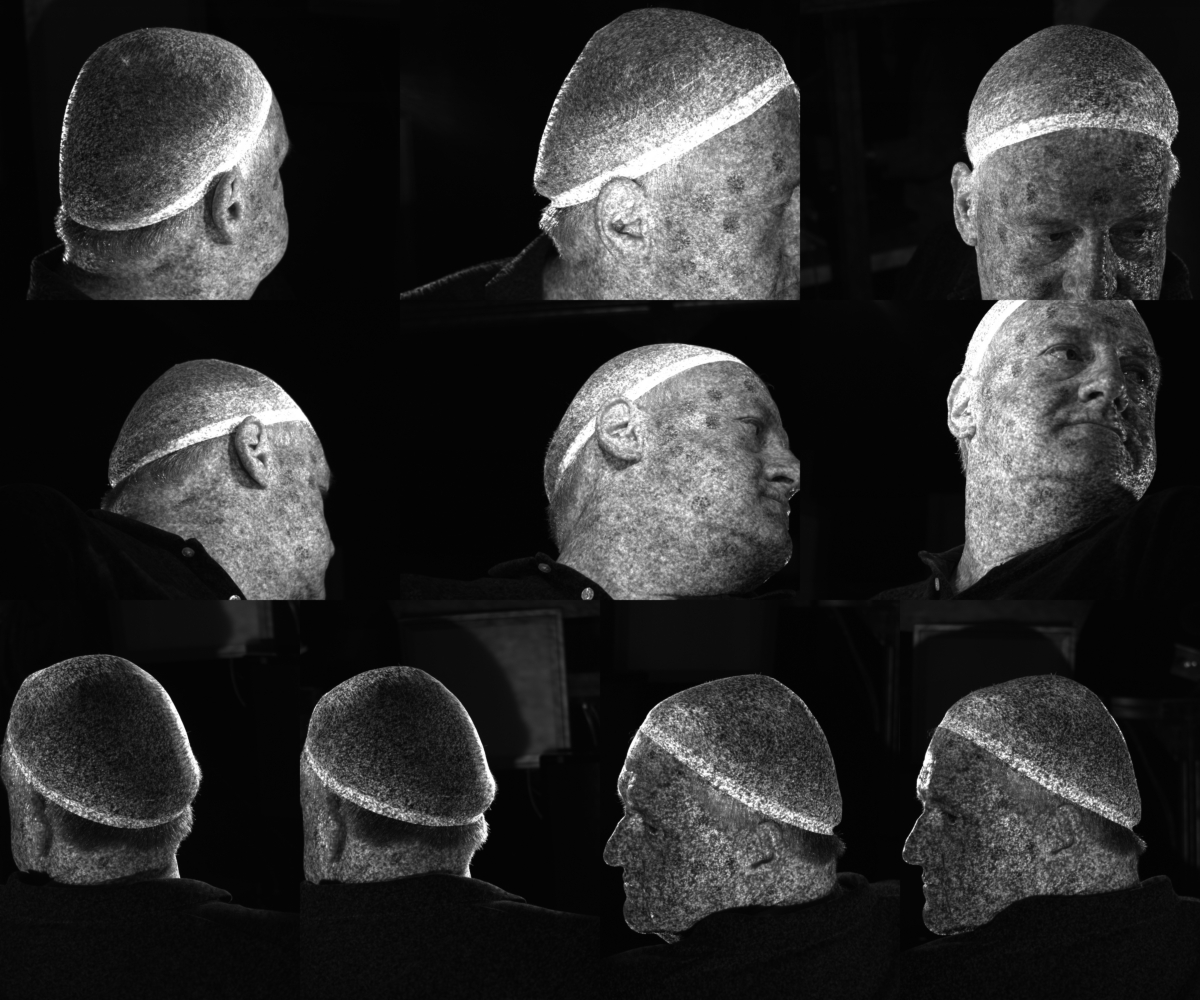}   
        \includegraphics[width=0.25\columnwidth]{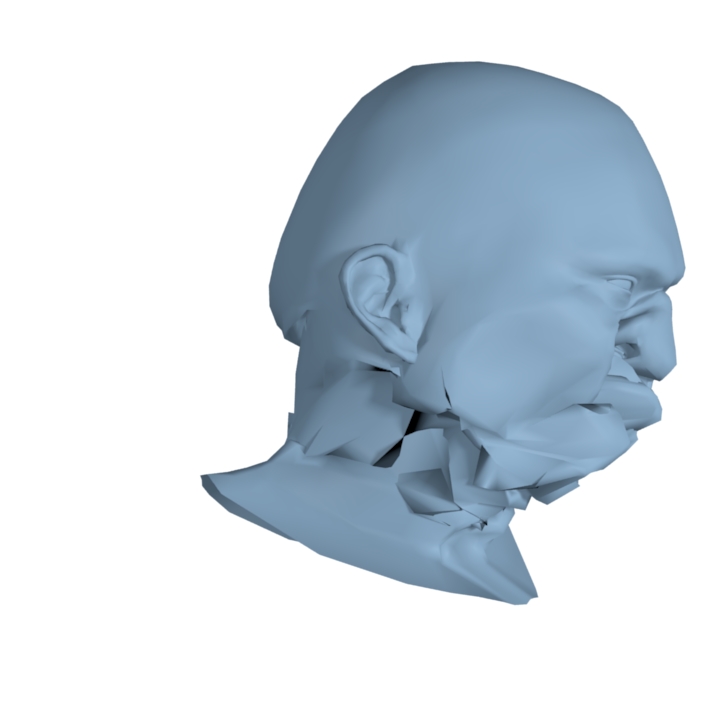} 
        \includegraphics[width=0.25\columnwidth]{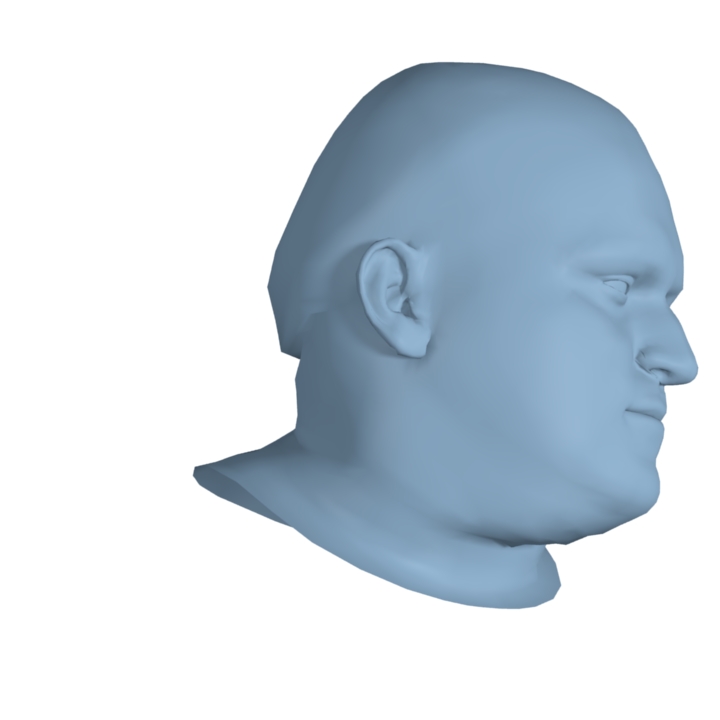}
    \end{tabular}
	\caption{\textbf{Failures}. For input images (left: 10 of 16 views) where the face is occluded in most views, the coarse stage reconstruction can fail (middle), resulting in poorly estimated head meshes. 
    Longer training of the coarse stage can improve the reconstruction performance for such extreme cases (right). 
	}
    \label{fig:failures}
\end{figure}

\paragraph{Failures:}
\modelname's coarse stage reconstruction can fail under large occlusions due to extreme head poses (see Figure~\ref{fig:failures}).
We found empirically that training the coarse stage for $250$K more iterations improves the quality of the reconstructed head meshes for such extreme head poses.

\paragraph{Computational requirements:}
\modelname and the baseline models are trained/evaluated on a computing unit with a single NVIDIA A100-SXM 80 GB GPU and 16 CPU cores.
Training \modelname/ToFu/ToFu+ allocates 26/4/14 GB GPU memory for the coarse stage, 37/34/37 GB for refinement, and up to 21 GB RAM. 
GPU memory is mainly allocated in the volumetric feature sampling and the probability volume prediction for the 2 (batch) $\times5023$ local grids of size $8^3$.
Training \modelname takes $6$ days ($3.5/2.5$ days for coarse / refinement).
Inference for \modelname/ToFu/ToFu+ allocates 6/4/6 GB GPU memory for the coarse stage and 10/6/8 GB for refinement.

\paragraph{Running time evaluation:}
\modelname targets the typical two-step process of reconstructing 3D meshes in correspondence, \ac{MVS}, followed by non-rigid registration. 
This pipeline takes $\geq$ 10 minutes per mesh (Tab.~1 \cite{Li2021_ToFu}), while \modelname takes 0.27s.
While ToFu/ToFu+ is even faster with 0.16/0.18s, \modelname reconstructs 3D heads with a 64\% lower error. 
The time difference between ToFu/ToFu+ and \modelname is mainly due to the visibility computation in the surface-aware feature fusion.
\modelname w/ na\"ive feature fusion requires 0.17s, comparable to ToFu/ToFu+.
The coarse model inference accounts for about 0.03s for all models.
The fast inference speed is due to downsampling of the input images, and due to parallelization. 
Specifically, the feature extraction is parallelized across images (stacked across the batch dimension), while feature sampling \& aggregation, head inference, and mesh refinement are parallelized across all points. 

\paragraph{Data diversity:}
FaMoS data are female/male: 52/41; age 18-34: 65, 35-50: 14, 51-69: 13, 70+: 1; Middle-Eastern: 6, South American: 10, Asian: 24, Pacific Ocean: 1, African: 3, European: 49.
We provide self-identified ethnicity labels as provided by each participant with the dataset. 

\paragraph{Model architecture:}
\modelname uses volumetric features to  localize, infer and then refine the output mesh. 
These features are extracted from the input images with two separate 2D feature extraction networks $\featurenet$, one for coarse head prediction (Section 3.1) and one for head refinement (Section 3.2). 
Both networks use a fully-convolutional U-Net \cite{Ronneberger2015_UNet} architecture with a ResNet34 \cite{He2016_ResNet} backbone. 
Both feature networks take downsampled images as input (i.e., images of size w = 200, h = 150 for the coarse stage, and w = 400, h = 300 for the refinement stage), and output a feature map $\featureimage$ with the same spatial resolution as the image, with a feature dimension of 8. 
We empirically found that adding two additional skip connections for the feature networks compared to ToFu's implementation improved the reconstruction performance of the refinement network. 
For a fair comparison to ToFu and ToFu+, we use the same feature extractor networks with added skip connections for all models.

The reconstruction networks in coarse and refinement stages, $\reconnetcoarse$ and $\reconnet$, respectively, are both 3D U-Nets \cite{Iskakov2019}. 
Similar to ToFu, the coarse stage reconstruction network $\reconnetcoarse$ has five down- and upsampling blocks, with a slight modification of the third last and second last convolution blocks, which output 64 and 128 channels (instead of 32 for ToFu). 
The refinement stage reconstruction network $\reconnet$ follows a similar structure, but with three down- and upsampling layers, same as ToFu.

\end{document}